\documentclass[twoside,11pt]{article}

\usepackage{blindtext}

\usepackage{jmlr2e}

\usepackage{amsmath, amssymb}
\usepackage{mathtools}
\usepackage{algorithm,algorithmic}
\usepackage{mathrsfs}
\usepackage{array}

\usepackage{bm}

\newcommand{\vecT}[1]{#1^\mathrm{T}}
\newcommand{\til}[1]{\tilde{#1}}
\newcommand{\inv}[1]{{#1}^{-1}}

\newcommand{\old}[1]{{#1}_{\textrm{old}}}
\newcommand{\oldInv}[1]{{#1}_{\textrm{old}}^{-1}}

\newcommand{\st}{\textrm{s.t.}}

\DeclareMathOperator*{\argmax}{arg\,max}
\DeclareMathOperator*{\argmin}{arg\,min}

\usepackage{xcolor}

\usepackage{tikz}
\usetikzlibrary{positioning, arrows, automata}
\usepackage{pgfplots}
\pgfplotsset{compat=1.8}
\usepackage{pgfplotstable}
\usepgfplotslibrary{groupplots}
\usepgfplotslibrary{colorbrewer}
\usetikzlibrary{external}
\usetikzlibrary{backgrounds, calc}
\usetikzlibrary{spy}
\usetikzlibrary{fit}
\usetikzlibrary{arrows.meta}
\usepackage{adjustbox}
\usepackage{graphicx}

\definecolor{C0}{HTML}{1f77b4}  
\definecolor{C1}{HTML}{ff7f0e}  
\definecolor{C4}{HTML}{2ca02c}  
\definecolor{C7}{HTML}{d62728}  
\definecolor{C2}{HTML}{9467bd}  
\definecolor{C5}{HTML}{8c564b}  
\definecolor{C6}{HTML}{e377c2}  
\definecolor{C3}{HTML}{7f7f7f}  
\definecolor{C8}{HTML}{bcbd22}  
\definecolor{C9}{HTML}{17becf}
\definecolor{C10}{HTML}{767171}

\definecolor{darkgray176}{RGB}{176,176,176}
\definecolor{lightgray204}{RGB}{204,204,204}

\def\Figref#1{Figure~\ref{#1}}

\def\twoFigref#1#2{Figures \ref{#1} and \ref{#2}}

\def\Apxref#1{Appendix~\ref{#1}}

\def\eqref#1{equation~\ref{#1}}
\def\Eqref#1{Equation~\ref{#1}}

\def\psubref#1{(\protect\subref{#1})}
\def\twopsubref#1#2{(\protect\subref{#1}) and (\protect\subref{#2})}

\newcommand{\ie}{i.\,e.\,}
\newcommand{\eg}{e.\,g.\,}

\usepackage[acronym]{glossaries}
\usepackage{mfirstuc}
\usepackage{hyperref}

\usepackage{multicol}
\usepackage{multirow}
\usepackage{subcaption}
\usepackage[belowskip=3pt]{subcaption}
\usepackage{comment}
\usepackage{setspace}
\usepackage[inline]{enumitem}
\usepackage{wrapfig}
\usepackage{colortbl}
\usepackage{booktabs}

\usepackage{lipsum} 
\usepackage{pgfplots}

\glsdisablehyper

\newacronym{acnmp}{ACNMP}{adaptive conditional neural movement primitive}
\newacronym{ba}{BA}{Bayesian aggregation}
\newacronym{bbrl}{BBRL}{black-box reinforcement learning}
\newacronym{bc}{BC}{boundary condition}
\newacronym{clv}{CLV}{conditional latent variable}
\newacronym{cmore}{CMORE}{contextual model-based relative entropy stochastic search}
\newacronym{cnmp}{CNMP}{conditional neural movement primitive}
\newacronym{cnn}{CNNs}{convolutional neural networks}
\newacronym{cnp}{CNP}{conditional neural processes}
\newacronym{ddpg}{DDPG}{deep deterministic policy gradient}
\newacronym{dmp}{DMP}{dynamic movement primitive}
\newacronym{dof}{DoF}{degree of freedom}
\newacronym{erl}{ERL}{episode-based RL}
\newacronym{es}{ES}{evolution strategies}
\newacronym{gp}{GPs}{Gaussian processes}
\newacronym{idmp}{IDMP}{integral form of dynamic movement primitive}
\newacronym{iqm}{IQM}{interquartile mean}
\newacronym{il}{IL}{imitation learning}
\newacronym{ik}{IK}{inverse kinematics}
\newacronym{ma}{MA}{mean aggregation}
\newacronym{mp}{MP}{movement primitive}
\newacronym{mse}{MSE}{mean squared error}
\newacronym{ndp}{NDP}{neural dynamic policies}
\newacronym{nmp}{ProDMP}{probabilistic dynamic movement primitive}
\newacronym{nn}{NN}{neural networks}
\newacronym{np}{NP}{neural processes}
\newacronym{ode}{ODE}{ordinary differential equation}
\newacronym{pdmp}{ProDMP}{probabilistic dynamic movement primitive}
\newacronym{ppo}{PPO}{proximal policy optimization}
\newacronym{promp}{ProMP}{probabilistic movement primitive}
\newacronym{rl}{RL}{reinforcement learning}
\newacronym{sac}{SAC}{soft actor critic}
\newacronym{srl}{SRL}{step-based RL}
\newacronym{sse}{SSE}{summed squared error}
\newacronym{td3}{TD3}{twin-delayed deep deterministic policy gradient}
\newacronym{trpl}{TRPL}{trust region projection layer}
\newacronym{trpo}{TRPO}{trust region policy optimization}

\newacronym{method}{MP3}{Movement Primitive-based Planning Policy}
\newacronym{method-bb}{MP3-BB}{MP3-Black Box}
\newacronym{method-ppo}{MP3-PPO}{MP3-PPO}
\newacronym{method-bb-ppo}{MP3-BB-PPO}{MP3-BB-PPO}
\newacronym{method-bb-pdmp}{MP3-BB-ProDMP}{MP3-BB-ProDMP}

\newacronym{mprl}{MPRL}{RL with MPs}

\newcommand{\ipb}{\bm{\Phi}}
\newcommand{\ivb}{\dot{\bm{\Phi}}}

\newcommand{\ipbb}{\bm{\Phi}_b}
\newcommand{\ivbb}{\dot{\bm{\Phi}}_b}

\newcommand{\ipbbt}{\bm{\Phi}^\intercal_b}
\newcommand{\ivbbt}{\dot{\bm{\Phi}}^\intercal_b}

\usepackage{lastpage}
\jmlrheading{}{2023}{1-\pageref{LastPage}}{6/23; Revised X/XX}{X/XX}{21-0000}{Fabian Otto and Hongyi Zhou}

\ShortHeadings{MP3:Movement Primitive-Based (Re-)Planning Policy}{Fabian Otto, Hongyi Zhou et al.}
\firstpageno{1}

\begin{document}
    \title{MP3: Movement Primitive-Based (Re-)Planning Policy}

\author{\name Fabian Otto\thanks{Equal contribution: order determined by alphabet.} \email fabian.otto@bosch.com \\
      \addr Bosch Center for Artificial Intelligence\\
      University of Tübingen, Tübingen, 72076, Germany
      \AND
      \name Hongyi Zhou$^*$ \email  hongyi.zhou@kit.edu \\
      \addr Intuitive Robots Lab\\
      Karlsruhe Institute of Technology, Karlsruhe, 76131, Germany
      \AND
      \name Onur Celik \email celik@kit.edu \\
      \addr Autonomous Learning Robots Lab\\
      Karlsruhe Institute of Technology, Karlsruhe, 76131, Germany
      \AND
      \name Ge Li \email ge.li@kit.edu \\
      \addr Autonomous Learning Robots Lab\\
      Karlsruhe Institute of Technology, Karlsruhe, 76131, Germany
      \AND
      \name Rudolf Lioutikov \email lioutikov@kit.edu \\
      \addr Intuitive Robots Lab\\
      Karlsruhe Institute of Technology, Karlsruhe, 76131, Germany
      \AND
      \name Gerhard Neumann \email gerhard.neumann@kit.edu \\
      \addr Autonomous Learning Robots Lab\\
      Karlsruhe Institute of Technology, Karlsruhe, 76131, Germany
      }

\editor{My editor}

\maketitle

\begin{abstract}
We introduce a novel deep \gls{rl} approach called \gls{method}. By integrating \glspl{mp} into the deep \gls{rl} framework, \gls{method} enables the generation of smooth trajectories throughout the whole learning process while effectively learning from sparse and non-Markovian rewards. Additionally, \gls{method} maintains the capability to adapt to changes in the environment during execution.
Although many early successes in robot \gls{rl} have been achieved by combining \gls{rl} with \glspl{mp}, these approaches are often limited to learning single stroke-based motions, lacking the ability to adapt to task variations or adjust motions during execution. 
Building upon our previous work, which introduced an episode-based \gls{rl} method for the non-linear adaptation of \gls{mp} parameters to different task variations, this paper extends the approach to incorporating replanning strategies. This allows adaptation of the \gls{mp} parameters throughout motion execution, addressing the lack of online motion adaptation in stochastic domains requiring feedback.
We compared our approach against state-of-the-art deep \gls{rl} and \gls{rl} with \glspl{mp} methods. The results demonstrated improved performance in sophisticated, sparse reward settings and in domains requiring replanning. The video demonstration can be accessed at \url{https://intuitive-robots.github.io/mp3_website/}.

\end{abstract}

\glsresetall
\section{Introduction}
\Glspl{mp} are a powerful and versatile method for representing robot trajectories with a concise set of parameters. 
This makes \glspl{mp} an easy-to-use and efficient tool for \gls{rl} tasks. 
By directly exploring the space of desired trajectories, \glspl{mp} simplify the exploration process and produce smooth, ``robot-friendly'' motions. 
As a result, \gls{mprl} has been responsible for many of the early successes in robot \gls{rl}, with notable applications in Table-Tennis \citep{mulling2013learning, gomez2016using}, Ball-In-A-Cup \citep{kormushev2010robot}, and Pan-Cake-Flipping \citep{kormushev2013reinforcement}.
However, these algorithms were previously limited to simple setups and could only learn single stroke-based open-loop motions. Consequently, these motions were difficult to adapt to task variations or during motion execution. 
With increasing computation power, the field of deep \gls{rl} rose. These methods can learn complex closed-loop sensorimotor policies, which is the reason why this research field dominated recent years.
In this paper, we extend our work given in \cite{Otto2021} and \cite{otto2022deep} to address the shortcomings of previous \gls{rl} with \glspl{mp} approaches and propose a new method that integrates \glspl{mp} into a deep \gls{rl} pipeline. 
Our method allows for non-linear adaptation and replanning during execution of the \gls{mp}, while still maintaining the beneficial exploration properties of the \gls{mp} framework in the context of \gls{rl}. 

Traditional deep \gls{rl} methods use a step-based policy, where at each time step the policy explores in the atomic action space. 
During interaction with the environment, the agent collects state, action, and reward data points at each time step, which are used to update the policy.
Although using every atomic action generates a vast amount of data-points for the policy update, it also complicates exploration due to the typical random walk behavior and introduces a lot of noise in the policy evaluation process (see \Figref{fig:replan_trajectories}). 
Therefore, these methods often rely on informative reward signals throughout the interaction sequence, making them less effective in sparse or temporally sparse settings where feedback from the environment is delayed.
Moreover, step-based exploration can result in slower convergence and jerky, potentially dangerous behavior for the agent. 

In contrast, \gls{mprl} is typically based on \gls{erl} \citep{Deisenroth2013,Abdolmaleki2015,Daniel2012,otto2022deep}. 
\gls{erl} methods learn to parameterize a desired trajectory used for a controller based on a task description known as the context, which remains fixed throughout the entire episode.
For example, in a table tennis scenario, the context is given by the target position where the robot has to return the ball. 
These methods explore the trajectory space, meaning that a parameter is sampled given the context only once at the beginning of the episode and executed without resampling.
This exploration strategy results in time-correlated exploration, smooth behaviors, and improved performance in sparse or non-Markovian reward settings \citep{otto2022deep}. 
Yet, only one data point is generated per executed trajectory, as these algorithms collect only one context-parameter pair per episode. This data collection procedure limits sample efficiency.
In our recent work \citep{otto2022deep}, we integrated \gls{erl} with \glspl{mp} into a deep policy gradient algorithm that is based on \gls{trpl} \citep{Otto2021}. 
While this algorithm can non-linearly adapt the parameters of the MP to the given context and achieve high-quality policies for complex robotic tasks, it is inherently constrained to generating open-loop trajectories that cannot be adapted or adjusted during execution.

This paper is an extension of \citet{otto2022deep}, where we also add learning non-linear replanning policies instead of just the initial adaptation of the \gls{mp} to the context, combining the benefits of \gls{erl} with \glspl{mp} and \gls{srl} methods. 
We still explore in the trajectory space, yet, the agent is now able to change the desired trajectory during an episode, enabling it to adapt its behavior to unpredictable changes in the environment. 
In the original paper \citep{otto2022deep}, \glspl{promp} \citep{Paraschos2013} were used as the \gls{mp} representation. However, \glspl{promp} are unable to generate smooth trajectories if the \gls{mp} parameters are changed during motion execution.
As a result, \glspl{promp} are not suitable for learning replanning policies. 
In contrast, in this paper, we employ the recently introduced \glspl{pdmp} \citep{li2022prodmps} to address the limitations of commonly used \glspl{mp} \citep{Schaal2005, Schaal2006, Paraschos2013}. 
\gls{pdmp} constitute a reformulation of the popular \gls{dmp} \citep{Schaal2006, Ijspeert2013} framework which is better suited to be integrated into a neural network architecture as it does not require expensive numerical integration. Additionally, \gls{pdmp} can use any initial state as the initial condition, allowing for the generation of smooth desired trajectories even when applying replanning. 
Our policies are parameterized with neural networks and efficiently trained using \gls{trpl} \citep{otto2022deep}, which has demonstrated significantly improved stability and quality of the learned policy in comparison to other policy gradient methods, such as \gls{ppo} \citep{Schulman2017}. 
We demonstrate the effectiveness of our method by presenting various complex simulated robotic tasks, such as robot table tennis, beer-pong, a complex box-pushing task, and large-scale manipulation tasks on Meta-World \citep{Yu2019}. 
We compare our approach to state-of-the-art \gls{srl} and \gls{erl} methods and illustrate improved performance in sophisticated, sparse reward settings and settings that require replanning.
\section{Background and Related Work}
\paragraph{(Step-Based) Reinforcement Learning.}
In \gls{srl}, direct interactions with the environment involve using raw action inputs based on the current state. 
These predicted actions typically correspond to positions, velocities, or torques in robotic tasks. They represent the specific control signals that are applied to the robot's actuators, influencing its movements and behavior.
On-policy \gls{rl} refers to the methods that use the same policy for exploration and update. 
Here, policy gradient methods \citep{Sutton1999, Schulman2015,Schulman2017} are often used as they are straightforward to implement and yield high-quality policies.
These methods typically apply some form of trust region to stabilize the on-policy update due to the large variance of the gradients. 
\gls{ppo} \citep{Schulman2017} is one of the most commonly used methods. It introduced a clipping heuristic to the policy gradient objective as an approximation of the trust region. 
It has shown robust and efficient performance even in large-scale problems, such as OpenAI Five \citep{berner2019dota} and GPT-4 \citep{openai2023gpt}, but can depend heavily on the implementation details \citep{Engstrom2020} and uses ad-hoc heuristics that do not work well for complex exploration problems \citep{Otto2021,otto2022deep}.
In contrast to the above methods, \gls{sac} and \gls{td3} are off-policy methods.
They tend to exhibit higher sample efficiency, but are often computationally more expensive. Additionally, they may introduce a higher bias in the policy update due to the reliance on the critic network to evaluate the actions. We choose to compare our algorithm against \gls{ppo} and \gls{sac} as these are currently the most popular on-policy and off-policy RL algorithms. 

\paragraph{Trust Region Projection Layers.}
Enforcing trust regions is an established technique to ensure the stability and convergence of policy gradient methods.
While existing efficient methods, such as \gls{ppo}, use surrogate cost functions as approximations of trust regions, they cannot enforce the trust region exactly.
This limitation was addressed in \acrfullpl{trpl} \citep{Otto2021}, which presents a mathematically sound and scalable approach to enforce exact trust regions in deep \gls{srl}.
The algorithm efficiently enforces a trust region for each input state of the policy using differentiable convex optimization layers \citep{agrawal2019differentiable}, providing more stability and control during training and at the same time reduce the dependency on implementation choices \citep{Engstrom2020}. 

The layer receives the standard output of a Gaussian policy, which includes the mean $\bm{\mu}$ and covariance $\bm{\Sigma}$, and applies a projection operation to enforce a trust region when the mean and covariance exceed their respective bounds. 
This projection is performed individually for each input state. 
The resulting Gaussian policy distribution, characterized by the adjusted parameters $\tilde{\bm{\mu}}$ and $\til{\bm{\Sigma}}$, is then used for subsequent computations, \eg sampling and loss computation.
Formally, the layer solves the following two optimization problems for each state $\bm{s}$
\begin{equation*}
    \label{eq:generic_projection}
    \begin{split}
        \argmin_{\til{\bm{\mu}}_s} d_\textrm{mean} \left(\til{\bm{\mu}}_s, \bm{\mu}(s) \right),  \quad &\st \quad d_\textrm{mean} \left(\til{\bm{\mu}}_s,  \old{\bm{\mu}}(s) \right) \leq \epsilon_{\bm{\mu}}, \quad \textrm{and}\\ 
        \argmin_{\til{\bm{\Sigma}}_s} d_\textrm{cov} \left(\til{\bm{\Sigma}}_s, \bm{\Sigma}(\bm{s}) \right), \quad &\st \quad d_\textrm{cov} \left(\til{\bm{\Sigma}}_s, \old{\bm{\Sigma}}(\bm{s}) \right) \leq \epsilon_\Sigma,
    \end{split}
\end{equation*}
where $\tilde{\bm{\mu}}_s$ and $\tilde{ \bm{\Sigma}}_s$ are the optimization variables for input state $\bm{s}$ and $\epsilon_\mu$ and $\epsilon_\Sigma$ are the trust region bounds for mean and covariance, respectively.
We measure the similarity between means and covariances using $d_\textrm{mean}$ and $d_\textrm{cov}$, which are decomposable similarity measures.
Although \citet{Otto2021} proposed three such measures(KL, Wasserstein, and Frobenius), we only use the decomposed KL-divergence in this work as it showed the most promising performance in the original work. 
This measure can be made fully differentiable by following the method described in \cite{Otto2021} and is explained in more detail in \Apxref{app:kl_trpl}.

\paragraph{Episode-based Reinforcement Learning.}
In the framework of contextual episode-based policy search \citep{Deisenroth2013,Daniel2012}, \gls{rl} is treated as a black-box optimization problem. 
The goal is to maximize the expected return $R(\bm{w},\bm{c})$ by optimizing a contextual search distribution $\pi(\bm{w}|\bm{c})$ over the controller parameters $\bm{w}$. Here, the context vector $\bm{c}$ characterizes the given task, for instance, given by a goal location or the location of an object, and the controller is typically given by a motion primitive. 
The objective  can be expressed as 
\begin{align*}
    \argmax_{\pi(\bm{w}|\bm{c})} \mathbb{E}_{p(\bm{c})}\left[\mathbb{E}_{\pi(\bm{w}|\bm{c})}[R(\bm{w},\bm{c})] \right],
    \label{eq:eps_obj_context}
\end{align*}
where $p(\bm{c})$ denotes the context distribution given by the task. 
The return function $R(\bm{w},\bm{c})$ is not subject to any structural assumptions, and it can be any non-Markovian function of the resulting trajectory due to the black-box nature of the problem.
Most \gls{erl} algorithms are focused on the non-contextual setting, where different optimization techniques have been used, such as policy gradients \citep{Sehnke2010}, natural gradients \citep{Wierstra2014}, stochastic search strategies \citep{Hansen2001,Mannor2003,Abdolmaleki2019}, or trust-region optimization techniques \citep{Abdolmaleki2015,Daniel2012,Tangkaratt2017}. 
Early methods that incorporate context adaptation \citep{Tangkaratt2017,Abdolmaleki2019} only consider a linear mapping from context to parameter space, which is a major limitation on the performance of these approaches. 
In contrast, we consider highly nonlinear context-parameter relationships using neural networks. 

\paragraph{Evolutionary Strategies.}
Recent work from the domain of evolutionary strategies \citep{Mania2018, Salimans2017, Chrabaszcz2018} have also proposed full gradient-free black-box approaches as an alternative to gradient- and step-based methods for finding optimal neural network parameters.
These approaches treat learning the neural network policy parameters (several thousand parameters) as the black-box optimization problem as opposed to learning movement primitive parameters (20-50 parameters). 
While these methods can be competitive for black-box optimization of neural networks, they typically do not consider contextual setups where different parameters have to be chosen for different contexts, as it is common for movement primitives.  
In the contextual case, the performance of the rollout depends on both the parameter vector and the context (for instance, the goal), introducing additional noise to the evaluation if the method is context ignorant. 
For example, a good neural network parametrization can still yield a poor performance because it has been evaluated in a difficult context. The above approaches are completely ignorant to the context and, therefore, are not competitive for complex contextual scenarios as shown in our experiments. 

In contrast, we do not engage in black-box optimization at the level of a global neural network control policy, which would involve thousands of parameters. Rather, we focus on local control parameters of movement primitives or similar controllers, which only have $10-50$ dimensions. 
We still use neural network policy with a substantial number of parameters to predict these local control parameters based on the context.
Instead of optimizing these neural network parameters in a black-box manner, we update them via policy gradient, which uses both the context information as well as the policy derivatives.

\paragraph{Reinforcement Learning with Movement Primitives.}
While most works of \gls{mprl} concentrate on learning a single \gls{mp} parameter vector for a single task configuration \citep{Abdolmaleki2015,Kober2008,Stulp2012,Stulp2012b}, some methods allow linear adaptation of the \gls{mp}'s parameter vector to the context \citep{Daniel2012, Kupcsik2017,celik2022specializing}. 
In addition, a few \gls{rl} approaches leverage non-linear policies combined with predefined action primitives, such as pushing or grasping motions \citep{dalal2021accelerating,zenkri2022hierarchical}.
One approach that directly uses \glspl{mp} and deep networks in a \gls{srl} setting is \gls{ndp} \citep{Bahl2020}. 
\gls{ndp} aims to embed the structure of \glspl{dmp} into deep policies by reparameterizing action spaces via second-order differential equations. 
This can be seen as an intersection between step-based and episode-based methods by learning sub-trajectories via \glspl{dmp} spanning multiple timesteps.
While their approach enables effective replanning, unlike our approach, their main exploration still takes place at the action level rather than at the trajectory level, similar to standard step-based approaches, which neglects the main benefit of using \glspl{mp} in an \gls{rl} context. Moreover, using \glspl{dmp} requires several numerical integration steps that also need to be differentiated, making it computationally expensive.
In our experiments, \gls{ndp} performs poorly, which contradicts the original paper. This is mainly because we use a different measure for sample efficiency. In contrast to the original paper where only the number of samples used in policy updates are counted, we report the total number of environment interactions as we believe this choice provides a more straightforward insight into the training cost associated with each method.

\section{Deep Reinforcement Learning with Movement Primitives}
In this work, we present a framework to effectively combine \glspl{mp} with deep \gls{rl} methods. 
This framework consists of three major components (see \Figref{fig:overview}):
\begin{itemize}
    \item[{\color{C0}\textbullet}] One \gls{rl} policy which takes the environment observation as input and outputs an MP weight vector that is used for multiple time steps.
    \item[{\color{C4}\textbullet}] One \gls{mp} model which uses the weight vector as input to generate a desired trajectory. 
    \item[{\color{C1}\textbullet}] One low-level controller which converts the desired trajectory into raw actions and interacts with the environment.
\end{itemize}

\begin{figure}[tbp]
    \centering
    \resizebox{\textwidth}{!}{%
    \input{tikz_plots/diagrams/updated_overview/updated_overview_tikz.tex}%
    }%
    \caption{
    This figure provides an overview of the proposed framework that combines deep \gls{rl} with \glspl{mp}. 
    Instead of generating a raw action directly, the policy generates a set of weights that parameterizes a \gls{mp}. 
    The \gls{mp} predicts a desired trajectory given the weights and initial conditions, which is then converted to raw actions using a tracking controller. 
    }%
    \label{fig:overview}
\end{figure}

This approach is simple but highly versatile. 
Theoretically, any policy search algorithm applicable to continuous action spaces can be used here. 
However, it is worth noting that the dimensionality of the weight space of the \gls{mp} is usually larger than the raw action space.
Therefore, the policy search algorithm used must be able to explore high-dimensional spaces efficiently. 
Furthermore, \glspl{mp} can be replaced by any parameterized trajectory generator, as long as the desired trajectory can be uniquely determined by a weight vector. We specifically chose \glspl{mp} in this work because they are capable of generating smooth trajectories and allowing effective replanning.
In addition, the planning horizon (length of the generated trajectory before a new weight vector is chosen) can vary from a single step to the entire episode. 
Two special cases correspond to two common \gls{rl} paradigms: 
(i) When the planning horizon is equal to one, our framework is similar to an \gls{srl} algorithm (although with a higher dimensional action space).
(ii) When it is equal to the episode length, the framework corresponds to an \gls{erl} algorithm.

\subsection{Movement Primitives}
\Glspl{mp} are a widely used tool for motion representation and generation in robotics. 
They are used as building blocks for movements, allowing for the modulation of motion behavior and the creation of more complex movements through combination or concatenation. 
With their concise parameterization and flexibility, \glspl{mp} have become a popular choice in imitation learning \citep{maeda2014learning, gomez2016using, ridge2020training, li2022prodmps, rozo2022orientation} and \gls{rl} \citep{celik2022specializing, li2017reinforcement}.
In this work, we focus on using trajectory-based movement representations, as discussed in previous work for \gls{rl} \citep{Schaal2006, Ijspeert2013, Paraschos2013, li2022prodmps}.
Given a parameter vector, these representations generate desired trajectories for the agent to follow, and the goal is often to optimize the weight vector or its distribution to improve the resulting movements.
\paragraph{Probabilistic Movement Primitives (ProMPs).}
ProMPs \citep{Paraschos2013} generate the trajectory $\bm{\lambda}$ by a linear basis function model 
\begin{equation*}
    y(t) = \bm{\Phi}^\intercal(t) \bm{w},  \quad \bm{\lambda} = [y_t]_{t=0:T} = \bm{\Phi}_{0:T}^\intercal \bm{w},
\end{equation*}
where $\bm{w}$ is the time-independent weight vector, $y(t)$ is the trajectory position at time step $t$ and $\bm{\Phi}$ are pre-defined time-dependent basis functions, for instance, normalized RBF basis. 
Due to the simplicity of the linear basis function representation, \glspl{promp} allow for fast trajectory computation and enable modeling the trajectory's statistics from the weight vector's distribution. 
Such statistics often include temporal correlations of the trajectories and the motion correlations across different \gls{dof}.
One of the key limitations of \glspl{promp} as a representation method is their lack of smoothness in trajectory replanning and concatenation, which means when choosing a $\bm{w}$, there is no guarantee that the trajectory will start from the desired initial conditions at the current time step.  
This lack of smoothness and inability to adjust the trajectory's start state poses limitations in the practical application of \glspl{promp}, especially in situations where the weight vector needs to be updated throughout motion execution due to unpredictable changes in the environment.

\paragraph{Dynamic Movement Primitives (DMPs).}
\glspl{dmp} \citep{Schaal2006, Ijspeert2013} form a trajectory by integrating a dynamic system, providing smooth replanning of both position and velocity \citep{brandherm2019learning, ginesi2019knowledge, lee2020trajectory}. 
However, this smoothness comes at a computational cost, as \glspl{dmp} require online numerical integration to compute a trajectory. 
When using \glspl{dmp} in neural networks, the forward and backward pass of the networks become coupled with this numerical integration \citep{Bahl2020, gams2018deep, ridge2020training}, leading to complex and slow models. 

\paragraph{Probabilistic Dynamic Movement Primitives (ProDMPs).}
To combine the advantages and address the limitations of \glspl{promp} and \glspl{dmp}, \citet{li2022prodmps} recently proposed \glspl{pdmp} by solving the \gls{dmp}'s underlying \gls{ode}. 
The expensive online numerical integration of \glspl{dmp} is replaced by position and velocity basis functions, which can be computed offline and shared by all trajectories.
This results in a representation of trajectory position and velocity that is similar to that of \glspl{promp}, expressed as:
\begin{equation*}
    y(t) = c_1y_1(t) + c_2y_2(t) + \ipb(t)^\intercal \bm{w}_g, \quad \dot{y}(t)=c_1\dot{y}_1(t) + c_2\dot{y}_2(t) + \ivb(t)^\intercal\bm{w}_g, \label{eq:prodmp}
\end{equation*}
where $y_1$ and $y_2$ denote the two linearly independent complementary functions of the governing \gls{ode} of \glspl{dmp}.
Their corresponding derivatives w.r.t time are $\dot{y}_1$, $\dot{y}_2$. 
The coefficients $c_1$ and $c_2$, which are shared by both position and velocity representations, are determined by solving an initial condition problem of the \gls{ode}. 
The position and velocity basis functions, denoted by $\ipb$ and $\ivb$ respectively, can be computed only once offline and later used as constant functions. 
The vector $\bm{w}_g$ in both equations concatenates the \gls{dmp}'s original weights $\bm{w}$ and goal attractor $g$ into one vector. 
To simplify our notation, we will refer to all learned parameters collectively as $\bm{w}$ from this point forward.
A brief derivation of these equations is provided in \Apxref{app:pdmp}, for the further details, we refer the reader to the original paper \citep{li2022prodmps}.

Theoretically, both \gls{dmp} and \gls{pdmp} can be used in domains requiring online replanning.
In this work, we specifically use the \glspl{pdmp} model as our trajectory generator because it ensures smooth replanning with low computational cost.

\subsection{Reinforcement Learning Objective with Movement Primitives}
While traditional \gls{srl} methods rely on single raw actions $\bm{a}_t \in \mathcal{A}$ per time step, we train a policy to select a weights vector $\bm{w}_t \in \mathcal{W}$ in \gls{mp}'s parameter space $\mathcal{W}$. The weights vector is then translated to a desired trajectory of the proprioceptive states $\bm{\lambda}^d = (\bm{q}^d_{t+1}, \bm{q}^d_{t+2}, \ldots, \bm{q}^d_{t+k})$, where $q^d_t=[\bm{y}^d_t, \dot{\bm{y}}^d_t]$ consists of desired position $\bm{y}^d_t$ and desired velocity $\dot{\bm{y}}^d_t$ at time step $t$, and $k$ denotes the planning horizon.
Given the desired trajectory and the measured proprioceptive state, a tracking controller $f(q_{t}^d, q_t)$ decides the action at each step, resulting in a trajectory in the raw action space $(\bm{a}_{t+1}, \bm{a}_{t+2}, \ldots, \bm{a}_{t+k}) \in \mathcal{A}$. 
In contrast to the step-wise sample $(\bm{s}_t, \bm{a}_t, R_t)$ used in \gls{srl}, we use temporarily-abstracted samples of the form $(\bm{s}_t, \bm{w}_t, R^k_t)$.   
The reward $R^k_t = R_{t:t+k-1}$ of each trajectory segment is defined as the cumulative reward over all the segment's time steps $t$ to $t+k-1$
\begin{align}
    R^k_{t}(\bm{s}_t,\bm{w}_t) = \sum_{i=0}^{{k-1}} \gamma^i r(\bm{s}_{t+i}, \bm{a}_{t+i}),
\end{align}
where $a_t$ and $s_t$ are the executed actions and observed states following the desired trajectory and tracked by the controller. 
While our approach supports different $k$ for each segment, we only consider planning segments with equal length in this work.
We can compute the episode return by taking the cumulative discounted sum of the segment rewards. 
Using the notation from above, we can express this as
\begin{align}
        G^k_t &= \sum_{i=0}^{\lceil T/k - 1\rceil} \gamma^{ik} R^{k}_{t+ki}(\bm{w}_{t+ki}, \bm{s}_{t+ki}),
\end{align}
where $\gamma \in (0,1]$ is the discount factor. 
It is worth noting that there are two special cases to consider. 
In the black-box setting, in other words, when the MP parameters are chosen only at the beginning of the episode, then  $k=T$ and the segment reward equals the episode return
\begin{align}
    R^{T-1}_0 = \sum_{t=0}^{T-1} \gamma^t r(a_t, s_t).
\end{align}
The second special case is step-based RL. That is, we choose a new parameter vector at every time step and  $k=1$. In this case, segment reward is equivalent to step reward
\begin{align}
        R^{1}_t = r(a_{t}, s_{t}).
\end{align}
This gives the insight that we can alter between \gls{srl} and \gls{erl} by choosing different planning horizons $k$.

\subsection{Policy-gradients for MP weight-selection policies}
With these rewards, we can now also define matching value and advantage functions 
\begin{align}
    V^\pi(\bm{s}) = \mathbb{E} \left[ G^k_t | \bm{s}_t = \bm{s}; \pi_{\theta} \right] \quad\quad
    A^{\pi}(\bm{s}, \bm{w}) = \mathbb{E} \left[ G^k_t | \bm{s}_t = \bm{s}, \bm{w}_t = \bm{w}; \pi_{\theta} \right] - V^{\pi}(\bm{s}).
\end{align}
Following the step-based policy gradient \citep{Williams1992,Schulman2015}, we optimize the advantage function using the likelihood ratio gradient and an importance sampling estimator. 
The resulting objective
\begin{align}
     \hat{J}(\pi_{\bm{\theta}}, \pi_{\old{\bm{\theta}}}) = \mathbb{E}_{({\bm{s}},{\bm{w}})\sim p(\bm{s}),\pi_{\old{\bm{\theta}}}} \left[\frac{{{\pi}_{\bm{\theta}}}({\bm{w}}\vert{\bm{s}})}{\pi_{\old{\bm{\theta}}}({\bm{w}}\vert{\bm{s}})} A^{\pi_{\old{\bm{\theta}}}}({\bm{s}}, {\bm{w}})\right],
    \label{eq:objective}
\end{align} 
is maximized w.r.t $\bm{\theta}$,
with $\pi_{\old{\bm{\theta}}}$ being the old behavior policy used for sampling. 
We can further make use of a learned state-value function $V_\phi(\bm{s}) \approx V^\pi(\bm{s})$ for the advantage estimator, which is approximated by optimizing
\begin{equation}
    \argmin_\phi \mathbb{E}_{(\bm{s},\bm{w})\sim p(\bm{s}),\pi_{\old{\bm{\theta}}}} \left[\big(V_\phi(\bm{s}) - G^k_t\big)^2\right].
\end{equation}
This formulation also enables the use of advantage estimation methods, such as general advantage estimation \citep{Schulman2016}. 
During the update of the policy, neither the \gls{mp} $\rho(\bm{w}, \bm{q})$ nor the controller $f(\bm{q}^d_{t+1}, \bm{q}_t)$ are needed, i.e., our approach would work with any form of parametrizable controller.

\subsection{Choice of the Deep Reinforcement Learning Algorithm}
Most deep \gls{rl} methods can theoretically be used to train policies in the weight space of MPs. 
Yet, training in this space requires learning policies with a higher degree of precision compared to the step-based case as the selected action is active for more time steps with no opportunity for correction during the trajectory segment.
To address this issue, we chose \gls{trpl} as it has been shown to be more stable and accurate than other \gls{rl} methods \citep{Otto2021}. \gls{trpl} implements exact trust regions for policy updates  and enforces them per state, while most other deep \gls{rl} methods \citep{Schulman2015, Schulman2017, Akrour2019} only provide approximate trust region updates that are enforced for the average policy change across all states.

\subsection{Choice of the Planning Horizon}
Our method harnesses the merits of two common \gls{rl} paradigms: \acrfull{srl} and \acrfull{erl}. The agent's behavior can seamlessly switch between the two paradigms according to the planning horizon $k$. 
As discussed previously, there are two special cases when selecting the planning horizon.

\paragraph{Black-Box Setting.}
The first special case arises when the planning horizon is equal to the episode length, that is, $k=T$. 
In this case, the agent generates only one desired trajectory for the entire episode, similar to an open-loop motion planner. 
We refer to this setting as \gls{method-bb} in the following discussion, since it treats reinforcement learning as a black-box optimization problem.
\gls{method-bb} does not assume the existence of a step reward, but instead evaluates the performance of the entire trajectory as a whole. 
The black-box nature facilitates dealing with sparse and non-Markovian rewards, leading to a more intuitive reward design. 
Another advantage is state abstraction. The \gls{method-bb} agent does not need any intermediate state information about the task execution that varies during the execution process (such as joints position/velocity, etc.), but focuses only on the critical information that defines the essence of the tasks (for example goals, obstacles, etc.), this is also referred to as context.
However, \gls{method-bb} presents some challenges.
Although the reduction of the observation dimension to the context space results in a modest impact on sample efficiency, the high cost of a single sample makes it less sample-efficient. 
Moreover, its black-box nature limits its applicability in dynamic environments, where the agent must adapt to environmental changes during execution.

\begin{figure}[!t]
	\begin{subfigure}{0.33\textwidth}
		\resizebox{\textwidth}{!}{
\begin{tikzpicture}
	
	\definecolor{darkgray176}{RGB}{176,176,176}
	\definecolor{lightgray204}{RGB}{204,204,204}
	\definecolor{slategray}{RGB}{112,128,144}
	\definecolor{tomato}{RGB}{234,67,53}
	\definecolor{royalblue}{RGB}{66,133,244}
	
	\begin{axis}[
		legend cell align={left},
		legend style={
			fill opacity=0.8,
			draw opacity=1,
			text opacity=1,
			at={(0.03,0.97)},
			anchor=north west,
			draw=lightgray204,
            height=0.2 \textwidth,
            width= 0.6\textwidth, 
            ylabel style={font=\Large}, 
		},
		tick align=outside,
		tick pos=left,
		x grid style={darkgray176},
		xmin=-5, xmax=105,
		xtick style={color=black},
		y grid style={darkgray176},
		ymin=-2.57852057512239, ymax=0.293536882958863,
		ytick style={color=black},
		ylabel={\textbf{Random Policy}}
		]
  		\addplot [line width=1.5pt, tomato]
		table {%
			0 -2.39914377237064
			1 -2.40530077939937
			2 -2.41366033897826
			3 -2.4240810312049
			4 -2.43901906621835
			5 -2.44797250884597
			6 -2.43892618718572
			7 -2.41589563650273
			8 -2.40312285358285
			9 -2.4031197898816
			10 -2.41220695220804
			11 -2.43078196790862
			12 -2.43059622866922
			13 -2.41486780817369
			14 -2.39736285609228
			15 -2.3891209828799
			16 -2.39227451508758
			17 -2.39544427085785
			18 -2.38900503701122
			19 -2.37427249363738
			20 -2.3566682183023
			21 -2.32780813544502
			22 -2.29526055892624
			23 -2.26853984399901
			24 -2.23703693645571
			25 -2.20034705091687
			26 -2.16494491905673
			27 -2.13797087055732
			28 -2.12434493092546
			29 -2.11754906773899
			30 -2.10046341223512
			31 -2.07328922359129
			32 -2.04177608319093
			33 -2.00134499464535
			34 -1.96041950723913
			35 -1.9234730983178
			36 -1.88346438596893
			37 -1.84212897028308
			38 -1.8036745984119
			39 -1.75866838335513
			40 -1.71204342825137
			41 -1.66417923983155
			42 -1.62069260027512
			43 -1.58158066708018
			44 -1.53985788541489
			45 -1.50417533602902
			46 -1.4800386369818
			47 -1.46380747976106
			48 -1.45403640145694
			49 -1.44002914654503
			50 -1.41085734721164
			51 -1.36483189675225
			52 -1.3105146223556
			53 -1.26567527338734
			54 -1.25175906250526
			55 -1.21202880559526
			56 -1.15307095403693
			57 -1.09422759275603
			58 -1.04129047615222
			59 -0.974581543045374
			60 -0.90877681662497
			61 -0.871946671077192
			62 -0.814926046684896
			63 -0.703856151512773
			64 -0.550789623144236
			65 -0.359747787669203
			66 -0.166521820561481
			67 0.0262333533011815
			68 0.162988816682443
			69 0.156754187323396
			70 0.128767295140383
			71 0.0982385372886511
			72 0.065924414477477
			73 0.0400852909263791
			74 0.0116804328990385
			75 -0.0235996382266243
			76 -0.0419665197745115
			77 -0.0334036977422563
			78 -0.0147990801768695
			79 -0.00922107414984927
			80 -0.00722797543770718
			81 -0.0154819399021489
			82 -0.021090989741567
			83 -0.0292122631718428
			84 -0.0633281292412497
			85 -0.0273793885053285
			86 0.042524145640539
			87 0.054229384183953
			88 0.0509589451195481
			89 0.0554220101047407
			90 0.0714094998611918
			91 0.0854723314621558
			92 0.0788506945838068
			93 0.0469299057326481
			94 0.00966246763640301
			95 -0.0222648334351823
			96 -0.053500058429833
			97 -0.0793910849925127
			98 -0.0984008534174017
			99 -0.112394082120461
			100 -0.0946608231112874
		};
  
		\addplot [line width=1.5pt, royalblue]
		table {%
			0 -2.39914377237064
			1 -2.39466691017151
			2 -2.38193321228027
			3 -2.3619396686554
			4 -2.33560180664062
			5 -2.30376243591309
			6 -2.26719999313354
			7 -2.22663760185242
			8 -2.18275737762451
			9 -2.13621187210083
			10 -2.08764266967773
			11 -2.03769779205322
			12 -1.98705005645752
			13 -1.93640649318695
			14 -1.88651621341705
			15 -1.83816230297089
			16 -1.79213869571686
			17 -1.74921989440918
			18 -1.71012043952942
			19 -1.67545652389526
			20 -1.64571821689606
			21 -1.62125217914581
			22 -1.60225760936737
			23 -1.58879566192627
			24 -1.58080399036407
			25 -1.57811689376831
			26 -1.57875835895538
			27 -1.58091425895691
			28 -1.58455717563629
			29 -1.58960866928101
			30 -1.59595561027527
			31 -1.60345780849457
			32 -1.61196088790894
			33 -1.62130188941956
			34 -1.6313111782074
			35 -1.6418240070343
			36 -1.65267789363861
			37 -1.66371631622314
			38 -1.67479038238525
			39 -1.68576073646545
			40 -1.69649267196655
			41 -1.70686316490173
			42 -1.71675038337708
			43 -1.72603678703308
			44 -1.73460340499878
			45 -1.74232769012451
			46 -1.74907958507538
			47 -1.75471973419189
			48 -1.75909829139709
			49 -1.76206183433533
			50 -1.76345252990723
			51 -1.76181423664093
			52 -1.75623106956482
			53 -1.74725651741028
			54 -1.73536789417267
			55 -1.72096931934357
			56 -1.70439171791077
			57 -1.6858948469162
			58 -1.66567718982697
			59 -1.64388251304626
			60 -1.62060523033142
			61 -1.59590005874634
			62 -1.56979024410248
			63 -1.54227113723755
			64 -1.51331925392151
			65 -1.48289310932159
			66 -1.45094668865204
			67 -1.41742384433746
			68 -1.3822705745697
			69 -1.34543478488922
			70 -1.30686974525452
			71 -1.26653850078583
			72 -1.22441613674164
			73 -1.18048989772797
			74 -1.13476181030273
			75 -1.08725214004517
			76 -1.04451596736908
			77 -1.01264584064484
			78 -0.990880012512207
			79 -0.978481590747833
			80 -0.974736630916595
			81 -0.978960812091827
			82 -0.990488231182098
			83 -1.00867760181427
			84 -1.03291141986847
			85 -1.06259608268738
			86 -1.09715986251831
			87 -1.13605546951294
			88 -1.17876374721527
			89 -1.22478592395782
			90 -1.27365207672119
			91 -1.32491517066956
			92 -1.37815070152283
			93 -1.43296563625336
			94 -1.48898696899414
			95 -1.54586887359619
			96 -1.60328722000122
			97 -1.6609423160553
			98 -1.71855890750885
			99 -1.77588212490082
			100 -1.83268082141876
		};

		\addplot[const plot, dashed, line width=1.5pt, slategray] coordinates {
			(24, -3)
			(24, 1)
		};
		
		\addplot[const plot, dashed, line width=1.5pt, slategray] coordinates {
			(49, -3)
			(49, 1)
		};
		\addplot[const plot, dashed, line width=1.5pt, slategray] coordinates {
			(74, -3)
			(74, 1)
		};
	\end{axis}
	
\end{tikzpicture}}
	\end{subfigure}
	\begin{subfigure}{0.3\textwidth}
		\resizebox{\textwidth}{!}{
\begin{tikzpicture}
	
	\definecolor{darkgray176}{RGB}{176,176,176}
	\definecolor{lightgray204}{RGB}{204,204,204}
	\definecolor{slategray}{RGB}{112,128,144}
	\definecolor{tomato}{RGB}{234,67,53}
	\definecolor{royalblue}{RGB}{66,133,244}
	\begin{axis}[
		legend cell align={left},
		legend style={fill opacity=0.8, draw opacity=1, text opacity=1, draw=lightgray204},
		tick align=outside,
		tick pos=left,
		x grid style={darkgray176},
		xmin=-5, xmax=105,
		xtick style={color=black},
		y grid style={darkgray176},
		ymin=-3.51394534470727, ymax=10.3341768543312,
		ytick style={color=black},
		]
	
		\addplot [line width=1.5pt, tomato]
		table {%
			0 0
			1 -0.560974041894714
			2 -0.299258070539498
			3 -0.70521450543591
			4 -0.773187736253609
			5 -0.154405191502374
			6 0.983781647998043
			7 1.11786655326217
			8 0.214386764723798
			9 -0.20061681569843
			10 -0.64833236212076
			11 -0.823253041758608
			12 0.665607158076472
			13 0.88398439372479
			14 0.86589143656103
			15 0.039691233155374
			16 -0.311073885992648
			17 -0.0276997065035702
			18 0.624593915758348
			19 0.830788260241645
			20 0.924648167632613
			21 1.86731841047227
			22 1.43048918000794
			23 1.26098965293265
			24 1.83204898829329
			25 1.84333249959531
			26 1.72388151831602
			27 1.04745853079941
			28 0.401146548713281
			29 0.330068708970805
			30 1.30197861801473
			31 1.41193811428181
			32 1.71428840765723
			33 2.2745110377322
			34 1.86786691958736
			35 1.83761869659761
			36 2.15052639123558
			37 1.99880041923064
			38 1.88215934502729
			39 2.56865228918024
			40 2.13638412844367
			41 2.62148028962151
			42 1.82067459152604
			43 2.02943452313323
			44 2.08783539347264
			45 1.46889841340303
			46 0.973794512590542
			47 0.52387515562871
			48 0.705053634724725
			49 0.773260411227082
			50 1.7633542628312
			51 2.73358281479835
			52 2.69531278515859
			53 1.51919077076869
			54 1.07773951798208
			55 2.47098435977153
			56 3.35522625513571
			57 2.51718222006382
			58 2.63969051847485
			59 3.85710567588095
			60 2.18679117910886
			61 1.71654022043975
			62 4.7858854385887
			63 5.61578337004391
			64 9.50949622396851
			65 9.70471675437488
			66 9.60331004419985
			67 9.61945335015958
			68 2.15053501787855
			69 -1.33215896031512
			70 -1.2752956160622
			71 -1.80358503356867
			72 -1.3814914089618
			73 -1.3942403297861
			74 -1.62627294304539
			75 -1.77149266445103
			76 -0.172516183200528
			77 0.869896057000265
			78 0.62177837783273
			79 -0.0181625365707488
			80 0.109883342006574
			81 -0.976430228967635
			82 0.272504361998486
			83 -0.92933842674926
			84 -2.01389582730734
			85 5.17902379141836
			86 1.93802381332819
			87 -0.328589096813013
			88 0.0313248020868777
			89 0.364774779653901
			90 1.16297371120197
			91 0.45680756474573
			92 -1.08333274741609
			93 -1.94353767128658
			94 -1.77932919659829
			95 -1.44872922507607
			96 -1.60339123154179
			97 -0.880076091693173
			98 -1.03603888593
			99 -0.396951259376272
			100 1.76920988068459
		};
		\addplot [line width=1.5pt, royalblue]
		table {%
			0 0
			1 0.439183354377747
			2 0.826389074325562
			3 1.16584038734436
			4 1.46138572692871
			5 1.716472864151
			6 1.93408870697021
			7 2.11666774749756
			8 2.2659854888916
			9 2.38303637504578
			10 2.46794176101685
			11 2.51993012428284
			12 2.53745985031128
			13 2.51852965354919
			14 2.46119999885559
			15 2.36425089836121
			16 2.22784328460693
			17 2.0539858341217
			18 1.8466557264328
			19 1.61154198646545
			20 1.35549533367157
			21 1.08585572242737
			22 0.809823870658875
			23 0.533981800079346
			24 0.263999700546265
			25 0.00450801849365234
			26 -0.0707821100950241
			27 -0.146193891763687
			28 -0.218849539756775
			29 -0.286619305610657
			30 -0.347991228103638
			31 -0.401954591274261
			32 -0.447899639606476
			33 -0.485530734062195
			34 -0.514795541763306
			35 -0.535821080207825
			36 -0.548869967460632
			37 -0.554290294647217
			38 -0.55248236656189
			39 -0.543851852416992
			40 -0.528781056404114
			41 -0.5075763463974
			42 -0.480431199073792
			43 -0.447395801544189
			44 -0.408339858055115
			45 -0.362969517707825
			46 -0.310854077339172
			47 -0.251514792442322
			48 -0.1845463514328
			49 -0.109756827354431
			50 -0.0272946357727051
			51 0.184860616922379
			52 0.367710143327713
			53 0.524789333343506
			54 0.659826338291168
			55 0.776556253433228
			56 0.878559172153473
			57 0.969141006469727
			58 1.05125904083252
			59 1.1274870634079
			60 1.20001423358917
			61 1.27066147327423
			62 1.34090387821198
			63 1.41190326213837
			64 1.48452794551849
			65 1.55938708782196
			66 1.6368488073349
			67 1.71706175804138
			68 1.79997658729553
			69 1.88536846637726
			70 1.97285652160645
			71 2.06192398071289
			72 2.15194511413574
			73 2.24220538139343
			74 2.33193016052246
			75 2.42030954360962
			76 1.857497215271
			77 1.3335063457489
			78 0.846984505653381
			79 0.396652102470398
			80 -0.0186871290206909
			81 -0.40023148059845
			82 -0.749135851860046
			83 -1.06656181812286
			84 -1.35368406772614
			85 -1.61168241500854
			86 -1.84175801277161
			87 -2.04513788223267
			88 -2.22305083274841
			89 -2.37674450874329
			90 -2.50746965408325
			91 -2.61649799346924
			92 -2.70508432388306
			93 -2.77447581291199
			94 -2.82590389251709
			95 -2.8605945110321
			96 -2.87973213195801
			97 -2.88448524475098
			98 -2.87599229812622
			99 -2.85533785820007
			100 -2.82359099388123
		};

	\addplot[const plot, dashed, line width=1.5pt, slategray] coordinates {
		(49, -3.5)
		(49, 10)
	};
	\addplot[const plot, dashed, line width=1.5pt, slategray] coordinates {
		(74, -3.5)
		(74, 10)
	};
	\addplot[const plot, dashed, line width=1.55pt, slategray] coordinates {
		(24, -3.5)
		(24, 10)
	};
	\end{axis}
	
\end{tikzpicture}}
	\end{subfigure}
	\begin{subfigure}{0.315\textwidth}
		\resizebox{\textwidth}{!}{
\begin{tikzpicture}
	
	\definecolor{darkgray176}{RGB}{176,176,176}
	\definecolor{lightgray204}{RGB}{204,204,204}
	\definecolor{slategray}{RGB}{112,128,144}
	\definecolor{tomato}{RGB}{234,67,53}
	\definecolor{royalblue}{RGB}{66,133,244}
	
	\begin{axis}[
		legend cell align={left},
		legend style={
			fill opacity=0.8,
			draw opacity=1,
			text opacity=1,
			at={(0.03,0.97)},
			anchor=north west,
			draw=lightgray204
		},
		tick align=outside,
		tick pos=left,
		x grid style={darkgray176},
		xmin=-4.95, xmax=103.95,
		xtick style={color=black},
		y grid style={darkgray176},
		ymin=-1.02990669906139, ymax=1.05558455884457,
		ytick style={color=black},
		]
  		\addplot [line width=1.5pt, tomato]
		table {%
			0 -0.54958564043045
			1 -0.393935143947601
			2 0.585323810577393
			3 -0.928639471530914
			4 -0.504970133304596
			5 0.660095453262329
			6 0.483541905879974
			7 -0.881496727466583
			8 0.179983392357826
			9 -0.195003405213356
			10 0.324429482221603
			11 -0.615073025226593
			12 -0.247424334287643
			13 -0.355673372745514
			14 0.315521448850632
			15 -0.289172440767288
			16 0.50595498085022
			17 0.586047291755676
			18 -0.307877123355865
			19 0.960789501667023
			20 0.756375730037689
			21 -0.498001515865326
			22 -0.203448876738548
			23 0.207558050751686
			24 0.256201714277267
			25 0.459978491067886
			26 -0.906695544719696
			27 -0.922028303146362
			28 0.0767927616834641
			29 0.603328824043274
			30 0.25322362780571
			31 0.76524144411087
			32 0.263909459114075
			33 -0.67697536945343
			34 -0.494450896978378
			35 0.661105036735535
			36 -0.0851486027240753
			37 -0.73784202337265
			38 0.665588617324829
			39 -0.806384563446045
			40 0.0375524684786797
			41 -0.748516976833344
			42 0.686050117015839
			43 0.442590415477753
			44 -0.22496272623539
			45 -0.724367678165436
			46 -0.753237187862396
			47 0.500784397125244
			48 -0.93488335609436
			49 0.777398586273193
			50 0.00667289597913623
			51 -0.468270242214203
			52 0.581627607345581
			53 -0.538665652275085
			54 -0.369298368692398
			55 0.657960951328278
			56 -0.664718568325043
			57 -0.689739465713501
			58 0.603593230247498
			59 0.670110583305359
			60 0.455684125423431
			61 0.461697697639465
			62 0.262145817279816
			63 -0.604315757751465
			64 0.240770861506462
			65 0.488584518432617
			66 -0.23807480931282
			67 0.720667362213135
			68 -0.633032619953156
			69 -0.256744861602783
			70 -0.0123333223164082
			71 -0.159514755010605
			72 0.0917779505252838
			73 0.63485985994339
			74 0.274327039718628
			75 -0.42172384262085
			76 -0.608855068683624
			77 0.24181355535984
			78 -0.716554462909698
			79 0.923418521881104
			80 -0.674155116081238
			81 0.63369083404541
			82 0.200695365667343
			83 -0.93511164188385
			84 0.140296876430511
			85 0.387842148542404
			86 -0.631455540657043
			87 0.706985831260681
			88 0.372657299041748
			89 0.630071759223938
			90 -0.451486855745316
			91 -0.414120197296143
			92 -0.225022211670876
			93 -0.061429213732481
			94 -0.457787543535233
			95 -0.454030603170395
			96 0.60921984910965
			97 -0.261990487575531
			98 0.937117278575897
			99 0.907151222229004
		};
		
  
		\addplot [line width=1.5pt, royalblue]
		table {%
			0 0.0492905700767336
			1 0.10003518868402
			2 0.151407852880357
			3 0.202504571190059
			4 0.252616298122262
			5 0.301142875328831
			6 0.347572907143158
			7 0.391442756574948
			8 0.432299314340192
			9 0.476754890977681
			10 0.533013411133828
			11 0.564770365307039
			12 0.586825861485413
			13 0.609968289995296
			14 0.649286650214576
			15 0.661909920521669
			16 0.650840492341305
			17 0.628924410014173
			18 0.596902120205
			19 0.555740912023789
			20 0.506612188384225
			21 0.450839769762358
			22 0.389869672639171
			23 0.325240793594512
			24 0.258570440788389
			25 0.210613588231842
			26 0.166049917924235
			27 0.125807497196451
			28 0.089649581751006
			29 0.0606548856500143
			30 0.0341562403691116
			31 0.0129874399906857
			32 -0.00711306837747651
			33 -0.0253417089181527
			34 -0.0425329434191315
			35 -0.0593463344618573
			36 -0.076378533702994
			37 -0.0941003581255803
			38 -0.1194473630174
			39 -0.164950200132638
			40 -0.179083059046249
			41 -0.193939347855653
			42 -0.20774563924319
			43 -0.221020797821512
			44 -0.231567139253131
			45 -0.238863899649088
			46 -0.241614256963635
			47 -0.239902288013211
			48 -0.231699150255907
			49 -0.2165122569505
			50 -0.18036732105555
			51 -0.14140786076449
			52 -0.0955567622357635
			53 -0.0545196002832777
			54 0.0240942339578087
			55 0.0662274220488699
			56 0.111426026534846
			57 0.159612006598787
			58 0.209325809180964
			59 0.263635699656585
			60 0.330181852275145
			61 0.370649413760567
			62 0.413221750756923
			63 0.46691909130617
			64 0.503409610779223
			65 0.544549540199929
			66 0.576533962523299
			67 0.605949574651796
			68 0.630797161367544
			69 0.652173236742573
			70 0.669889124398316
			71 0.683774867032976
			72 0.708345272529546
			73 0.717613160390522
			74 0.721334462689012
			75 0.647459986791836
			76 0.563168661117941
			77 0.47123621849106
			78 0.377961132729209
			79 0.271543757129383
			80 0.181336681928968
			81 0.0944226101237058
			82 -0.00279364164107671
			83 -0.075169245243926
			84 -0.13704040591848
			85 -0.186547230973681
			86 -0.24531615465763
			87 -0.279343106939115
			88 -0.319533034147961
			89 -0.350137162740177
			90 -0.385379511508051
			91 -0.423750442451019
			92 -0.467023589440492
			93 -0.513812645245775
			94 -0.564674383381722
			95 -0.620624224791732
			96 -0.677980128850342
			97 -0.744640436142357
			98 -0.815575955574559
			99 -0.886412938453453
		};

		\addplot[const plot, dashed, line width=1.5pt, slategray] coordinates {
			(24, -1.5)
			(24, 1.5)
		};
		
		\addplot[const plot, dashed, line width=1.5pt, slategray] coordinates {
			(49, -1.5)
			(49, 1.5)
		};
		\addplot[const plot, dashed, line width=1.5pt, slategray] coordinates {
			(74, -1.5)
			(74, 1.5)
		};
	\end{axis}
	
\end{tikzpicture}}
    \end{subfigure}
  \\
  	\begin{subfigure}{0.33\textwidth}
		\resizebox{\textwidth}{!}{
\begin{tikzpicture}
	
	\definecolor{darkgray176}{RGB}{176,176,176}
	\definecolor{lightgray204}{RGB}{204,204,204}
	\definecolor{slategray}{RGB}{112,128,144}
	\definecolor{tomato}{RGB}{234,67,53}
	\definecolor{royalblue}{RGB}{66,133,244}
	
	\begin{axis}[
		legend cell align={left},
		legend style={
			fill opacity=0.8,
			draw opacity=1,
			text opacity=1,
			at={(0.03,0.97)},
			anchor=north west,
			draw=lightgray204,
            ylabel style={font=\Large}, 
		},
		tick align=outside,
		tick pos=left,
		x grid style={darkgray176},
		xmin=-5, xmax=105,
		xtick style={color=black},
		y grid style={darkgray176},
		ymin=-3.2, ymax=-0.714,
		ytick style={color=black},
		xlabel={Environment Steps},
		ylabel={\textbf{Trained Policy}}
		]
  		\addplot [line width=1.5pt, tomato]
		table {%
0 -2.39859659789173
1 -2.40141167004485
2 -2.41015413421354
3 -2.42142607964673
4 -2.42839516381388
5 -2.43645931148195
6 -2.44706290866536
7 -2.46111662926664
8 -2.47909929005742
9 -2.49932595167948
10 -2.52233640095211
11 -2.54617470041011
12 -2.56861140822461
13 -2.59074639297812
14 -2.61170072774335
15 -2.63188849714292
16 -2.65080998219766
17 -2.66771701393044
18 -2.68372317531326
19 -2.69859159521355
20 -2.71206104639465
21 -2.7240039918081
22 -2.73464192144822
23 -2.74412506622366
24 -2.75276897721009
25 -2.76080775347932
26 -2.76825917650607
27 -2.77519324714855
28 -2.78161862499441
29 -2.78751100763058
30 -2.79282521504571
31 -2.79751588911468
32 -2.8015191978699
33 -2.80481944438633
34 -2.80743788532325
35 -2.80942869374404
36 -2.81087426226844
37 -2.81186149079341
38 -2.81248983868578
39 -2.81286007462086
40 -2.81307176721573
41 -2.81321768625663
42 -2.81338209674364
43 -2.81363937882087
44 -2.81405429751094
45 -2.81474019929194
46 -2.81752033479653
47 -2.81987389084832
48 -2.82422104588398
49 -2.82861672176649
50 -2.83369303144135
51 -2.83980218609173
52 -2.84685122736883
53 -2.85482673701193
54 -2.86316408497477
55 -2.87495011238051
56 -2.88853388576371
57 -2.90173502964618
58 -2.91554345335262
59 -2.93054866081686
60 -2.94652635623642
61 -2.96282182932611
62 -2.97901521583905
63 -2.99507703769356
64 -3.01112343498504
65 -3.02720242683296
66 -3.04330419810559
67 -3.05946466542107
68 -3.07564802207521
69 -3.0900638026408
70 -3.09369098578983
71 -3.09236575938137
72 -3.09010975963477
73 -3.08877661448225
74 -3.08834780010137
75 -3.0881283662136
76 -3.08804021281567
77 -3.08802666963041
78 -3.08806557353363
79 -3.08813078161502
80 -3.08820518096205
81 -3.08828183763666
82 -3.08835895750723
83 -3.08843679579761
84 -3.08851622836883
85 -3.08859817196632
86 -3.08868338525843
87 -3.08877242757433
88 -3.08886567519425
89 -3.0889633553959
90 -3.08906557965883
91 -3.08917237250258
92 -3.08928369709767
93 -3.08939947122937
94 -3.08951958031041
95 -3.08964388736906
96 -3.08977224166552
97 -3.0899044836235
98 -3.09004044848525
		};
  
		\addplot [line width=1.5pt, royalblue]
		table {%
0 -2.39898460838239
1 -2.39839892010853
2 -2.39715511255023
3 -2.39505435183554
4 -2.39193030622164
5 -2.3876475441445
6 -2.3820996845451
7 -2.37520736512342
8 -2.36691618703083
9 -2.35719470841361
10 -2.34611454749537
11 -2.33508401642262
12 -2.3232488958011
13 -2.30951277204682
14 -2.29364135362319
15 -2.27682429787024
16 -2.25854675143935
17 -2.23881031436297
18 -2.21784284866217
19 -2.19565417657761
20 -2.17233400247876
21 -2.14821896905276
22 -2.12324793407687
23 -2.0975412890929
24 -2.07140406376969
25 -2.04469772036114
26 -2.01800752514858
27 -1.99095243378544
28 -1.96332806310053
29 -1.93493135192112
30 -1.9060714490368
31 -1.87704510220599
32 -1.84812805200141
33 -1.81956797396343
34 -1.79157899122227
35 -1.76433778756638
36 -1.7379813312452
37 -1.71260617202747
38 -1.68826923862531
39 -1.66498998998616
40 -1.64275377310267
41 -1.62151620283396
42 -1.60120841707946
43 -1.5817430180858
44 -1.56302051016226
45 -1.54493602430865
46 -1.52738601916364
47 -1.51027461680997
48 -1.49351918361727
49 -1.47705472524282
50 -1.46083497702314
51 -1.44483725172552
52 -1.42906573249361
53 -1.41354763179895
54 -1.39832867371366
55 -1.38346782173321
56 -1.3690314793394
57 -1.35508751871039
58 -1.34169949289068
59 -1.32892134209083
60 -1.31679286568623
61 -1.30533617636831
62 -1.29455326958713
63 -1.2844247637027
64 -1.27490983234271
65 -1.26594727134212
66 -1.25745751934922
67 -1.24934548903523
68 -1.24150397193475
69 -1.23381740267696
70 -1.22616576291682
71 -1.21842842565606
72 -1.21048776497305
73 -1.20223242457476
74 -1.19356013889038
75 -1.18432056654272
76 -1.17434788258051
77 -1.16352784624865
78 -1.15179590111379
79 -1.1391330999124
80 -1.12556088483412
81 -1.11113492902238
82 -1.0959381872149
83 -1.08007406366743
84 -1.06365861440127
85 -1.04681452647111
86 -1.02966303213912
87 -1.01232031503132
88 -0.994889806870124
89 -0.977461870668825
90 -0.960107875728988
91 -0.942882035787223
92 -0.925819520803663
93 -0.908937994589352
94 -0.892240615551172
95 -0.875716763692438
96 -0.859347196545998
97 -0.843105513104602
98 -0.826962381490609
		};

		\addplot[const plot, dashed, line width=1.5pt, slategray] coordinates {
			(24, -3.5)
			(24, 1)
		};
		
		\addplot[const plot, dashed, line width=1.5pt, slategray] coordinates {
			(49, -3.5)
			(49, 1)
		};
		\addplot[const plot, dashed, line width=1.5pt, slategray] coordinates {
			(74, -3.5)
			(74, 1)
		};
	\end{axis}
	
\end{tikzpicture}}
	  \caption[]{Position ($\text{rad}$)}
    \end{subfigure}
	\begin{subfigure}{0.305\textwidth}
		\resizebox{\textwidth}{!}{
\begin{tikzpicture}
	
	\definecolor{darkgray176}{RGB}{176,176,176}
	\definecolor{lightgray204}{RGB}{204,204,204}
	\definecolor{slategray}{RGB}{112,128,144}
	\definecolor{tomato}{RGB}{234,67,53}
	\definecolor{royalblue}{RGB}{66,133,244}
	\begin{axis}[
		legend cell align={left},
		legend style={fill opacity=0.8, draw opacity=1, text opacity=1, draw=lightgray204},
		tick align=outside,
		tick pos=left,
		x grid style={darkgray176},
		xmin=-5, xmax=105,
		xtick style={color=black},
		y grid style={darkgray176},
		ymin=-1.351394534470727, ymax=1.5941768543312,
		ytick style={color=black},
		xlabel={Environment Steps},
		]
	
		\addplot [line width=1.5pt, tomato]
		table {%
0 0
1 0.0482465294565408
2 -0.295792109812867
3 -0.590043355784424
4 -0.532121593414087
5 -0.198871003963167
6 -0.567541951144376
7 -0.49969086603097
8 -0.865404934041102
9 -0.927227240970709
10 -1.07947284558029
11 -1.20878800639576
12 -1.17863532961843
13 -1.07527031688896
14 -1.13021878558028
15 -0.978516596440742
16 -1.03790163782234
17 -0.874007201289607
18 -0.824590031344774
19 -0.782107868671556
20 -0.713340482069232
21 -0.642646878145877
22 -0.56171875033673
23 -0.508615979078939
24 -0.447296077096693
25 -0.420415420381308
26 -0.387298773587932
27 -0.360603407466768
28 -0.335061198354243
29 -0.309355378026093
30 -0.281581810706949
31 -0.251428095267748
32 -0.219052090410187
33 -0.182628084411458
34 -0.148192565581661
35 -0.113987657122636
36 -0.0846246189699583
37 -0.0587552924026212
38 -0.0379823446391161
39 -0.022093820885993
40 -0.0114036033148228
41 -0.00553782682752608
42 -0.00419013918891997
43 -0.00683783102616553
44 -0.0130127740365311
45 -0.0222242787444614
46 -0.0628510972320342
47 -0.149832484398374
48 -0.113015239379351
49 -0.249632578477163
50 -0.189618165402709
51 -0.29756794697842
52 -0.330400582905834
53 -0.360008884525412
54 -0.402292163800092
55 -0.446603304507548
56 -0.707046407855175
57 -0.657626796034778
58 -0.661578155127144
59 -0.71329985311515
60 -0.779323063997124
61 -0.813337617417329
62 -0.814188710436772
63 -0.804450645415814
64 -0.80082112455702
65 -0.80242345087758
66 -0.803922130304703
67 -0.804900528767903
68 -0.809686711080942
69 -0.807863410257411
70 -0.57790061980603
71 0.0281174761589923
72 0.106398004920899
73 0.0975252236415018
74 0.0373453784976333
75 0.0143633555256776
76 0.00615042803386034
77 0.00234087083499735
78 -0.00107737802139612
79 -0.00307253874453027
80 -0.00383987449392043
81 -0.0040533482672357
82 -0.00408833523854848
83 -0.00410480871736369
84 -0.00415844513575035
85 -0.00426046233906329
86 -0.00440567761781374
87 -0.00458415167192801
88 -0.00478602229506033
89 -0.00500280140302319
90 -0.00522794380212781
91 -0.0054563153108609
92 -0.0056842359322785
93 -0.00590915625165364
94 -0.00612904758349194
95 -0.00634263435654611
96 -0.00654892923833833
97 -0.00674749099937268
98 -0.00693787893614975
99 -0.00711998329669568
		};
		\addplot [line width=1.5pt, royalblue]
		table {%
0 0
1 0.0144628635096039
2 0.0413803493701849
3 0.079152468928088
4 0.126112255811278
5 0.180669912266275
6 0.241318486660619
7 0.306651447499917
8 0.375364939072111
9 0.446258062967061
10 0.518228123980138
11 0.563216766953452
12 0.555028834936016
13 0.625372169936343
14 0.737086662271182
15 0.830048417169667
16 0.865935326756342
17 0.938022843143499
18 1.00892435086609
19 1.06967876783057
20 1.12228295814855
21 1.17824205613778
22 1.21709170989602
23 1.26530424603301
24 1.31130625415385
25 1.32172842007808
26 1.32865912630085
27 1.34636351868266
28 1.35557459552342
29 1.40272808701352
30 1.4342125828577
31 1.45040530847476
32 1.45214394323795
33 1.44065945264959
34 1.41750029508814
35 1.38445168103021
36 1.34345623071633
37 1.29653273027731
38 1.24570005891672
39 1.19290359067666
40 1.13995295099213
41 1.08846379543038
42 1.03981300214259
43 0.995096026383748
44 0.955099564832266
45 0.920278715588551
46 0.890749389061262
47 0.866296350807256
48 0.846399479088296
49 0.830280603706324
50 0.816972810228777
51 0.805577883859779
52 0.794660097606133
53 0.782955750021888
54 0.769405681943456
55 0.75319427637009
56 0.73378700970646
57 0.710947809823597
58 0.684740303415228
59 0.655509724998793
60 0.623852329304637
61 0.59057031129195
62 0.55661900153683
63 0.523048416090888
64 0.490944568790878
65 0.461369182956739
66 0.435306103457521
67 0.413618790488235
68 0.397011701110077
69 0.386010061232963
70 0.380943911015406
71 0.381948565835827
72 0.388969616936438
73 0.401780068015045
74 0.419997143820503
75 0.443109022386674
76 0.475964917214697
77 0.515943995291884
78 0.560494553555939
79 0.607187460830305
80 0.653856890828521
81 0.69860436887328
82 0.739861479915785
83 0.776437070808796
84 0.807432044203017
85 0.832446339230008
86 0.851180398368439
87 0.864039792670507
88 0.871112507474652
89 0.87353090073507
90 0.871501824912097
91 0.866639372858269
92 0.859124561036524
93 0.850591384633138
94 0.841252436846875
95 0.832283495034796
96 0.823957617283127
97 0.816846836677205
98 0.811131523132794
99 0.806934017596924
		};

	\addplot[const plot, dashed, line width=1.5pt, slategray] coordinates {
		(49, -3)
		(49, 10)
	};
	\addplot[const plot, dashed, line width=1.5pt, slategray] coordinates {
		(74, -3)
		(74, 10)
	};
	\addplot[const plot, dashed, line width=1.55pt, slategray] coordinates {
		(24, -3)
		(24, 10)
	};
	\end{axis}
	
\end{tikzpicture}}
  		\caption[]{Velocity ($\text{rad}/\text{s}$)}
	\end{subfigure}
	\begin{subfigure}{0.31\textwidth}
		\resizebox{\textwidth}{!}{
\begin{tikzpicture}
	
	\definecolor{darkgray176}{RGB}{176,176,176}
	\definecolor{lightgray204}{RGB}{204,204,204}
	\definecolor{slategray}{RGB}{112,128,144}
	\definecolor{tomato}{RGB}{234,67,53}
	\definecolor{royalblue}{RGB}{66,133,244}
	
	\begin{axis}[
		legend cell align={left},
		legend style={
			fill opacity=0.8,
			draw opacity=1,
			text opacity=1,
			at={(0.03,0.97)},
			anchor=north west,
			draw=lightgray204
		},
		tick align=outside,
		tick pos=left,
		x grid style={darkgray176},
		xmin=-4.95, xmax=103.95,
		xtick style={color=black},
		y grid style={darkgray176},
		ymin=-1.02990669906139, ymax=1.05558455884457,
		ytick style={color=black},
		xlabel={Environment Steps},
		]
  		\addplot [line width=1.5pt, tomato]
		table {%
                0 -1
                1 -1
                2 -1
                3 0.154752776026726
                4 0.326385676860809
                5 -0.776353478431702
                6 -0.0562855750322342
                7 -0.588217198848724
                8 0.113217487931252
                9 0.227626875042915
                10 0.715880513191223
                11 1
                12 0.789272487163544
                13 0.297317862510681
                14 0.384492099285126
                15 0.220683887600899
                16 0.437604427337646
                17 0.410150706768036
                18 0.319280624389648
                19 0.247285410761833
                20 0.334481596946716
                21 0.289196670055389
                22 0.236333951354027
                23 0.154150769114494
                24 0.0881581753492355
                25 0.0603598207235336
                26 0.0334096848964691
                27 0.0198009759187698
                28 0.0139187127351761
                29 0.0197566896677017
                30 0.0297990590333939
                31 0.0429670065641403
                32 0.0633415132761002
                33 0.0634991973638535
                34 0.0676978975534439
                35 0.0553949028253555
                36 0.0462214797735214
                37 0.0316179096698761
                38 0.0180700272321701
                39 0.00434798002243042
                40 -0.00753825902938843
                41 -0.017676904797554
                42 -0.0255670100450516
                43 -0.0313551723957062
                44 -0.0350703448057175
                45 -0.0370051860809326
                46 -0.0151099860668182
                47 -0.0438652858138084
                48 -0.135712817311287
                49 -0.158681690692902
                50 -0.199788689613342
                51 -0.232448592782021
                52 -0.267716526985168
                53 -0.263033121824265
                54 -0.347802996635437
                55 0.137122988700867
                56 0.0549045354127884
                57 0.0529424995183945
                58 -0.113965585827827
                59 -0.119582116603851
                60 -0.0467691272497177
                61 0.0153631716966629
                62 0.0296569019556046
                63 0.00350536406040192
                64 -0.0258810967206955
                65 -0.028642013669014
                66 -0.0187100917100906
                67 -0.0304847806692123
                68 -0.0452723056077957
                69 -0.0178706794977188
                70 -0.13666196167469
                71 -0.951874434947968
                72 -0.151472002267838
                73 -0.0520435273647308
                74 0.00601281225681305
                75 -0.0304150357842445
                76 0.00202635675668716
                77 -0.00348947942256927
                78 -0.00271157175302505
                79 0.000825099647045135
                80 0.00296333432197571
                81 0.004078209400177
                82 0.00468531996011734
                83 0.00505510717630386
                84 0.0053119957447052
                85 0.00551043450832367
                86 0.00567542016506195
                87 0.00581773370504379
                88 0.00594329088926315
                89 0.00605487078428268
                90 0.00615476071834564
                91 0.00624407082796097
                92 0.00632369518280029
                93 0.00639535486698151
                94 0.00645942986011505
                95 0.00651716440916061
                96 0.00656852871179581
                97 0.00661495327949524
                98 0.00665640830993652
                99 0.0066937580704689
		};
		
  
		\addplot [line width=1.5pt, royalblue]
		table {%
			0 0.0131234521234331
            1 0.0265773497879685
            2 0.0400989221477049
            3 0.053433717709628
            4 0.0663787409891201
            5 0.0787664336581599
            6 0.0904621933953992
            7 0.10135485960436
            8 0.111358245920885
            9 0.120400289120738
            10 0.128424794388158
            11 0.138187545149713
            12 0.153861072123578
            13 0.161110225820465
            14 0.162201150852541
            15 0.162741919815003
            16 0.16788487704984
            17 0.167599239871521
            18 0.165561216849124
            19 0.162897339905609
            20 0.159392410391328
            21 0.153991126588949
            22 0.149140318634573
            23 0.142120992525367
            24 0.134240716959652
            25 0.131649611095863
            26 0.128653684605106
            27 0.124508369571318
            28 0.120667031615494
            29 0.11222531686364
            30 0.104289415739239
            31 0.097180532188729
            32 0.0911601594604783
            33 0.0864278523296001
            34 0.0831151394822841
            35 0.0812918407164208
            36 0.0809655102389172
            37 0.0820883554952535
            38 0.0845587318706367
            39 0.0882324634963921
            40 0.092923280652736
            41 0.0984158183919524
            42 0.104463193731235
            43 0.110802309018022
            44 0.117155441941685
            45 0.123244009884824
            46 0.128802095234036
            47 0.133590369875174
            48 0.137412540337476
            49 0.140133251519771
            50 0.140730570121395
            51 0.140059401012234
            52 0.138246606471693
            53 0.13546686130979
            54 0.131928986443044
            55 0.12786094135062
            56 0.123488781043117
            57 0.119024223720924
            58 0.114650750335433
            59 0.110517395409365
            60 0.106730857839377
            61 0.103354754106936
            62 0.100409022474612
            63 0.0978751282287773
            64 0.0956956343772285
            65 0.0937839580454855
            66 0.0920315460394762
            67 0.0903109381536955
            68 0.088488820040133
            69 0.0864269819772678
            70 0.0839930924907576
            71 0.0810631435227747
            72 0.077527286127091
            73 0.0732913636811845
            74 0.0682815911190182
            75 0.0647049148661956
            76 0.0599971906722896
            77 0.0542030106153258
            78 0.0474187659651628
            79 0.0397855238197208
            80 0.0314703497245322
            81 0.0226613976778276
            82 0.013554612717693
            83 0.00433657066385276
            84 -0.00480724850393339
            85 -0.0137294957209863
            86 -0.0222818334571834
            87 -0.0303909126834492
            88 -0.0379551748072003
            89 -0.0449921987748542
            90 -0.0514437325551822
            91 -0.0574111007432402
            92 -0.0628753150007211
            93 -0.067981251819459
            94 -0.0727535429806576
            95 -0.0773292934687971
            96 -0.0817686279328855
            97 -0.086174192383827
            98 -0.0906138605966986
            99 -0.0951531310872036
		};

		\addplot[const plot, dashed, line width=1.5pt, slategray] coordinates {
			(24, -1.5)
			(24, 1.5)
		};
		
		\addplot[const plot, dashed, line width=1.5pt, slategray] coordinates {
			(49, -1.5)
			(49, 1.5)
		};
		\addplot[const plot, dashed, line width=1.5pt, slategray] coordinates {
			(74, -1.5)
			(74, 1.5)
		};
	\end{axis}
	
\end{tikzpicture}}
    	\caption[]{Torque ($\text{N}\text{m}$)}
	\end{subfigure}
    \caption{This figure presents a comparison between step-level exploration (red) and the proposed trajectory-level exploration using \glspl{mp} (blue). 
    We generated rollouts in box-pushing environments where the raw action space is joint torques. For the \gls{mp} setting, we used \gls{pdmp} and performed replanning every 25 steps. 
    The trajectories of the 3rd joint are plotted. 
    In the top row, we randomly sampled rollouts from untrained policies to demonstrate the initial exploration.
    In the bottom row, we evaluated two policies trained with PPO (red) and MP3 (blue) respectively, using 40M environment interactions. 
    The results show that MP3 enables smooth trajectory generation at both exploration and evaluation processes.}
    \label{fig:replan_trajectories}
    \vspace{-10pt}
\end{figure}

\paragraph{Step-Based Setting.}
At the opposite end of the planning horizon spectrum is the case where $k=1$.
Here, the agent only executes the desired trajectory for one step, after which it generates a new plan, repeating this loop throughout the episode similar to the \gls{srl} setting. 
However, using \glspl{mp} as a trajectory generator instead of raw actions has two essential differences. 
Firstly, the use of \glspl{mp} guarantees second-order smoothness (position and velocity), resulting in a more consistent and smooth behavior during exploration and evaluation (see \Figref{fig:replan_trajectories}). 
Secondly, the number of \gls{mp} basis functions is a hyperparameter that allows for scaling the dimension of the action space from the size of raw actions to arbitrary dimensions, enriching the model's design choices but also increasing the complexity.
Although increasing the dimensionality of the action space may seem to increase exploration difficulty, it can still benefit from the smooth trajectories generated by \glspl{mp}.
Nonetheless, we did not observe significant improvements using this setting over the standard \gls{srl} setting, and our discussion only aims to highlight the flexibility of the proposed method.
\paragraph{Re-planning with MPs.}
The more general case falls somewhere between the two extremes of \gls{srl} and \gls{erl}. 
In this approach, the agent generates a new desired trajectory after executing the current trajectory for a predefined number of steps ($1<k<T$). 
This method leverages the strengths of both \gls{srl} and \gls{erl} while addressing some of their shortcomings:

\begin{enumerate}
    \item In \gls{srl}, stochastic raw action selection often results in jerky random walk behavior that does not fully explore the trajectory space of the agent. 
    In contrast, our approach explores the weight space of \glspl{mp}, leveraging the \gls{mp}'s smoothness guarantees for more consistent and effective exploration (see \Figref{fig:replan_trajectories}).
    \item Stochastic raw action selection in \gls{srl} also results in noisy returns (see \Figref{fig:replan_trajectories}), translating to high variance in the policy gradient estimation. 
    The smaller number of exploration steps in the re-planning setting yields less variance in gradient estimation, leading to a more stable policy update.
    \item Trajectory-level exploration encapsulates the temporal abstraction within each trajectory segment, reducing the number of decisions to make for each episode and improving the agent's ability to handle the sparsity in the reward function.
    \item The policies trained by \gls{srl} often struggled to generate smooth trajectories  (see \Figref{fig:replan_trajectories}) due to the lack of continuity between consecutive steps. In contrast, policies trained with our approach are able to generate smooth trajectories in both the exploration and evaluation processes. 
    \item The ERL agent only makes decisions at the beginning of each episode and treats each episode as a black-box, thus can deal with the temporal sparsity and non-Markovian property in rewards. However, the black-box perspective also limits their ability to address observation noises and dynamics in the environment, as a result, ERL agents cannot adapt their plan according to the changes in the environment during execution, which makes them less flexible and robust compared to SRLs. Our approach addresses this shortcoming by incorporating periodic re-planning during online execution.
\end{enumerate}

Many \gls{srl} algorithms use a similar design called \emph{frame-skipping}, which can help with the partial observability of some Atari games \citep{braylan2015frame}. 
However, frame-skipping just repeats the same action for the ``skipped" frames, limiting the trajectory's expressive capacity.
In contrast, planning with \gls{mp} can ``skip''more frames without compromising expressiveness.

\subsection{Adapting to Different Reward Settings}
The proposed algorithm offers a significant advantage in the flexibility to adapt to different reward signal designs. By considering the information available at each step, the rewards can be classified into three distinct categories.

\paragraph{Dense Rewards.} Dense rewards provide task-related feedback signals at each time step. For example, in a reacher task where the objective is to reach a desired point with the end-effector, a common task-related feedback could be the distance between the end-effector and the desired point. Well-shaped dense rewards are crucial for the success of most \gls{srl} algorithms. 
However, designing efficient dense rewards can already be challenging for tasks where evaluating the quality of the action step-wise is difficult, such as in beerpong and table tennis.

\paragraph{Sparse Rewards.} Sparse rewards provide task-related feedback only when specified conditions are satisfied. These conditions can be either temporal-related (for instance, providing reward only at episode end) or task-related metrics (for instance, providing reward only when end-effector is close enough to the desired point). Our approach is less affected by the temporal sparse setting because we use highly temporal-abstracted samples in policy updates. In contrast to dense reward signals, sparse rewards are usually more intuitive to design and more suitable for the task where task completion at a specific time point (for instance, the episode end) is desired. Take the aforementioned reacher task as an example, dense reward based on distance error at every step implicitly encourages the agent to reach the desired point as fast as possible, leading to policies with large acceleration and overshooting. This is undesirable if we only want to reach the target at the end. Temporal sparse rewards address this issue by rewarding the agent based on the final state.  

\paragraph{Non-Markovian Rewards.} Non-Markovian rewards provide task-related feedback without adhering to the Markovian condition, which means the reward signal is not fully determined by the current state-action pair but also incorporates the past states and actions. The non-Markovian property exists widely in \gls{rl} tasks. For instance, in playing table tennis, once the ball is hit, the agent's actions no longer influence the trajectory of the ball. Therefore, the reward is not conditioned solely on the action at that time step but also on the actions preceding the hit. This setting is extremely challenging for \gls{srl} algorithms as their policy updates rely on the Markovian assumption. Our approach with the black-box setting can leverage non-Markovian rewards effectively by treating the entire episode trajectory as a single sample. 

\section{Experimental Results}

For our evaluation, we begin by demonstrating the effectiveness of our method in handling sparse rewards, improving precision, and energy efficiency without replanning, that is, in the black-box setting with $k=T$. 
We investigate challenging control problems that are typically difficult to solve in the standard step-based setting. 
Next, we conduct a large-scale study on all 50 Meta-World tasks \citep{Yu2019} to showcase our competitive performance on various robot manipulation tasks that come with highly shaped dense rewards with and without replanning.
Finally, we evaluate our method with replanning for several tasks with changing goals or high uncertainties and perform a thorough ablation study.
We compare our methods, which we call \acrshort{method} and \gls{method-bb} for the replanning and black-box cases respectively, against several other step-based methods, including \gls{ppo} \citep{Schulman2017}, \gls{trpl} \citep{Otto2021}, \gls{sac} \citep{Haarnoja2018}, and \gls{ndp} \citep{Bahl2020}, as well as a deep \gls{es} \citep{Salimans2017}, the linear adaption method \gls{cmore} with \glspl{promp} \citep{Tangkaratt2017} as well as \gls{method-ppo} and \gls{method-bb-ppo}, which are equivalent to \acrshort{method} and \acrshort{method-bb} but trained with \gls{ppo} instead of \gls{trpl}.
For the \gls{erl} methods (\gls{method-bb} and \gls{method-bb-ppo}), we leverage \glspl{promp} for motion generation and for the replanning versions (\acrshort{method} and \acrshort{method-ppo}), we use \glspl{pdmp}.

It is worth noting that the authors of \gls{ndp} report their performance in terms of the used samples rather than environment interactions (the original work only uses every fifth interaction).
However, we believe that reporting the total number of environment interactions leads to a fairer comparison and also helps to explain the relatively poor performance of \gls{ndp} in our experiments.
For both \gls{method-bb} and \gls{method-bb-ppo}, we provide only the context information $\bm{c}$ instead of leveraging the full state observation $\bm{s}$.
The context information $\bm{c}$ is a subset of the observation space that is randomly initialized after each reset and includes the stochastic elements, such as the goal or object positions. 
Unless otherwise specified, we measure the trajectory segment performance $R_{t:t+k}(\bm{w}, \bm{s})$ as the cumulative trajectory return.
We evaluate our method on $20$ different seeds and compute ten evaluation runs after each iteration.
To report our results, we use the \acrfull{iqm} with a $95\%$ stratified bootstrap confidence interval and performance profiles where feasible \citet{Agarwal2021}. 
For a detailed description of the hyperparameters used in the evaluation, please refer to \Apxref{app:parameters}.

\subsection{Black-Box Reinforcement Learning}
\label{sec:bbrl}
\paragraph{Performance}
\label{sec:effectiveness}
As an introductory task, we extend the reacher from OpenAI gym \citep{Brockman2016} by using five actuated joints and limiting the context space, that is, the location of the goal, to $y \geq 0$.
This results in an increased control complexity but slightly decreased task complexity. For a detailed environment description, please see \Apxref{app:reacher}.
We investigate two types of rewards: a dense reward equivalent to the original reacher, and a sparse reward that provides only the distance to the goal in the last episode time step. We study the sparse reward setting as it is better suited to generate energy-efficient motions. Yet, it is also more difficult to learn. 
While \gls{method-bb} and \gls{method-bb-ppo} can solve the task for both rewards, \gls{ndp} and \gls{es} fail in both cases (\twoFigref{fig:reacher:dense}{fig:reacher:sparse}). 
\gls{ppo} and \gls{trpl} achieve a slightly better asymptotic performance than \gls{method-bb} in the dense setting but are unable to consistently reach the goal for the sparse reward signal. 
Although \gls{sac} achieves a comparable performance to \gls{method-bb} in the dense setting, it cannot leverage the sparse reward (see \Apxref{app:evaluations}).
\gls{cmore} performs reasonably well, however, it is only able to cover part of context space due to its linear adaption strategy.

To demonstrate the learning capabilities of all algorithms in handling sparse rewards within a more complex scenario, we conduct evaluations on a box pushing task. 
The goal is to move a box to a given goal location and orientation using a simulated Franka Emika Panda. The goal location and orientation are randomized at the beginning of each episode. For a detailed environment description, please see \Apxref{app:box_pushing}.
Similar to the reacher task, we consider a dense reward signal and a temporal sparse reward signal. The dense reward is based on the position and orientation error for each time step, while the temporal sparse reward depends only on the distance errors at the last time step.
We observed that the temporal sparse reward used in the original work leads to policies that pass through the target location at episode end. 
To address this issue, we increased the control cost penalty and introduced a joint velocity penalty at the episode end. We re-tuned baselines that exhibited competitive performance in the original setting to accommodate the new reward setup. 

While \acrshort{method} and \acrshort{method-bb} achieve the highest precision and sample efficiency in dense reward setting (\Figref{fig:box_pushing:dense}), all \gls{srl} algorithms, excluding \gls{sac}, yielded acceptable performance at the end. The performance of \gls{sac} was adversely affected due to the penalties associated with constraint violation and control cost.
In the sparse reward setting(\Figref{fig:box_pushing:Tsparse}), all the algorithms experience a certain degree of performance decline, while \gls{srl} algorithms encounter substantial performance degradation, the \gls{mp}-based algorithms are capable of maintaining reasonable performance.


Although dense rewards may perform well in certain tasks (for instance, reacher and box-pushing), there are two main reasons to consider sparse rewards. Firstly, sparse rewards are usually easier to design, as we only need to consider the state at the last time step. Another reason is that dense rewards force the agent to reach the goal as fast as possible which typically yields energy inefficient motions. In contrast, sparse rewards only penalize the goal distance in the final time step, while accounting for energy costs in each time step.

\paragraph{Energy Efficiency}
To illustrate the trade-off between precision and energy efficiency, we analyzed the final behaviors of both reward setups with different action penalty factors in the reward function. 
For each of these factors, we computed the average precision and energy consumption.
Our results, shown in \twoFigref{fig:reacher:energy}{fig:box_pushing:energy}, demonstrate that decreasing the action penalty factor leads to higher task precision for all methods.
However, for the dense reward, a high task precision comes with the cost of high energy penalties, whereas the sparse reward generates behavior of similar precision with one (box pushing) or even two (reacher) orders of magnitude less energy consumption. 
When analyzing the final behaviors, agents trained with the dense reward quickly move to the target and stay there, while the agents with the sparse reward reach the target only slightly before the specified end of the episode, resulting in a much slower, smoother, and more energy efficient motion. 

\begin{figure}[t]
    \centering
    \captionsetup*[subfigure]{margin={20pt,0pt}}
    \begin{subfigure}{.33\textwidth}
        \resizebox{\textwidth}{!}{\input{tikz_plots/reacher/reacher_iqm_sample_efficiency}}
        \caption{\scriptsize 5D Reacher - Dense}
        \label{fig:reacher:dense}
    \end{subfigure}%
    \begin{subfigure}{.33\textwidth}
        \resizebox{\textwidth}{!}{\input{tikz_plots/reacher/reacher_sparse_iqm_sample_efficiency}}%
        \caption{\scriptsize 5D Reacher - Sparse}
        \label{fig:reacher:sparse}
    \end{subfigure}%
    \begin{subfigure}{.325\textwidth}
        \centering
        \resizebox{\textwidth}{!}{
\begin{tikzpicture}

\begin{axis}[
legend cell align={left},
legend style={fill opacity=0.8, draw opacity=1, text opacity=1, draw=lightgray204},
tick align=outside,
tick pos=left,
scaled ticks=false,
xmajorgrids=true,
ymajorgrids=true,
x grid style={darkgray176},
xlabel={Control Cost},
xmin=-0.25, xmax=4.2,
xtick style={color=black},
y tick label style=
y grid style={darkgray176},
ylabel={Goal Distance in [m]},
ymin=-0.00367045595988746, ymax=0.055,
ytick style={color=black},
yticklabel style={
    /pgf/number format/.cd,
        fixed,
        fixed zerofill,
        precision=2,
    /tikz/.cd
    },
name=ax1
]
\addplot [semithick, C0, mark=*, mark size=1, mark options={solid}, only marks]
table {%
0.0761166694301032 0.00623786380625012
0.0411478153793777 0.00619509787901616
0.031088966871415 0.00628147933958715
0.0251961040414871 0.00634073501763791
0.0214513908835172 0.00629758784836845
0.0184212210635465 0.00624427343893159
0.0173518556623966 0.00638589499080119
0.01603448353249 0.00647677607995042
0.0148634360948103 0.00663652402964895
0.0136176084619929 0.00633819598099539
0.0134472266595547 0.00637335626376129
0.0126815693038882 0.0065409688534958
0.0120948011780637 0.00642188957511391
0.0115717561073237 0.00672099153661967
0.01121387752385 0.00674908865800919
0.0107545200762243 0.00677682833080307
0.0104365622360233 0.0067526624030303
0.0101881087252347 0.0066776582073486
0.00995761586453003 0.00670390494515648
0.00982982525372756 0.00684677378427535
0.00864207984755357 0.00676497440959448
0.00812241622408941 0.00705769167792255
0.00744198995442658 0.00696257644232432
0.00690940418346233 0.00677424623385555
0.00669187797962886 0.00712548366863515
0.00640357019395304 0.00690846517266827
0.00620540754974669 0.00672489799913334
0.00592054160218229 0.00677735042878557
0.00581994379123407 0.00698476330228835
0.00565695619773785 0.00673265475110774
0.00558332282324088 0.00681844268259598
0.00543127728503315 0.0069693471752895
0.0052867827797334 0.00697669383096753
0.00525639717410556 0.00700346508879379
0.00516329085387839 0.00673125769100021
0.00512979855953202 0.006958519926207
};
\addplot [semithick, C1, mark=square*, mark size=1, mark options={solid}, only marks]
table {%
2.30136734979368 0.00588652261519048
1.27332182842184 0.00664409919433378
0.807665313558108 0.0126162869495419
0.585508823335014 0.0154168203163644
0.475020124694149 0.0188085550792419
0.41424913279561 0.024038040825114
0.372571646090098 0.025414768803837
0.330546979921109 0.0325025029728306
0.28412789313202 0.0262784864598892
0.275806702286927 0.0288070999612936
0.243942821192354 0.0353430405649638
0.236187284953288 0.0305203359443709
0.241589341581954 0.0399671290529157
0.219505133939767 0.0433266188950769
0.202733665419738 0.0389417708835401
0.202820029203675 0.0530862775640351
0.185785765188443 0.051339273415053
0.16865183737666 0.0443723265927089
0.167585348762573 0.0473135541341738
0.172533656009557 0.063451622913814
0.147160257158381 0.0417540033954945
0.0922615828703012 0.111748807856199
0.0642936777343629 0.175241275254231
};



\addplot [semithick, C4, mark=triangle*, mark size=1, mark options={solid}, only marks]
table {%
1.88809231595219 0.0419138091979824
0.488300004041918 0.0409058616718117
0.148885609873042 0.0424843978263697
0.092004522100548 0.0435045211904908
0.0654392334003469 0.0446367679207029
0.0519977063859921 0.0474636669253055
0.0439879058350099 0.0479194044035264
0.0381092402819536 0.0480859642954231
0.0366574065599025 0.0495564969060759
0.031566668295903 0.0502429973242361
0.0312272459135759 0.049879877632215
0.0273447185363163 0.0512717010595057
0.0269520525092751 0.0514918811941105
0.0250286300347991 0.0526169156793315
0.0240012931824594 0.0569464537598505
0.021574867622267 0.0552146779493538
0.0210599171136494 0.0570260946729313
0.0198102320234284 0.0562238041059758
0.0187341631885884 0.0574198876357938
};


\coordinate (c1) at (axis cs: 0.15,0.005);
\coordinate (c2) at (axis cs:-0.05,0.008);
\draw (c1) rectangle (c2);

\coordinate (insetSW) at (axis cs:\pgfkeysvalueof{/pgfplots/xmax},\pgfkeysvalueof{/pgfplots/ymax}); 
\end{axis}

\begin{axis}[
name=ax2,
at={(insetSW)}, anchor=north east, width=150,
scaled ticks=false,
axis background/.style={fill=white},
clip=false,
yticklabels={,,},
xticklabels={,,}
]
\addplot [semithick, C0, mark=*, mark size=1, mark options={solid}, only marks]
table {%
0.0411478153793777 0.00619509787901616
0.031088966871415 0.00628147933958715
0.0251961040414871 0.00634073501763791
0.0214513908835172 0.00629758784836845
0.0184212210635465 0.00624427343893159
0.0173518556623966 0.00638589499080119
0.01603448353249 0.00647677607995042
0.0148634360948103 0.00663652402964895
0.0136176084619929 0.00633819598099539
0.0134472266595547 0.00637335626376129
0.0126815693038882 0.0065409688534958
0.0120948011780637 0.00642188957511391
0.0115717561073237 0.00672099153661967
0.01121387752385 0.00674908865800919
0.0107545200762243 0.00677682833080307
0.0104365622360233 0.0067526624030303
0.0101881087252347 0.0066776582073486
0.00995761586453003 0.00670390494515648
0.00982982525372756 0.00684677378427535
0.00864207984755357 0.00676497440959448
0.00812241622408941 0.00705769167792255
0.00744198995442658 0.00696257644232432
0.00690940418346233 0.00677424623385555
0.00669187797962886 0.00712548366863515
0.00640357019395304 0.00690846517266827
0.00620540754974669 0.00672489799913334
0.00592054160218229 0.00677735042878557
0.00581994379123407 0.00698476330228835
0.00565695619773785 0.00673265475110774
0.00558332282324088 0.00681844268259598
0.00543127728503315 0.0069693471752895
0.0052867827797334 0.00697669383096753
0.00525639717410556 0.00700346508879379
0.00516329085387839 0.00673125769100021
0.00512979855953202 0.006958519926207
};
\end{axis}

\draw [dashed, opacity=0.2] (c1) -- (ax2.south east);
\draw [dashed, opacity=0.2] (c2) -- (ax2.north west);


\end{tikzpicture}}%
        \caption{\scriptsize 5D Reacher - Energy}
        \label{fig:reacher:energy}
    \end{subfigure}%
    \\
    \resizebox{\textwidth}{!}{
    \begin{tikzpicture} 
    \begin{axis}[%
    hide axis,
    xmin=10,
    xmax=50,
    ymin=0,
    ymax=0.4,
    legend style={
        draw=white!15!black,
        legend cell align=left,
        legend columns=-1, 
        legend style={
            draw=none,
            column sep=1ex,
            line width=1pt
        }
    },
    ]
    \addlegendimage{C1}
    \addlegendentry{\acrshort{ppo}};
    \addlegendimage{C5}
    \addlegendentry{\acrshort{trpl}};
    \addlegendimage{C2}
    \addlegendentry{\acrshort{ndp}};
    \addlegendimage{C6}
    \addlegendentry{\acrshort{es}};
    \addlegendimage{C7}
    \addlegendentry{\acrshort{cmore}};
    \addlegendimage{C8}
    \addlegendentry{\acrshort{sac}};
    \addlegendimage{C4}
    \addlegendentry{\acrshort{method-bb-ppo}};
    \addlegendimage{C0}
    \addlegendentry{\acrshort{method-bb}};
    \end{axis}
\end{tikzpicture}
    }%
    \caption{
    The figures \twopsubref{fig:reacher:dense}{fig:reacher:sparse} show the learning curve for the 5D reacher task with dense and sparse reward signals. Although \gls{trpl} and \gls{ppo} achieve the best performance in the dense reward setting, both of them are struggling under the sparse reward setting. In contrast, \gls{method-bb} achieves the best performance in the sparse reward setting.
    Figure \psubref{fig:reacher:energy} shows the tradeoff between the energy efficiency (sum of the squared control cost) and the task precision (distance to the goal at the last step). We average over 100 evaluation runs and all seeds and choose action penalty factors in the intervals $(0,100]$. \gls{method-bb} with sparse reward achieves the highest precision with much lower energy consumption compared to the \gls{ppo} trained with the dense reward. 
    %
    }
    \label{fig:reacher}
\end{figure}

\begin{figure}[t]
    \centering
    \captionsetup*[subfigure]{margin={0pt,10pt}}
    \begin{subfigure}{0.33\textwidth}
        \resizebox{\textwidth}{!}{\input{tikz_plots/box_pushing/box_pushing_dense_iqm}}
        \caption{\scriptsize Box Push - Dense}
        \label{fig:box_pushing:dense}
    \end{subfigure}%
    \begin{subfigure}{0.33\textwidth}
        \resizebox{\textwidth}{!}{\input{tikz_plots/box_pushing/box_pushing_sparse1_NEW}}%
        \caption{\scriptsize Box Push - Sparse}
        \label{fig:box_pushing:Tsparse}
    \end{subfigure}%
    \begin{subfigure}{0.325\textwidth}
        \resizebox{\textwidth}{!}{
\begin{tikzpicture}

\begin{axis}[
legend cell align={left},
legend style={fill opacity=0.8, draw opacity=1, text opacity=1, draw=lightgray204},
tick align=outside,
tick pos=left,
scaled x ticks=false,
xticklabels={,0,0.2,0.4,0.6,0.8,1},
x grid style={darkgray176},
xlabel={Control Cost ($\times 10^4$)},
xmajorgrids,
xmin=-500, xmax=10500,
xtick style={color=black},
y grid style={darkgray176},
ylabel={Success Rate},
ymajorgrids,
ymin=-0.05, ymax=1.05,
ytick style={color=black},
]
\addplot [draw=C0, fill=C0, mark=*, mark size=1, only marks]
table{%
x  y
13668.9782187624 0.904
12086.2467034566 0.925
9664.30419948519 0.921
8488.49515119024 0.928
6558.62511260695 0.934
4124.11519022654 0.938
3164.72114581158 0.931
2483.78145496797 0.921
1795.27888533972 0.904
1497.59629020714 0.918
1250.29004889935 0.915
915.592041823842 0.887
783.279365729972 0.835
692.854325586116 0.806
604.602725836629 0.674
};

\addplot [draw=C1, fill=C1, mark=square*, mark size=1, only marks]
table{%
x  y
36242.9563120944 0.923
11176.0713471751 0.967
7941.60221306806 0.965
7811.54141590076 0.963
7124.93820995149 0.959
5996.26589576516 0.96
5440.95930375086 0.938
4799.04647205061 0.952
4296.29281453872 0.906
2707.04020837567 0.81
3124.24833204274 0.852
1482.65692171329 0.523
971.730483101388 0.332
1048.87689528689 0.18
476.732573017216 0.194
};

\addplot [draw=C4, fill=C4, mark=triangle*, mark size=1, only marks]
table{%
x  y
5270.00758245373 0.837
3274.28637876125 0.865
2578.3765427498 0.876
2020.00002333305 0.879
1624.01703040571 0.865
1312.70799644618 0.82
959.323209226482 0.799
739.358036506305 0.762
624.188351409005 0.712
613.387146267677 0.591
};

\end{axis}

\end{tikzpicture}}%
        \caption{\scriptsize Box Push - Energy}
        \label{fig:box_pushing:energy}
    \end{subfigure}%
    \\
    \resizebox{0.8\textwidth}{!}{
        \begin{tikzpicture} 
    \begin{axis}[%
    hide axis,
    xmin=10,
    xmax=50,
    ymin=0,
    ymax=0.4,
    legend style={
        draw=white!15!black,
        legend cell align=left,
        legend columns=-1, 
        legend style={
            draw=none,
            column sep=1ex,
            line width=1pt
        }
    },
    ]
    \addlegendimage{C1}
    \addlegendentry{\acrshort{ppo}};
    \addlegendimage{C5}
    \addlegendentry{\acrshort{trpl}};
    \addlegendimage{C8}
    \addlegendentry{\acrshort{sac}};
    \addlegendimage{C4}
    \addlegendentry{\acrshort{method-bb-ppo}};
    \addlegendimage{C0}
    \addlegendentry{\acrshort{method-bb}};
    \addlegendimage{C9}
    \addlegendentry{\acrshort{method}}
    \end{axis}
\end{tikzpicture}
    }%
    \caption{
    Figure \psubref{fig:box_pushing:dense} shows the learning curve for the box pushing task with a dense reward signal, while \psubref{fig:box_pushing:Tsparse} with sparse reward signal. 
    For the dense reward all the method except \gls{sac} achieve remarkable performance, while \acrshort{method} and \acrshort{method-bb} achieve higher sample efficiency, this is due to the efficient exploration in parameters space. 
    All methods suffer performance decrease with sparse reward signal \psubref{fig:box_pushing:Tsparse}, \acrshort{method} and \acrshort{method-bb} are less influenced as they leverage highly temporal abstracted samples. \gls{method-bb} achieve similar sample efficiency with \acrshort{method} because it can take advantage of more compact observations (context observation). Figure \psubref{fig:box_pushing:energy} shows the tradeoff between energy efficiency(sum of squared action) and task precision (success rate). Similar to the energy plot in 5D reacher task (\Figref{fig:reacher:energy}), \gls{method-bb} shows better energy-efficient behavior.    
    %
    }
    \label{fig:box_pushing}
\end{figure}

\begin{figure}[!t]
    \centering
    \captionsetup*[subfigure]{margin={0pt,10pt}}
    \begin{subfigure}{0.33\textwidth}
        \resizebox{\textwidth}{!}{\input{tikz_plots/hopper_jump/hj_iqm_sample_efficiency_max_heights}}%
        \caption{\scriptsize Hopper Jump - Max Height}
        \label{fig:non-Markovian:height}
    \end{subfigure}%
    \begin{subfigure}{0.33\textwidth}
        \resizebox{\textwidth}{!}{\input{tikz_plots/hopper_jump/hj_iqm_sample_efficiency_goal_dists}}%
        \caption{\scriptsize Hopper Jump - Goal Distance}
        \label{fig:non-Markovian:distance}
    \end{subfigure}%
    \begin{subfigure}{0.33\textwidth}
        \resizebox{\textwidth}{!}{\input{tikz_plots/beer_pong/bp_iqm_sample_efficiency}}%
        \caption{\scriptsize Beer Pong}
        \label{fig:non-Markovian:beer}
    \end{subfigure}%
        \\
    \begin{subfigure}[t]{0.33\textwidth}
        \resizebox{\textwidth}{!}{\input{tikz_plots/table_tennis/table_tennis_iqm_hit_rate}}%
        \caption{\scriptsize Table Tennis - Hit Rate}
        \label{fig:non-Markovian:hit}
    \end{subfigure}%
    \begin{subfigure}[t]{0.33\textwidth}
        \resizebox{\textwidth}{!}{\input{tikz_plots/table_tennis/table_tennis_iqm_success_rate}}%
        \caption{\scriptsize Table Tennis - Success Rate}
        \label{fig:non-Markovian:success}
    \end{subfigure}%
    \\
    \resizebox{\textwidth}{!}{
        \begin{tikzpicture} 
    \begin{axis}[%
    hide axis,
    xmin=10,
    xmax=50,
    ymin=0,
    ymax=0.4,
    legend style={
        draw=white!15!black,
        legend cell align=left,
        legend columns=-1, 
        legend style={
            draw=none,
            column sep=1ex,
            line width=1pt
        }
    },
    ]
    \addlegendimage{C1}
    \addlegendentry{\acrshort{ppo}};
    \addlegendimage{C5}
    \addlegendentry{\acrshort{trpl}};
    \addlegendimage{C6}
    \addlegendentry{\acrshort{es}};
    \addlegendimage{C7}
    \addlegendentry{\acrshort{cmore}};
    \addlegendimage{C8}
    \addlegendentry{\acrshort{sac}};
    \addlegendimage{C4}
    \addlegendentry{\acrshort{method-bb-ppo}};
    \addlegendimage{C0}
    \addlegendentry{\acrshort{method-bb}};
    \addlegendimage{C9}
    \addlegendentry{\acrshort{method}}
    \end{axis}
\end{tikzpicture}
    }%
    \caption{
    The figures \twopsubref{fig:non-Markovian:height}{fig:non-Markovian:distance} show the maximum jumping height of the hopper's center of mass and the target distance, respectively. 
    With the non-Markovian reward, the hopper can jump approximately 20cm higher with increased goal precision.
    Figure \psubref{fig:non-Markovian:beer} shows the beer pong task, where \gls{ppo} struggles to throw the ball into the cup, even with a fixed optimal release point, while \gls{method-bb} can consistently succeed in the task with dynamic release points.
    The success and hit rate of the table tennis task are shown in \twopsubref{fig:non-Markovian:hit}{fig:non-Markovian:success}, respectively. 
    The episode is considered a success when the agent hits the ball and successfully returns the ball near the goal position. 
    The results demonstrate that \gls{method-bb} consistently hits the ball and returns it in most cases, and \acrshort{method} is even able to improve that performance further.}
    \label{fig:non-Markovian}
\end{figure}

\paragraph{Dealing with non-Markovian Rewards}
To assess the effectiveness of our method in complex reward settings, we test it with non-Markovian rewards, which are particularly useful for robot learning tasks that require the agent to use feedback from the full trajectory history.
We first use a modified version of the OpenAI Gym hopper \citep{Brockman2016}, which aims to jump as high as possible and land at a target location (see \Apxref{app:hopper_jump}).
The non-Markovian reward measures the highest point of the jump and the shortest distance to the target during the episode.
We compare our approach with two other methods, \gls{cmore} and \gls{es}, which also use the non-Markovian reward.
In addition, we train three different algorithms, \gls{ppo}, \gls{sac}, and \gls{trpl}, using a Markovian version that provides height and target distance at each time step.
For this setting, we performed extensive reward shaping to optimize for maximum height and minimum target distance.
Overall, our results show that \gls{method-bb} achieves higher jump heights with smaller target distances compared to most other methods (\twoFigref{fig:non-Markovian:height}{fig:non-Markovian:distance}). 
While \gls{method-bb-ppo} and \gls{cmore} can match the target distance, \gls{sac} can even exceed it, none of them can reliably learn a good jump height.
This can also be seen in the agent's behavior.
\gls{method-bb} charges energy and then jumps only once, whereas the step-based methods try to maximize height in each time step, resulting in multiple jumps in one episode.
This illustrates the need for non-Markovian rewards to describe certain behaviors.

To further strengthen the ability of \gls{method-bb} in solving tasks with non-Markovian rewards, we conduct experiments in a Beer pong environment\citep{celik2022specializing}.
In this task, the goal is to throw a ball into a cup at various locations on a table.
The return depends on the entire trajectory of the ball, which can be calculated by using information such as table contacts or the minimum distance to the cup (a detailed description of the task can be found at   \Apxref{app:beer_pong}). 
However, directly training \gls{ppo} on such a reward as well as designing a Markovian version of the reward is both challenging in this case.
To address this issue and make \gls{ppo} a stronger baseline, we simplified the task for \gls{ppo} by fixing the ball release time and considering the time between the ball release and the end of the ball trajectory as the last time step, allowing \gls{ppo} to compute the reward in a similar manner to the non-Markovian setting.
This kind of simplification is unnecessary for \gls{erl} algorithms.
We use \gls{method-bb} and \gls{cmore} with the non-Markovian reward and learn the ball release time as an additional controller parameter. See results at \Figref{fig:non-Markovian:beer}.
We see that both \gls{method-bb} and \gls{method-bb-ppo} are able to throw the ball into the cup, while \gls{ppo} struggles.
Even \gls{cmore} can throw the ball reliably, but only for a subset of the context space.
Interestingly, we again observe that \gls{method-bb-ppo} has a larger confidence interval compared to \gls{method-bb}, indicating that it is not always able to solve the task consistently.
This behavior is similar to the jumping task, where we also observed a larger confidence interval for \gls{method-bb-ppo}.

Finally, we trained agents in a simulated table tennis environment\citep{celik2022specializing}.
The context is four-dimensional, given by features of the incoming ball trajectory and the desired location for returning the ball (a detailed environment description is provided at \Apxref{app:table_tennis}). 
The episode return depends on several factors, such as the minimum distance between the racket trajectory and ball trajectory, whether the racket hits the ball, and the distance error between the ball's landing position and the target. It is worth noting that the agent's action cannot influence the return after hitting the ball. Thus the return after hitting is not conditioned on the action at the current step but on the previous actions, which poses a significant challenge for \gls{srl} algorithms.
For step-based methods, we consider the time after hitting the ball as one time step, akin to the beer pong task. 
For the \acrshort{method} and \gls{method-bb} approaches, we also learn the start time of the the trajectory and the speed of the desired trajectory (which is a parameter of the \gls{mp}). 
Both parameters help to learn the precise timing required to play table tennis. 
The results (\twoFigref{fig:non-Markovian:hit}{fig:non-Markovian:success}) show that \gls{method-bb} succeeds in hitting the ball and returning it within the vicinity of the goal in more than $60\%$ of the cases.
\gls{method-bb-ppo} and \gls{trpl} can hit the ball but fail to return it accurately, \gls{ppo} even fails to hit the ball consistently. 

In summary, \gls{method-bb} provides an effective solution for handling non-Markovian reward structures, which are often more natural and easier to define than engineered dense rewards. 
By leveraging these reward structures, the method can facilitate the learning of more sophisticated and efficient behaviors, leading to improved overall performance.

\subsection{Large Scale Robot Manipulation}
We also showcase our ability to learn high quality policies on the Meta-World benchmark suite \citep{Yu2019}.
To verify our algorithms can adapt to task variations and solve tasks consistently, we use a more rigorous evaluation protocol compared to the one used by \citet{Yu2019}. In contrast to using a fixed context for each episode, we randomly generate new contexts with each reset. Additionally, instead of considering any instance of success during the episode as a task solved, we consider a task successfully solved only if the last time step successfully solves the task. The last time step success metric rules out cases where the task is only momentarily solved but subsequently disrupted by random agent motion. 
We train individual policies for each environment but use the same hyperparameters.
 Our results (\Figref{fig:metaworld:sample}) show that \gls{ppo} and \gls{trpl} achieve the best sample complexity, but \gls{method-bb} performs competitively in terms of asymptotic performance and even outperforms \gls{ppo} slightly in terms of asymptotic performance.
Although the gap between \gls{ppo} and \gls{method-bb} in the aggregated view is relatively small, the corresponding performance profiles (\Figref{fig:metaworld:profile}) reveal that \gls{method-bb} performs better above the $80\%$ threshold.
This means that \gls{method-bb} finds more consistent solutions than \gls{ppo} with higher precision and solves these tasks without failures. 
\gls{sac} performs similar to \gls{ppo}, whereas \gls{ndp}, \gls{es}, and \gls{method-bb-ppo} are not achieving a competitive performance.

We conducted an additional ablation study in which we trained \gls{method-bb} using sparse rewards, meaning only the final step reward of each episode was used. 
We denote this variant as \gls{method-bb} sparse. 
While the \gls{iqm} score is lower, it still able to complete $50\%$ of the tasks with $100\%$ success rate.
This is a better performance than what we observed with \gls{ppo} trained with dense rewards.
Furthermore, the slope of the performance profile is quite small, indicating that nearly all the tasks that can be solved by the agent are solved with a high degree of accuracy. 

\begin{figure}
    \centering
    \captionsetup*[subfigure]{margin={20pt,0pt}}
    \hspace*{\fill}%
    \begin{subfigure}{0.4\textwidth}
        \resizebox{\textwidth}{!}{\input{tikz_plots/metaworld/metaworld_IQM_sample_efficiency}}%
        \caption{MetaWorld - Sample Efficiency}
        \label{fig:metaworld:sample}
    \end{subfigure}\hfill%
    \begin{subfigure}{0.4\textwidth}
        \resizebox{\textwidth}{!}{\input{tikz_plots/metaworld/metaworld_performance_profile_6e7}}%
        \caption{MetaWorld - Perf. Profile}
        \label{fig:metaworld:profile}
    \end{subfigure}%
    \hspace*{\fill}%
    \newline
    \resizebox{\textwidth}{!}{
        \begin{tikzpicture} 
    \begin{axis}[%
    hide axis,
    xmin=10,
    xmax=50,
    ymin=0,
    ymax=0.4,
    legend style={
        draw=white!15!black,
        legend cell align=left,
        legend columns=-1, 
        legend style={
            draw=none,
            column sep=1ex,
            line width=1pt
        }
    },
    ]
    \addlegendimage{C1}
    \addlegendentry{\acrshort{ppo}};
    \addlegendimage{C5}
    \addlegendentry{\acrshort{trpl}};
    \addlegendimage{C2}
    \addlegendentry{\acrshort{ndp}};
    \addlegendimage{C6}
    \addlegendentry{\acrshort{es}};
    \addlegendimage{C8}
    \addlegendentry{\acrshort{sac}};
    \addlegendimage{C4}
    \addlegendentry{\acrshort{method-bb-ppo}};
    \addlegendimage{C0}
    \addlegendentry{\acrshort{method-bb}};
    \addlegendimage{C3}
    \addlegendentry{\acrshort{method-bb} sparse};
    \addlegendimage{C9}
    \addlegendentry{\acrshort{method}};
    \end{axis}
\end{tikzpicture}
    }%
    \caption{
    This Plot shows the success rate \psubref{fig:metaworld:sample} for all 50 Meta-World tasks, and the corresponding performance profile \psubref{fig:metaworld:profile}, which is the fraction of runs that perform above the threshold given by the x-axis. 
    Despite having lower sample efficiency, \gls{method-bb} achieves a higher asymptotic policy quality than \gls{ppo}. 
    Finally, \gls{method} can further improve the asymptotic performance while being slightly less sample efficient. According to the performance profile \psubref{fig:metaworld:profile}, \acrshort{method-bb} can solve more tasks (fewer runs with zero success rate) while \acrshort{method} can solve the tractable tasks more consistently (higher fraction of runs with $\geq 80\%$ success rate).
    }
    \label{fig:metaworld}
\end{figure}

\subsection{Replanning with Movement Primitives}
\label{sec:replanning}
We evaluate our approach in the online replanning case by decreasing the planning horizon, such that $1 \leq k < T$, positioning it between \gls{srl} ($k=1$) and \gls{erl} ($k=T$). 
This approach, which we refer to as \gls{method}, offers two significant benefits. 
Firstly, it leads to a more precise policy due to the closed-loop nature of the method.
Secondly, it enables the handling of environmental dynamics through online replanning.
However, there are two reasons why it is advantageous to consider it as a complementary approach rather than a complete replacement for \gls{method-bb}. First, this design choice may limit the ability to address non-Markovian rewards, although we have also observed good results in this setting. Second, using replanning necessitates incorporating the internal proprioceptive state into the observation space, whereas the black-box setting can leverage a more concise observation space known as the context space.

\paragraph{Quality of the Learned Policy.}
We evaluated the performance of \gls{method} agents by conducting experiments in three challenging environments: the Meta-World benchmark suite \citep{Yu2019} for large-scale robot manipulation, Box Pushing with dense and sparse rewards, and Table Tennis with non-Markovian reward.

We kept the same planning horizon ($20\%$ of the max episode length) in all 50 Meta-World environments. While the step-based methods achieved the best sample efficiency for Meta-World (\Figref{fig:replanning:metaworld}), \gls{method} slightly outperforms the black-box approach and  reaches the best asymptotic \gls{iqm} score. 
The performance profile (\Figref{fig:metaworld:profile}) further indicates that although the number of unsolvable tasks remains constant, the quality of the solved tasks improves.

In all the box-pushing experiments (\Figref{fig:replanning:box_pushing}), we kept the planning horizon to $25\%$ of the max episode length. \gls{method} achieved the same performance with \gls{method-bb} regarding success rate and sample efficiency in dense reward setting. In sparse reward setting, \gls{method} exhibited better precision and sample efficiency compared to \gls{method-bb}. This is primarily due to the use of different \glspl{mp} representations (\glspl{pdmp} for \gls{method} and \glspl{promp} for \gls{method-bb}). \glspl{pdmp}-based policies tend to generate trajectories with lower episode energy, which helps the agent to focus on the main target (move the box to the target) instead of regularized by the control penalty. We conducted an ablation in \Figref{fig:ablation_promps_vs_prodmps:bb} to verify this assumption.

While non-Markovian rewards cannot be used as freely as in the \gls{method-bb} case, it is still possible to leverage them as long as the non-Markovian behavior is limited to one trajectory segment.  
For table tennis, we choose planning horizon $k$ such that we ensure the last trajectory segment starts before the racket hits the ball.
In this setting, \gls{method} learns table tennis skills with a better success rate and much fewer samples than \gls{method-bb} (\Figref{fig:replanning:table_tennis}). 

In conclusion, we observed that the use of replanning yields better asymptotic performance in all cases while it can harm slightly the sample efficiency (observed in Meta-World experiments). We attribute this to the higher dimensional state space that must be considered in the replanning case compared to the black-box case. 

\begin{figure}[t]
    \centering
    \captionsetup*[subfigure]{margin={0pt,10pt}}
    \begin{subfigure}{0.33\textwidth}
        \resizebox{\textwidth}{!}{\input{tikz_plots/replanning/metaworld_static_success_rate}}
        \caption{\scriptsize MetaWorld - 50 Tasks}
        \label{fig:replanning:metaworld}
    \end{subfigure}%
    \begin{subfigure}{0.33\textwidth}
        \resizebox{\textwidth}{!}{\input{tikz_plots/replanning/box_pushing_static_success_rate}}%
        \caption{\scriptsize Box Pushing}
        \label{fig:replanning:box_pushing}
    \end{subfigure}%
    \begin{subfigure}{0.33\textwidth}
        \resizebox{\textwidth}{!}{\input{tikz_plots/replanning/table_tennis_static_success_rate}}%
        \caption{\scriptsize Table Tennis}
        \label{fig:replanning:table_tennis}
    \end{subfigure}%
    \hspace*{\fill}%
    \newline
    \resizebox{0.25\textwidth}{!}{
        \begin{tikzpicture} 
    \begin{axis}[%
    hide axis,
    xmin=10,
    xmax=50,
    ymin=0,
    ymax=0.4,
    legend style={
        draw=white!15!black,
        legend cell align=left,
        legend columns=-1, 
        legend style={
            draw=none,
            column sep=1ex,
            line width=1pt
        }
    },
    ]
    \addlegendimage{C0}
    \addlegendentry{\acrshort{method-bb}};
    \addlegendimage{C9}
    \addlegendentry{\acrshort{method}}
    \end{axis}
\end{tikzpicture}
    }%
    \caption{
    Figure \psubref{fig:replanning:metaworld} presents a comparison of the learning curves between \gls{method} and \gls{method-bb} across Meta-World's 50 tasks.
    Due to its closed-loop nature, \gls{method} achieves higher asymptotic performance with a marginal compromise in sample efficiency. This is attributed to the advantage of \gls{method-bb} in utilizing a more compact observation space. 
    Figure \psubref{fig:replanning:box_pushing} shows the performance of both methods in the box pushing task, considering both dense (solid) and sparse (dashed) reward settings. While both methods achieve similar sample efficiency and asymptotic performance in dense reward settings, \gls{method} reaches a higher success rate in the sparse reward setting.
    Figure \psubref{fig:replanning:table_tennis} demonstrates the hit rate (dashed) and success rate (solid) in the robot table tennis task. Both methods can consistently hit the ball, but \gls{method} outperforms \gls{method-bb} by returning the ball with higher precision and requiring significantly fewer samples to converge. 
    }
    \label{fig:replanning}
\end{figure}

\begin{figure}[!t]
    \centering
    \captionsetup*[subfigure]{margin={15pt,0pt}}
    \begin{subfigure}{0.33\textwidth}
        \resizebox{\textwidth}{!}{
\begin{tikzpicture}

\begin{axis}[
legend cell align={left},
legend style={fill opacity=0.8, draw opacity=1, text opacity=1, draw=lightgray204, at={(0.03,0.95)},  anchor=north west},
tick align=outside,
tick pos=left,
x grid style={darkgray176},
scaled x ticks=false,
xticklabels={,0,1,2,3,4},
xlabel={Number Environment Interactions ($\times 10^7$)},
xmajorgrids,
xmin=-2500000, xmax=42000000.05,
xtick style={color=black},
y grid style={darkgray176},
ylabel={Success Rate},
ymajorgrids,
ymin=-0.05, ymax=1.05,
ytick style={color=black}
]

\path [draw=C1, fill=C1, opacity=0.2]
(axis cs:0,0)
--(axis cs:0,0)
--(axis cs:1520000,0)
--(axis cs:3040000,0)
--(axis cs:4560000,0)
--(axis cs:6080000,0)
--(axis cs:7600000,0)
--(axis cs:9120000,0)
--(axis cs:10640000,0)
--(axis cs:12160000,0.0166666666666667)
--(axis cs:13680000,0)
--(axis cs:15200000,0)
--(axis cs:16720000,0)
--(axis cs:18240000,0)
--(axis cs:19760000,0)
--(axis cs:21280000,0)
--(axis cs:22800000,0)
--(axis cs:24320000,0.0166666666666667)
--(axis cs:25840000,0)
--(axis cs:27360000,0)
--(axis cs:28880000,0)
--(axis cs:30400000,0)
--(axis cs:31920000,0)
--(axis cs:33440000,0)
--(axis cs:34960000,0)
--(axis cs:36480000,0)
--(axis cs:38000000,0)
--(axis cs:39520000,0.0166666666666667)
--(axis cs:39520000,0.333333333333333)
--(axis cs:38000000,0.267083333333327)
--(axis cs:36480000,0.333333333333333)
--(axis cs:34960000,0.316666666666667)
--(axis cs:33440000,0.266666666666667)
--(axis cs:31920000,0.316666666666667)
--(axis cs:30400000,0.2)
--(axis cs:28880000,0.133333333333333)
--(axis cs:27360000,0.233333333333333)
--(axis cs:25840000,0.233333333333333)
--(axis cs:24320000,0.233333333333333)
--(axis cs:22800000,0.266666666666667)
--(axis cs:21280000,0.2)
--(axis cs:19760000,0.133333333333333)
--(axis cs:18240000,0.25)
--(axis cs:16720000,0.133333333333333)
--(axis cs:15200000,0.133333333333333)
--(axis cs:13680000,0.1)
--(axis cs:12160000,0.233333333333333)
--(axis cs:10640000,0.133333333333333)
--(axis cs:9120000,0.116666666666667)
--(axis cs:7600000,0.116666666666667)
--(axis cs:6080000,0.05)
--(axis cs:4560000,0.0333333333333333)
--(axis cs:3040000,0.0333333333333333)
--(axis cs:1520000,0)
--(axis cs:0,0)
--cycle;

\addplot [thick, solid, C1, mark=*, mark size=0, mark options={solid}]
table {%
0 0
1520000 0
3040000 0
4560000 0
6080000 0
7600000 0.0166666666666667
9120000 0.0166666666666667
10640000 0.0333333333333333
12160000 0.0833333333333333
13680000 0.0166666666666667
15200000 0.0333333333333333
16720000 0
18240000 0.0333333333333333
19760000 0.0333333333333333
21280000 0.05
22800000 0.0333333333333333
24320000 0.1
25840000 0.0333333333333333
27360000 0.0333333333333333
28880000 0
30400000 0.0166666666666667
31920000 0.0833333333333333
33440000 0.0333333333333333
34960000 0.0833333333333333
36480000 0.1
38000000 0.05
39520000 0.1
};

\path [draw=C5, fill=C5, opacity=0.2]
(axis cs:0,0)
--(axis cs:0,0)
--(axis cs:1596000,0)
--(axis cs:3192000,0)
--(axis cs:4788000,0)
--(axis cs:6384000,0)
--(axis cs:7980000,0)
--(axis cs:9576000,0)
--(axis cs:11172000,0)
--(axis cs:12768000,0)
--(axis cs:14364000,0.02)
--(axis cs:15960000,0)
--(axis cs:17556000,0.01)
--(axis cs:19152000,0)
--(axis cs:20748000,0)
--(axis cs:22344000,0)
--(axis cs:23940000,0.01)
--(axis cs:25536000,0.03)
--(axis cs:27132000,0.03)
--(axis cs:28728000,0.07)
--(axis cs:30324000,0.02)
--(axis cs:31920000,0.02)
--(axis cs:33516000,0.02)
--(axis cs:35112000,0.03)
--(axis cs:36708000,0.03)
--(axis cs:38304000,0.04)
--(axis cs:39900000,0.02)
--(axis cs:39900000,0.31)
--(axis cs:38304000,0.24)
--(axis cs:36708000,0.26)
--(axis cs:35112000,0.24)
--(axis cs:33516000,0.23)
--(axis cs:31920000,0.2)
--(axis cs:30324000,0.28)
--(axis cs:28728000,0.3)
--(axis cs:27132000,0.2)
--(axis cs:25536000,0.23)
--(axis cs:23940000,0.15)
--(axis cs:22344000,0.15)
--(axis cs:20748000,0.1)
--(axis cs:19152000,0.08)
--(axis cs:17556000,0.13)
--(axis cs:15960000,0.11)
--(axis cs:14364000,0.12)
--(axis cs:12768000,0.08)
--(axis cs:11172000,0.04)
--(axis cs:9576000,0.07)
--(axis cs:7980000,0)
--(axis cs:6384000,0.01)
--(axis cs:4788000,0)
--(axis cs:3192000,0)
--(axis cs:1596000,0)
--(axis cs:0,0)
--cycle;

\addplot [thick, solid, C5, mark=*, mark size=0, mark options={solid}]
table {%
0 0
1596000 0
3192000 0
4788000 0
6384000 0
7980000 0
9576000 0.03
11172000 0
12768000 0.03
14364000 0.06
15960000 0.04
17556000 0.05
19152000 0.01
20748000 0.02
22344000 0.02
23940000 0.05
25536000 0.12
27132000 0.09
28728000 0.14
30324000 0.09
31920000 0.08
33516000 0.11
35112000 0.11
36708000 0.1
38304000 0.1
39900000 0.11
};

\path [draw=C9, fill=C9, opacity=0.2]
(axis cs:0,0)
--(axis cs:0,0)
--(axis cs:1873508.57142857,0)
--(axis cs:3747017.14285714,0)
--(axis cs:5620525.71428571,0)
--(axis cs:7494034.28571429,0)
--(axis cs:9367542.85714286,0.11)
--(axis cs:11241051.4285714,0.13)
--(axis cs:13114560,0.3)
--(axis cs:14988068.5714286,0.36)
--(axis cs:16861577.1428571,0.37)
--(axis cs:18735085.7142857,0.35)
--(axis cs:20608594.2857143,0.37)
--(axis cs:22482102.8571429,0.41)
--(axis cs:24355611.4285714,0.45)
--(axis cs:26229120,0.43)
--(axis cs:28102628.5714286,0.39)
--(axis cs:29976137.1428571,0.49)
--(axis cs:31849645.7142857,0.55)
--(axis cs:33723154.2857143,0.6)
--(axis cs:35596662.8571429,0.53)
--(axis cs:37470171.4285714,0.53)
--(axis cs:39343680,0.63)
--(axis cs:39343680,0.78)
--(axis cs:37470171.4285714,0.68)
--(axis cs:35596662.8571429,0.66)
--(axis cs:33723154.2857143,0.73)
--(axis cs:31849645.7142857,0.7)
--(axis cs:29976137.1428571,0.7)
--(axis cs:28102628.5714286,0.670249999999996)
--(axis cs:26229120,0.62)
--(axis cs:24355611.4285714,0.7)
--(axis cs:22482102.8571429,0.66)
--(axis cs:20608594.2857143,0.56)
--(axis cs:18735085.7142857,0.56)
--(axis cs:16861577.1428571,0.54)
--(axis cs:14988068.5714286,0.61)
--(axis cs:13114560,0.45)
--(axis cs:11241051.4285714,0.35)
--(axis cs:9367542.85714286,0.220249999999996)
--(axis cs:7494034.28571429,0.07)
--(axis cs:5620525.71428571,0.03)
--(axis cs:3747017.14285714,0)
--(axis cs:1873508.57142857,0)
--(axis cs:0,0)
--cycle;

\addplot [thick, solid, C9, mark=*, mark size=0, mark options={solid}]
table {%
0 0
1873508.57142857 0
3747017.14285714 0
5620525.71428571 0
7494034.28571429 0.02
9367542.85714286 0.15
11241051.4285714 0.22
13114560 0.37
14988068.5714286 0.48
16861577.1428571 0.46
18735085.7142857 0.45
20608594.2857143 0.47
22482102.8571429 0.56
24355611.4285714 0.58
26229120 0.51
28102628.5714286 0.53
29976137.1428571 0.61
31849645.7142857 0.61
33723154.2857143 0.66
35596662.8571429 0.59
37470171.4285714 0.6
39343680 0.7
};

\end{axis}

\end{tikzpicture}}%
        \caption{\scriptsize Box Push - Switch}
        \label{fig:dynamics:switch20}
    \end{subfigure}%
    \begin{subfigure}{0.33\textwidth}
        \resizebox{\textwidth}{!}{\input{tikz_plots/table_tennis/table_tennis_4d_goal_switching.tex}}%
        \caption{\scriptsize Table Tennis - Switch}
        \label{fig:dynamics:switchTT}
    \end{subfigure}%
    \begin{subfigure}{0.325\textwidth}
        \resizebox{\textwidth}{!}{\input{tikz_plots/table_tennis/table_tennis_4d_wind.tex}}%
        \caption{\scriptsize Table Tennis - Wind}
        \label{fig:dynamics:wind}
    \end{subfigure}%
    \\%
    \resizebox{0.5\textwidth}{!}{
    \begin{tikzpicture} 
    \begin{axis}[%
    hide axis,
    xmin=10,
    xmax=50,
    ymin=0,
    ymax=0.4,
    legend style={
        draw=white!15!black,
        legend cell align=left,
        legend columns=-1, 
        legend style={
            draw=none,
            column sep=1ex,
            line width=1pt
        }
    },
    ]
    \addlegendimage{C1}
    \addlegendentry{\acrshort{ppo}};
    \addlegendimage{C5}
    \addlegendentry{\acrshort{trpl}};
    \addlegendimage{C0}
    \addlegendentry{\acrshort{method-bb}};
    \addlegendimage{C9}
    \addlegendentry{\acrshort{method}}
    \end{axis}
\end{tikzpicture}
    }%
    \caption{
    This figure displays the success rate of perturbed tasks with and without replanning. 
    The success rate of box pushing with goal switching is shown at \psubref{fig:dynamics:switch20}, as well as the success rate of table tennis with goal switching \psubref{fig:dynamics:switchTT} and with wind \psubref{fig:dynamics:wind}. 
    In the box pushing tasks \psubref{fig:dynamics:switch20}, the solid lines represent learning curves of dense reward, while the dashed lines are learning curves from temporal-sparse reward. While step-based algorithms with dense reward already struggle in the easier $20\%$ setting, \gls{method} trained with sparse reward can solve both settings with a remarkable success rate.
    In table tennis tasks \twopsubref{fig:dynamics:switchTT}{fig:dynamics:wind}, the solid lines represent the success rates, and the dashed lines are hit rate.
    \gls{method} can effectively handle changes in tasks and environmental perturbations.
    In contrast, the \gls{method-bb} fails in these cases, as it only relies on a single observation at the beginning of the episode, which lacks critical information about the environment and task dynamics.}
    \label{fig:dynamics}
\end{figure}

\paragraph{Dealing with Uncertainties in the Environments.}
To demonstrate the robustness of \gls{method} in handling unforeseen events in the environment, we modified the box pushing and the table tennis tasks to include uncertainties that require incorporating feedback throughout the execution of the episode.

In the box-pushing experiments, we randomly switch to a new target position and orientation during execution after $20\%$ of the max episode length. 
We compared the performance of our method against step-based \gls{ppo} and \gls{trpl} in the dense reward setting. 
The results in \Figref{fig:dynamics:switch20} show that \gls{method} achieved the best performance under this setting.
To investigate the reasons behind this performance gap, we conducted a qualitative analysis of the trained policies with all three methods. We observed that the policies generated by \gls{ppo} and \gls{trpl} accelerate much faster and keep a high velocity from the beginning of the episode, which makes adapting to the new targets more difficult with the presence of control cost penalty and limited episode length. On the other hand, the acceleration and velocity of \gls{method} agents are regularized by the \gls{pdmp}'s representation, leading to uniform motion, thus yielding an amenable success rate in this challenging goal-switching setting.

For the table tennis environment, we test two kinds of uncertainties. 
We compare \gls{method} only with the \gls{method-bb} as the \gls{srl} algorithms have shown to be incapable of solving the table tennis task even in a static environment.
Firstly, we modify the desired landing position of the ball, similar to the goal change for the box pushing task.
Specifically, we initialize the desired landing position at a random location on the left side of the table. 
After half of the maximum episode steps, there is a $50\%$ chance that the target landing position will change to a new random position on the right side of the table. 
Our results in \Figref{fig:dynamics:switchTT} suggest that the \gls{method} agent is able to adapt its behavior and return the ball to the new target point with high precision.
In contrast, the \gls{method-bb} agent, which only receives the initial observation containing the initial target position, can only hit the ball but cannot solve this task.
While we would expect the \gls{method-bb} agent to solve at least those cases where the goal is not altered, we found that the conflicting reward feedback hinders the agent from learning high-quality policies. 
Secondly, we add wind to the environment by applying a random force to the ball, which is unknown to the agent and constant for an entire episode. 
However, the agent can still infer the underlying applied force according to the velocity of the ball, but only after observing the ball for a certain number of time steps.
Due to the wind, the \gls{method-bb} agent is not able to hit the ball consistently, while the \gls{method} agent slightly drops in performance but can still achieve reasonably good results (\Figref{fig:dynamics:wind}).

\subsection{Ablation Studies}
We conduct ablation studies to evaluate each component's influence on the proposed method that aim to answer the following questions: 
\begin{itemize}
    \item[\textbf{Q1}] What is the impact of varying the number of bases and the length of the replanning horizon on the performance of \gls{method}?
    \item[\textbf{Q2}] How does the performance of the non dynamic-based \gls{promp} with replanning compare to \gls{method}? 
    \item[\textbf{Q3}] Can the policy be effectively learned in the parameter space without incorporating proper trust regions? 
    \item[\textbf{Q4}] How does the performance of dynamic-based \glspl{pdmp} compare to non dynamic-based \glspl{promp} in \gls{method-bb} setting?
\end{itemize}

\begin{figure}[!t]
    \centering
    \captionsetup*[subfigure]{margin={0pt,15pt}}
    \begin{subfigure}{0.5\textwidth}
        \resizebox{\textwidth}{!}{\includegraphics[width=0.5\textwidth]{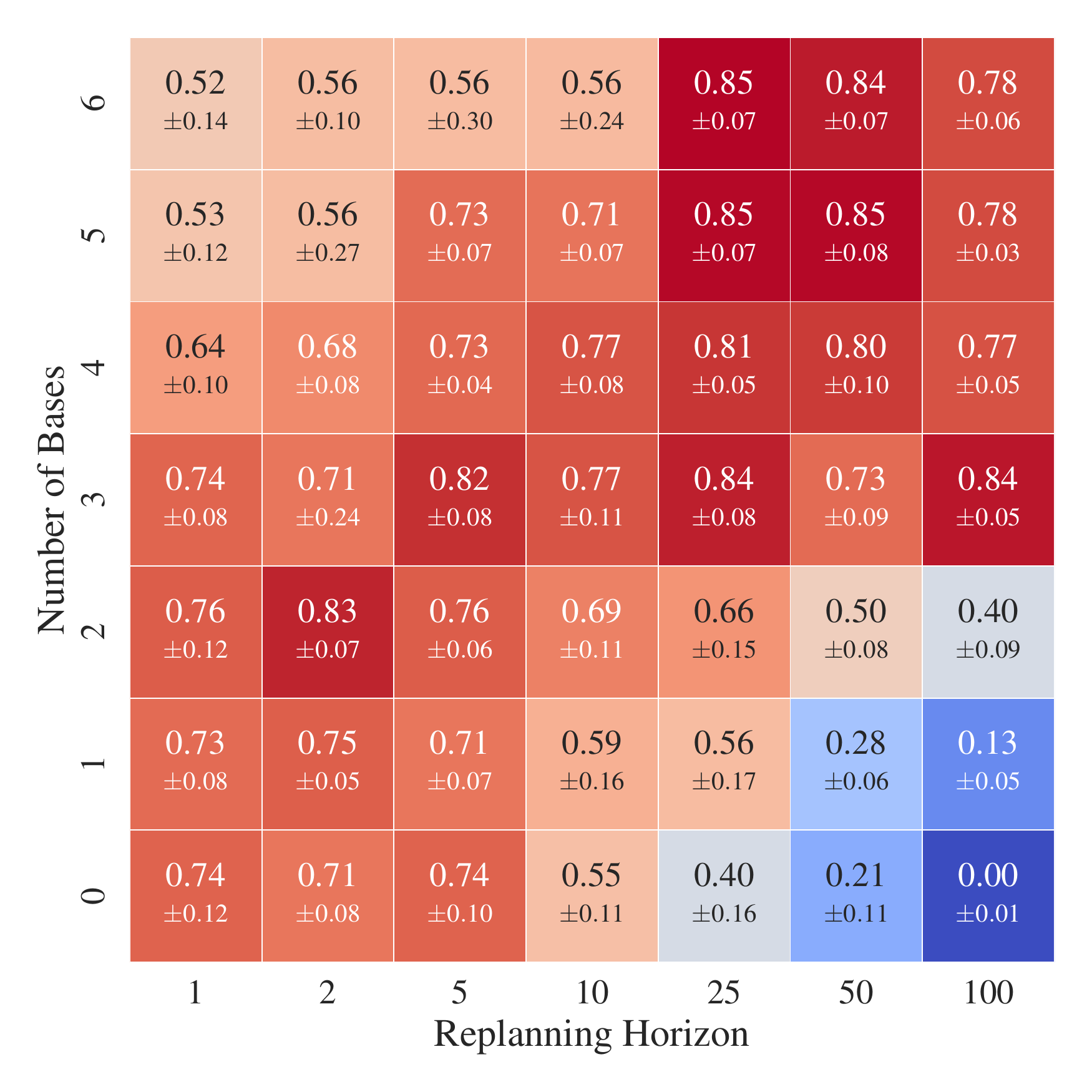}}
        \caption{\scriptsize Median Success Rate for Dense Reward}
        \label{fig:ablation_basis_steps_pdf:dense}
    \end{subfigure}%
    \begin{subfigure}{0.5\textwidth}
        \resizebox{\textwidth}{!}{\includegraphics[width=0.5\textwidth]{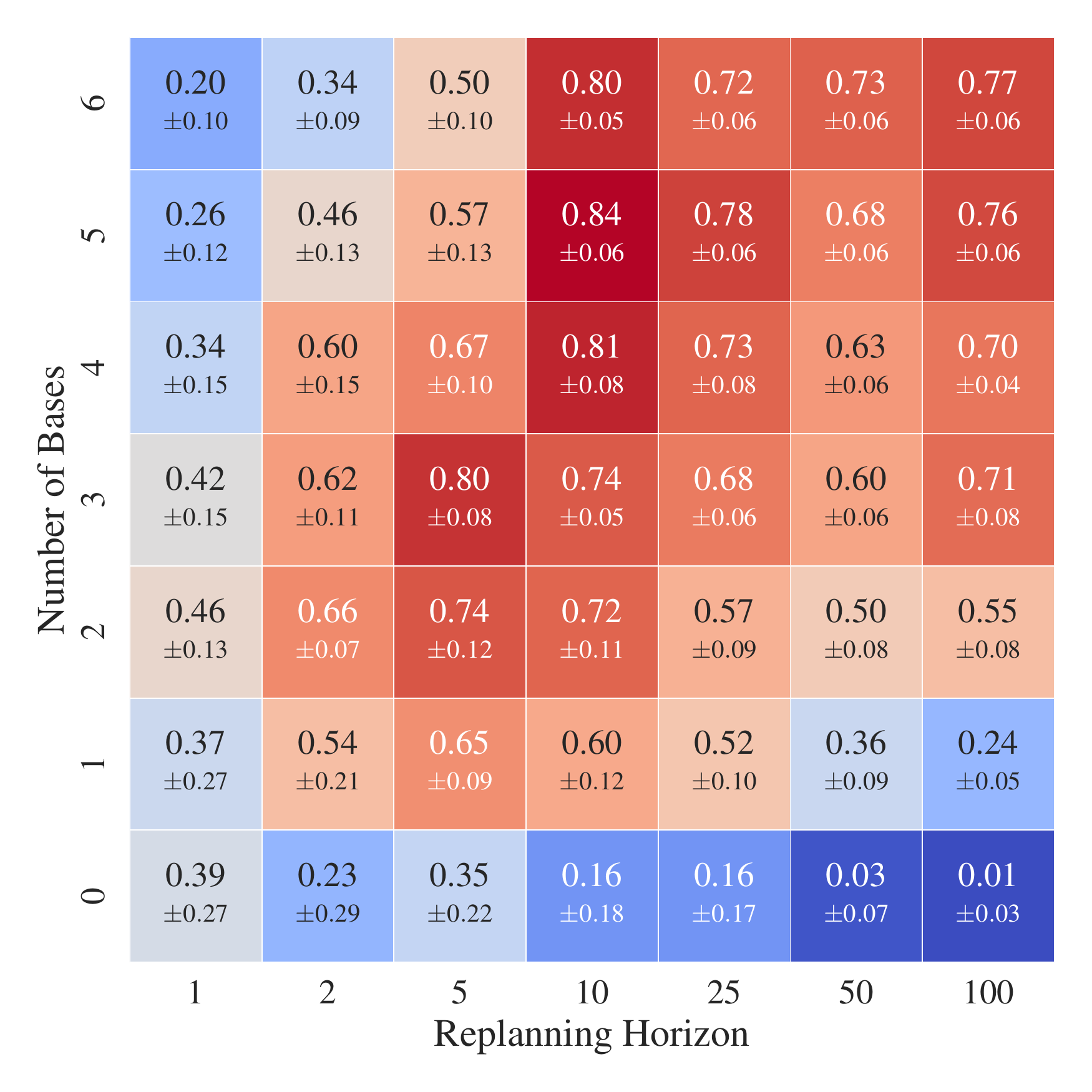}}
        \caption{\scriptsize Median Success Rate for Sparse Reward}
        \label{fig:ablation_basis_steps_pdf:sparse}
    \end{subfigure}%
    \caption{
    This figure shows the median success rate and standard deviation for different configurations of box pushing environments using dense \psubref{fig:ablation_basis_steps_pdf:dense} and sparse reward \psubref{fig:ablation_basis_steps_pdf:sparse}, with varying numbers of bases $N$ and replanning horizons $k$.
    When $N = 0$, the weight bases are disabled, and only the goal basis of the \gls{pdmp} is used, with the action space dimension equal to that of \gls{srl}.
    $k=1$ and $k=100=T$ correspond to \gls{srl} with \glspl{mp} and \gls{method-bb}, respectively.
    We evaluate each combination using ten random seeds and $20$ contexts per seed. 
    In general, the results suggest that: 1) longer planning horizons require more bases for optimal performance, 2) long planning horizons can help improve performance in sparse reward settings.
    }
    \label{fig:ablation_basis_steps_pdf}
\end{figure}

Firstly, we study the correlations between the number of bases of \glspl{mp} and the length of the planning horizon (replanning steps) in \Figref{fig:ablation_basis_steps_pdf}.
We train agents in box pushing environments with both dense and sparse rewards, using different combinations of planning horizons $k \in \{1, 2, 5, 10, 25, 50, 100\}$ and the number of bases $N \in \{0, 1, 2, 3, 4, 5, 6\}$. 
A value of $0$ for the number of bases indicates that the agent only uses the goal basis of the \glspl{pdmp}, leading to the same action space dimension as \gls{srl} algorithms.
When the planning horizon is equal to $1$, the learning objective of replanning reduces to a \gls{srl} objective. 
In this case, the only difference between replanning and \gls{srl} is that the agent explores the parameters space of the \gls{mp}, which usually has a higher dimensionality ((N + 1) $\times$ DoF) compared to the action space that a step-based agent explores. 
Another special case is when the replanning horizon equals the episode length ($k=100=T$), which corresponds to the \gls{method-bb} setting.
\\
\textbf{Q1} can be answered according to results in \Figref{fig:ablation_basis_steps_pdf}.
First, planning with a longer horizon requires a greater number of bases to achieve optimal performance. A longer planning horizon means less chance for the agent to adapt trajectories by adjusting the weights, limiting the ability to generate complex trajectories. This limitation, in turn, reduces performance in tasks that require fine manipulation, such as box pushing.
Second, longer planning horizons contribute to improved performance in the (temporal) sparse reward setting. This is attributed to the usage of high temporal abstracted samples in the policy updates. However, it does not necessarily mean \gls{method-bb} will always perform better in the sparse reward setting, as the black-box setting lacks the ability to correct its behavior due to the absence of the feedback signals from inter-execution observations.

For \textbf{Q2}, we compare replanning with dynamic-based (\glspl{pdmp}) and non dynamic-based \glspl{mp} (\glspl{promp}) in \twoFigref{fig:ablation_promps_vs_prodmps:prompVSpdmp}{fig:ablation_promps_vs_prodmps:cost}. 
The results demonstrate that the policy with dynamic-based \glspl{mp} yields a policy with a higher success rate and lower control cost.
This is largely due to the fact that non-dynamic \glspl{mp} can result in abrupt transitions between different planning segments, leading to discontinuities in the motion.

\begin{figure}
    \centering
    \begin{subfigure}{0.25\textwidth}
        \resizebox{\textwidth}{!}{\input{tikz_plots/box_pushing/box_pushing_promp_vs_prodmp_is_success.tex}}
        \caption{\scriptsize \acrshort{method}: Success}
        \label{fig:ablation_promps_vs_prodmps:prompVSpdmp}
    \end{subfigure}%
    \begin{subfigure}{0.25\textwidth}
        \resizebox{\textwidth}{!}{\input{tikz_plots/box_pushing/box_pushing_promp_vs_prodmp_energy.tex}}
        \caption{\scriptsize \acrshort{method}: Control Cost}
        \label{fig:ablation_promps_vs_prodmps:cost}
    \end{subfigure}%
    \begin{subfigure}{0.25\textwidth}
        \resizebox{\textwidth}{!}{\input{tikz_plots/box_pushing/box_pushing_replan_ppo_vs_trpl_is_success.tex}}
        \caption{\scriptsize \acrshort{method}: \acrshort{ppo} vs \acrshort{trpl}}
        \label{fig:ablation_promps_vs_prodmps:ppoVStrpl}
    \end{subfigure}%
    \begin{subfigure}{0.25\textwidth}
        \resizebox{\textwidth}{!}{\input{tikz_plots/box_pushing/box_pushing_prompbb_vs_prodmpbb.tex}}
        \caption{\scriptsize BB: \acrshort{promp} vs \acrshort{pdmp}}
        \label{fig:ablation_promps_vs_prodmps:bb}
    \end{subfigure}%
    \\
    \resizebox{0.55\textwidth}{!}{
        \begin{tikzpicture} 
    \begin{axis}[%
    hide axis,
    xmin=10,
    xmax=50,
    ymin=0,
    ymax=0.4,
    legend style={
        draw=white!15!black,
        legend cell align=left,
        legend columns=-1, 
        legend style={
            draw=none,
            column sep=1ex,
            line width=1pt
        }
    },
    ]
    \addlegendimage{C4}
    \addlegendentry{Dense Reward};
    \addlegendimage{C1}
    \addlegendentry{Temporal Sparse Reward};
    
    \end{axis}
\end{tikzpicture}
    }%
    \caption{
    This figure presents an ablation study for \gls{method} and \gls{method-bb} with different MPs and learning algorithms.
    The figures \twopsubref{fig:ablation_promps_vs_prodmps:prompVSpdmp}{fig:ablation_promps_vs_prodmps:cost} show the success rate and episode control cost for the box-pushing task, respectively.
    Here, we compare \gls{method} with \glspl{pdmp} (solid) to \gls{method} with \glspl{promp} (dashed) to show the need for \glspl{pdmp} when replanning.
    Figure \psubref{fig:ablation_promps_vs_prodmps:ppoVStrpl} demonstrates the need for using \gls{trpl} in \gls{method} (solid). 
    \gls{method-ppo} with \glspl{pdmp} using replanning (dashed) on box pushing tasks is not able to achieve the same performance. 
    Lastly, the figure \psubref{fig:ablation_promps_vs_prodmps:bb} shows that \gls{method-bb} with \glspl{pdmp} (dashed) is performing similarly to the \gls{method-bb} with \glspl{promp} (solid) in dense reward setting, and slightly better in the sparse reward setting. 
    }
    \label{fig:ablation_promps_vs_prodmps}
\end{figure}

To address \textbf{Q3}, we evaluate policy search algorithms without trust regions in the replanning setting, and present the results in \Figref{fig:ablation_promps_vs_prodmps:ppoVStrpl}. 
In both dense and sparse reward settings of box pushing, \gls{method} outperforms \gls{method-ppo} in terms of sample efficiency and success rate. 
The need for a more stable optimization and the higher dimensional nature of learning in parameter spaces could account for this observed improvement.

Finally, to answer \textbf{Q4}, we compare the performance between the black-box agent with \glspl{pdmp} and with \glspl{promp}. 
The results in \Figref{fig:ablation_promps_vs_prodmps:bb} show that in dense and sparse reward settings, both algorithms' sample efficiency and success rate are similar. In the sparse reward setting, \gls{method-bb} with \glspl{promp} shows slightly higher asymptotic performance. This difference is due to the different shapes of bases, and we believe the minor performance gap can be mitigated by selecting MP's parameters that minimize the differences in bases. The overall results suggest that the types of \gls{mp} make no significant difference in the black-box setting.

\section{Conclusion and Limitations}
\label{sec:conclusion}
Our work presents a new approach for combining \gls{srl} and \gls{erl} by integrating recent advancements in trust-region-based policy search \citep{Otto2021} and \glspl{mp} \citep{li2022prodmps}.
Unlike the commonly used step-based exploration in \gls{srl}, our method incorporates consistent and effective exploration at the trajectory level.
This approach is a promising way to handle tasks with sparse and non-Markovian rewards, enabling a more intuitive reward design. 
Furthermore, our method showed competitive performance against state-of-the-art \gls{srl} algorithms in large-scale robot manipulation tasks, as confirmed by thorough empirical evaluations.

Although our proposed method shows promise, there remain some limitations that require addressing in future work. 
Firstly, our current approach only considers fixed-length planning horizons and relies solely on time-based replanning triggers. 
Yet, in real-world applications, it may be necessary to incorporate event-based replanning triggers, such as detecting unforeseen obstacles or changing targets. 
Therefore, we will investigate how to integrate event-based triggers into our method in the future.
Secondly, our method, and \gls{erl} approaches in general, typically require more interaction time than \gls{srl} in dense reward settings. 
This is mainly due to the encapsulation of temporal-correlated information in highly abstracted samples. 
To address this issue, we will leverage the information within each planning segment to improve efficiency.
Lastly, our evaluation of the Meta-World benchmark suite identified that most failure cases occur in tasks requiring sub-goal achievement and skill sequencing. Therefore, we will investigate how to achieve long-horizon planning by incorporating sub-goals into our framework in future work.


\acks{
The presented work was funded by the Carl Zeiss Foundation through the Project JuBot (Jung Bleiben mit Robotern), the Helmholtz Association of German Research Centers and  by the Deutsche Forschungsgemeinschaft (DFG),
German Research Foundation under Grant 448648559 (Intuitive Robot Intelligence).
This work was supported by the state of Baden-Württemberg through bwHPC.
This work was performed on the HoreKa supercomputer funded by the
Ministry of Science, Research and the Arts Baden-Württemberg and by
the Federal Ministry of Education and Research.
}


\vskip 0.2in
\bibliography{bibliography}

\begin{thebibliography}{54}
\providecommand{\natexlab}[1]{#1}
\providecommand{\url}[1]{\texttt{#1}}
\expandafter\ifx\csname urlstyle\endcsname\relax
  \providecommand{\doi}[1]{doi: #1}\else
  \providecommand{\doi}{doi: \begingroup \urlstyle{rm}\Url}\fi

\bibitem[Abdolmaleki et~al.(2015)Abdolmaleki, Lioutikov, Peters, Lau,
  Pualo~Reis, and Neumann]{Abdolmaleki2015}
Abbas Abdolmaleki, Rudolf Lioutikov, Jan~R Peters, Nuno Lau, Luis Pualo~Reis,
  and Gerhard Neumann.
\newblock Model-based relative entropy stochastic search.
\newblock \emph{Advances in Neural Information Processing Systems}, 28, 2015.

\bibitem[Abdolmaleki et~al.(2019)Abdolmaleki, Sim{\~o}es, Lau, Reis, and
  Neumann]{Abdolmaleki2019}
Abbas Abdolmaleki, David Sim{\~o}es, Nuno Lau, Lu{\'\i}s~Paulo Reis, and
  Gerhard Neumann.
\newblock Contextual direct policy search.
\newblock \emph{Journal of Intelligent \& Robotic Systems}, 96\penalty0
  (2):\penalty0 141--157, 2019.

\bibitem[Agarwal et~al.(2021)Agarwal, Schwarzer, Castro, Courville, and
  Bellemare]{Agarwal2021}
Rishabh Agarwal, Max Schwarzer, Pablo~Samuel Castro, Aaron Courville, and
  Marc~G Bellemare.
\newblock Deep reinforcement learning at the edge of the statistical precipice.
\newblock \emph{Advances in Neural Information Processing Systems}, 2021.

\bibitem[Agrawal et~al.(2019)Agrawal, Amos, Barratt, Boyd, Diamond, and
  Kolter]{agrawal2019differentiable}
Akshay Agrawal, Brandon Amos, Shane Barratt, Stephen Boyd, Steven Diamond, and
  J~Zico Kolter.
\newblock Differentiable convex optimization layers.
\newblock \emph{Advances in neural information processing systems}, 32, 2019.

\bibitem[Akrour et~al.(2019)Akrour, Pajarinen, Peters, and Neumann]{Akrour2019}
Riad Akrour, Joni Pajarinen, Jan Peters, and Gerhard Neumann.
\newblock Projections for approximate policy iteration algorithms.
\newblock In \emph{Proceedings of Machine Learning Research}, pages 181--190,
  2019.

\bibitem[Bahl et~al.(2020)Bahl, Mukadam, Gupta, and Pathak]{Bahl2020}
Shikhar Bahl, Mustafa Mukadam, Abhinav Gupta, and Deepak Pathak.
\newblock Neural dynamic policies for end-to-end sensorimotor learning.
\newblock \emph{Advances in Neural Information Processing Systems}, 12 2020.
\newblock URL \url{https://arxiv.org/abs/2012.02788v1}.

\bibitem[Berner et~al.(2019)Berner, Brockman, Chan, Cheung, Debiak, Dennison,
  Farhi, Fischer, Hashme, Hesse, et~al.]{berner2019dota}
Christopher Berner, Greg Brockman, Brooke Chan, Vicki Cheung, Przemyslaw
  Debiak, Christy Dennison, David Farhi, Quirin Fischer, Shariq Hashme, Chris
  Hesse, et~al.
\newblock Dota 2 with large scale deep reinforcement learning.
\newblock \emph{arXiv preprint arXiv:1912.06680}, 2019.

\bibitem[Brandherm et~al.(2019)Brandherm, Peters, Neumann, and
  Akrour]{brandherm2019learning}
Florian Brandherm, Jan Peters, Gerhard Neumann, and Riad Akrour.
\newblock Learning replanning policies with direct policy search.
\newblock \emph{IEEE Robotics and Automation Letters}, 4\penalty0 (2):\penalty0
  2196--2203, 2019.

\bibitem[Braylan et~al.(2015)Braylan, Hollenbeck, Meyerson, and
  Miikkulainen]{braylan2015frame}
Alex Braylan, Mark Hollenbeck, Elliot Meyerson, and Risto Miikkulainen.
\newblock Frame skip is a powerful parameter for learning to play atari.
\newblock In \emph{Workshops at the Twenty-Ninth AAAI Conference on Artificial
  Intelligence}, 2015.

\bibitem[Brockman et~al.(2016)Brockman, Cheung, Pettersson, Schneider,
  Schulman, Tang, and Zaremba]{Brockman2016}
Greg Brockman, Vicki Cheung, Ludwig Pettersson, Jonas Schneider, John Schulman,
  Jie Tang, and Wojciech Zaremba.
\newblock Openai gym, 2016.

\bibitem[Celik et~al.(2022)Celik, Zhou, Li, Becker, and
  Neumann]{celik2022specializing}
Onur Celik, Dongzhuoran Zhou, Ge~Li, Philipp Becker, and Gerhard Neumann.
\newblock Specializing versatile skill libraries using local mixture of
  experts.
\newblock In \emph{Conference on Robot Learning}, pages 1423--1433. PMLR, 2022.

\bibitem[Chrabaszcz et~al.(2018)Chrabaszcz, Loshchilov, and
  Hutter]{Chrabaszcz2018}
Patryk Chrabaszcz, Ilya Loshchilov, and Frank Hutter.
\newblock Back to basics: Benchmarking canonical evolution strategies for
  playing atari.
\newblock In \emph{IJCAI}, 2018.

\bibitem[Dalal et~al.(2021)Dalal, Pathak, and
  Salakhutdinov]{dalal2021accelerating}
Murtaza Dalal, Deepak Pathak, and Russ~R Salakhutdinov.
\newblock Accelerating robotic reinforcement learning via parameterized action
  primitives.
\newblock \emph{Advances in Neural Information Processing Systems},
  34:\penalty0 21847--21859, 2021.

\bibitem[Daniel et~al.(2012)Daniel, Neumann, and Peters]{Daniel2012}
Christian Daniel, Gerhard Neumann, and Jan Peters.
\newblock Hierarchical relative entropy policy search.
\newblock In \emph{Artificial Intelligence and Statistics}, pages 273--281.
  PMLR, 2012.

\bibitem[Deisenroth et~al.(2013)Deisenroth, Neumann, Peters,
  et~al.]{Deisenroth2013}
Marc~Peter Deisenroth, Gerhard Neumann, Jan Peters, et~al.
\newblock A survey on policy search for robotics.
\newblock \emph{Foundations and trends in Robotics}, 2\penalty0 (1-2):\penalty0
  388--403, 2013.

\bibitem[Engstrom et~al.(2020)Engstrom, Ilyas, Santurkar, Tsipras, Janoos,
  Rudolph, and Madry]{Engstrom2020}
Logan Engstrom, Andrew Ilyas, Shibani Santurkar, Dimitris Tsipras, Firdaus
  Janoos, Larry Rudolph, and Aleksander Madry.
\newblock {Implementation Matters in Deep Policy Gradients: A Case Study on PPO
  and TRPO}.
\newblock In \emph{International Conference on Learning Representations}, 2020.
\newblock URL \url{http://arxiv.org/abs/2005.12729}.

\bibitem[Ginesi et~al.(2019)Ginesi, Meli, Nakawala, Roberti, and
  Fiorini]{ginesi2019knowledge}
Michele Ginesi, Daniele Meli, Hirenkumar Nakawala, Andrea Roberti, and Paolo
  Fiorini.
\newblock A knowledge-based framework for task automation in surgery.
\newblock In \emph{2019 19th International Conference on Advanced Robotics
  (ICAR)}, pages 37--42. IEEE, 2019.

\bibitem[Gomez-Gonzalez et~al.(2016)Gomez-Gonzalez, Neumann, Sch{\"o}lkopf, and
  Peters]{gomez2016using}
Sebastian Gomez-Gonzalez, Gerhard Neumann, Bernhard Sch{\"o}lkopf, and Jan
  Peters.
\newblock Using probabilistic movement primitives for striking movements.
\newblock In \emph{2016 IEEE-RAS 16th International Conference on Humanoid
  Robots (Humanoids)}, pages 502--508. IEEE, 2016.

\bibitem[Haarnoja et~al.(2018)Haarnoja, Zhou, Abbeel, and Levine]{Haarnoja2018}
Tuomas Haarnoja, Aurick Zhou, Pieter Abbeel, and Sergey Levine.
\newblock Soft actor-critic: Off-policy maximum entropy deep reinforcement
  learning with a stochastic actor.
\newblock In \emph{International conference on machine learning}, pages
  1861--1870. PMLR, 2018.

\bibitem[Hansen and Ostermeier(2001)]{Hansen2001}
N.~Hansen and A.~Ostermeier.
\newblock Completely derandomized self-adaptation in evolution strategies.
\newblock \emph{Evolutionary computation}, 9\penalty0 (2):\penalty0 159--195,
  2001.

\bibitem[Ijspeert et~al.(2013)Ijspeert, Nakanishi, Hoffmann, Pastor, and
  Schaal]{Ijspeert2013}
Auke~Jan Ijspeert, Jun Nakanishi, Heiko Hoffmann, Peter Pastor, and Stefan
  Schaal.
\newblock Dynamical {{Movement Primitives}}: {{Learning Attractor Models}} for
  {{Motor Behaviors}}.
\newblock \emph{Neural Computation}, 25\penalty0 (2), 2013.

\bibitem[Kober and Peters(2008)]{Kober2008}
Jens Kober and Jan Peters.
\newblock Policy search for motor primitives in robotics.
\newblock In D.~Koller, D.~Schuurmans, Y.~Bengio, and L.~Bottou, editors,
  \emph{Advances in Neural Information Processing Systems}, volume~21. Curran
  Associates, Inc., 2008.
\newblock URL
  \url{https://proceedings.neurips.cc/paper/2008/file/7647966b7343c29048673252e490f736-Paper.pdf}.

\bibitem[Kormushev et~al.(2010)Kormushev, Calinon, and
  Caldwell]{kormushev2010robot}
Petar Kormushev, Sylvain Calinon, and Darwin~G Caldwell.
\newblock Robot motor skill coordination with em-based reinforcement learning.
\newblock In \emph{2010 IEEE/RSJ international conference on intelligent robots
  and systems}, pages 3232--3237. IEEE, 2010.

\bibitem[Kormushev et~al.(2013)Kormushev, Calinon, and
  Caldwell]{kormushev2013reinforcement}
Petar Kormushev, Sylvain Calinon, and Darwin~G Caldwell.
\newblock Reinforcement learning in robotics: Applications and real-world
  challenges.
\newblock \emph{Robotics}, 2\penalty0 (3):\penalty0 122--148, 2013.

\bibitem[Kupcsik et~al.(2017)Kupcsik, Deisenroth, Peters, Poha, Vadakkepata,
  and Neumann]{Kupcsik2017}
Andras Kupcsik, Marc~P. Deisenroth, Jan Peters, Loh~Ai Poha, Prahlad
  Vadakkepata, and Gerhard Neumann.
\newblock Model-based contextual policy search for data-efficient
  generalization of robot skills.
\newblock \emph{Artificial Intelligence}, 247:\penalty0 415--439, 2017.
\newblock \doi{10.1016/j.artint.2014.11.005}.
\newblock URL \url{http://eprints.lincoln.ac.uk/25774/1/Kupcsik_AIJ_2015.pdf}.
\newblock Impact Factor: 3.333.

\bibitem[Lee et~al.(2020)Lee, Seo, and Kim]{lee2020trajectory}
Hyeonbeom Lee, Hoseong Seo, and Hyeong-Geun Kim.
\newblock Trajectory optimization and replanning framework for a micro air
  vehicle in cluttered environments.
\newblock \emph{Ieee Access}, 8:\penalty0 135406--135415, 2020.

\bibitem[Li et~al.(2022)Li, Jin, Volpp, Otto, Lioutikov, and
  Neumann]{li2022prodmps}
Ge~Li, Zeqi Jin, Michael Volpp, Fabian Otto, Rudolf Lioutikov, and Gerhard
  Neumann.
\newblock Prodmps: A unified perspective on dynamic and probabilistic movement
  primitives.
\newblock \emph{arXiv preprint arXiv:2210.01531}, 2022.

\bibitem[Li et~al.(2017)Li, Zhao, Chen, Hu, Su, and
  Fukuda]{li2017reinforcement}
Zhijun Li, Ting Zhao, Fei Chen, Yingbai Hu, Chun-Yi Su, and Toshio Fukuda.
\newblock Reinforcement learning of manipulation and grasping using dynamical
  movement primitives for a humanoidlike mobile manipulator.
\newblock \emph{IEEE/ASME Transactions on Mechatronics}, 23\penalty0
  (1):\penalty0 121--131, 2017.

\bibitem[Maeda et~al.(2014)Maeda, Ewerton, Lioutikov, Amor, Peters, and
  Neumann]{maeda2014learning}
Guilherme Maeda, Marco Ewerton, Rudolf Lioutikov, Heni~Ben Amor, Jan Peters,
  and Gerhard Neumann.
\newblock Learning interaction for collaborative tasks with probabilistic
  movement primitives.
\newblock In \emph{2014 IEEE-RAS International Conference on Humanoid Robots},
  pages 527--534. IEEE, 2014.

\bibitem[Mania et~al.(2018)Mania, Guy, and Recht]{Mania2018}
Horia Mania, Aurelia Guy, and Benjamin Recht.
\newblock Simple random search of static linear policies is competitive for
  reinforcement learning.
\newblock In S.~Bengio, H.~Wallach, H.~Larochelle, K.~Grauman, N.~Cesa-Bianchi,
  and R.~Garnett, editors, \emph{Advances in Neural Information Processing
  Systems}, volume~31. Curran Associates, Inc., 2018.
\newblock URL
  \url{https://proceedings.neurips.cc/paper/2018/file/7634ea65a4e6d9041cfd3f7de18e334a-Paper.pdf}.

\bibitem[Mannor et~al.(2003)Mannor, Rubinstein, and Gat]{Mannor2003}
Shie Mannor, Reuven~Y Rubinstein, and Yohai Gat.
\newblock The cross entropy method for fast policy search.
\newblock In \emph{Proceedings of the 20th International Conference on Machine
  Learning (ICML-03)}, pages 512--519, 2003.

\bibitem[M{\"u}lling et~al.(2013)M{\"u}lling, Kober, Kroemer, and
  Peters]{mulling2013learning}
Katharina M{\"u}lling, Jens Kober, Oliver Kroemer, and Jan Peters.
\newblock Learning to select and generalize striking movements in robot table
  tennis.
\newblock \emph{The International Journal of Robotics Research}, 32\penalty0
  (3):\penalty0 263--279, 2013.

\bibitem[OpenAI(2023)]{openai2023gpt}
R~OpenAI.
\newblock Gpt-4 technical report.
\newblock \emph{arXiv}, 2023.

\bibitem[Otto et~al.(2021)Otto, Becker, Anh~Vien, Ziesche, and
  Neumann]{Otto2021}
Fabian Otto, Philipp Becker, Ngo Anh~Vien, Hanna~Carolin Ziesche, and Gerhard
  Neumann.
\newblock Differentiable trust region layers for deep reinforcement learning.
\newblock In \emph{International Conference on Learning Representations}, 2021.

\bibitem[Otto et~al.(2022)Otto, Celik, Zhou, Ziesche, Vien, and
  Neumann]{otto2022deep}
Fabian Otto, Onur Celik, Hongyi Zhou, Hanna Ziesche, Ngo~Anh Vien, and Gerhard
  Neumann.
\newblock Deep black-box reinforcement learning with movement primitives.
\newblock \emph{arXiv preprint arXiv:2210.09622}, 2022.

\bibitem[Pahič et~al.(2018)Pahič, Gams, Ude, and Morimoto]{gams2018deep}
Rok Pahič, Andrej Gams, Ale{\v{s}} Ude, and Jun Morimoto.
\newblock Deep encoder-decoder networks for mapping raw images to dynamic
  movement primitives.
\newblock In \emph{2018 IEEE International Conference on Robotics and
  Automation}, 2018.

\bibitem[Pahič et~al.(2020)Pahič, Ridge, Gams, Morimoto, and
  Ude]{ridge2020training}
Rok Pahič, Barry Ridge, Andrej Gams, Jun Morimoto, and Ale{\v{s}} Ude.
\newblock Training of deep neural networks for the generation of dynamic
  movement primitives.
\newblock \emph{Neural Networks}, 2020.

\bibitem[Paraschos et~al.(2013)Paraschos, Daniel, Peters, and
  Neumann]{Paraschos2013}
Alexandros Paraschos, Christian Daniel, Jan~R Peters, and Gerhard Neumann.
\newblock Probabilistic movement primitives.
\newblock In C.~J.~C. Burges, L.~Bottou, M.~Welling, Z.~Ghahramani, and K.~Q.
  Weinberger, editors, \emph{Advances in Neural Information Processing
  Systems}, volume~26. Curran Associates, Inc., 2013.
\newblock URL
  \url{https://proceedings.neurips.cc/paper/2013/file/e53a0a2978c28872a4505bdb51db06dc-Paper.pdf}.

\bibitem[Rozo and Dave(2022)]{rozo2022orientation}
Leonel Rozo and Vedant Dave.
\newblock Orientation probabilistic movement primitives on riemannian
  manifolds.
\newblock In \emph{Conference on Robot Learning}, pages 373--383. PMLR, 2022.

\bibitem[Salimans et~al.(2017)Salimans, Ho, Chen, Sidor, and
  Sutskever]{Salimans2017}
Tim Salimans, Jonathan Ho, Xi~Chen, Szymon Sidor, and Ilya Sutskever.
\newblock Evolution strategies as a scalable alternative to reinforcement
  learning.
\newblock \emph{arXiv preprint arXiv:1703.03864}, 2017.

\bibitem[Schaal(2006)]{Schaal2006}
Stefan Schaal.
\newblock \emph{Dynamic Movement Primitives -A Framework for Motor Control in
  Humans and Humanoid Robotics}, pages 261--280.
\newblock Springer Tokyo, Tokyo, 2006.
\newblock ISBN 978-4-431-31381-6.
\newblock \doi{10.1007/4-431-31381-8_23}.
\newblock URL \url{https://doi.org/10.1007/4-431-31381-8_23}.

\bibitem[Schaal et~al.(2005)Schaal, Peters, Nakanishi, and
  Ijspeert]{Schaal2005}
Stefan Schaal, Jan Peters, Jun Nakanishi, and Auke Ijspeert.
\newblock Learning movement primitives.
\newblock In \emph{Robotics research. the eleventh international symposium},
  pages 561--572. Springer, 2005.

\bibitem[Schulman et~al.(2015)Schulman, Levine, Abbeel, Jordan, and
  Moritz]{Schulman2015}
John Schulman, Sergey Levine, Pieter Abbeel, Michael Jordan, and Philipp
  Moritz.
\newblock {Trust Region Policy Optimization}.
\newblock In \emph{Proceedings of Machine Learning Research}, pages 1889--1897,
  2015.
\newblock URL \url{http://proceedings.mlr.press/v37/schulman15.html}.

\bibitem[Schulman et~al.(2016)Schulman, Moritz, Levine, Jordan, and
  Abbeel]{Schulman2016}
John Schulman, Philipp Moritz, Sergey Levine, Michael Jordan, and Pieter
  Abbeel.
\newblock High-dimensional continuous control using generalized advantage
  estimation.
\newblock In \emph{Proceedings of the International Conference on Learning
  Representations (ICLR)}, 2016.

\bibitem[Schulman et~al.(2017)Schulman, Wolski, Dhariwal, Radford, and
  Klimov]{Schulman2017}
John Schulman, Filip Wolski, Prafulla Dhariwal, Alec Radford, and Oleg Klimov.
\newblock {Proximal Policy Optimization Algorithms}.
\newblock In \emph{arXiv preprint}, 2017.
\newblock URL \url{http://arxiv.org/abs/1707.06347}.

\bibitem[Sehnke et~al.(2010)Sehnke, Osendorfer, R{\"u}ckstiess, Graves, Peters,
  and Schmidhuber]{Sehnke2010}
F.~Sehnke, C.~Osendorfer, T.~R{\"u}ckstiess, A.~Graves, J.~Peters, and
  J.~Schmidhuber.
\newblock Parameter-exploring policy gradients.
\newblock \emph{Neural Networks}, 21\penalty0 (4):\penalty0 551--559, May 2010.
\newblock \doi{10.1016/j.neunet.2009.12.004}.

\bibitem[Stulp and Sigaud(2012{\natexlab{a}})]{Stulp2012}
Freek Stulp and Olivier Sigaud.
\newblock Path integral policy improvement with covariance matrix adaptation.
\newblock In \emph{Proceedings of the 29th International Coference on
  International Conference on Machine Learning}, ICML'12, page 1547–1554,
  Madison, WI, USA, 2012{\natexlab{a}}. Omnipress.
\newblock ISBN 9781450312851.

\bibitem[Stulp and Sigaud(2012{\natexlab{b}})]{Stulp2012b}
Freek Stulp and Olivier Sigaud.
\newblock Policy improvement methods: Between black-box optimization and
  episodic reinforcement learning.
\newblock 2012{\natexlab{b}}.

\bibitem[Sutton et~al.(1999)Sutton, McAllester, Singh, and Mansour]{Sutton1999}
Richard~S Sutton, David McAllester, Satinder Singh, and Yishay Mansour.
\newblock Policy gradient methods for reinforcement learning with function
  approximation.
\newblock In S.~Solla, T.~Leen, and K.~M\"{u}ller, editors, \emph{Advances in
  Neural Information Processing Systems}, volume~12. MIT Press, 1999.
\newblock URL
  \url{https://proceedings.neurips.cc/paper/1999/file/464d828b85b0bed98e80ade0a5c43b0f-Paper.pdf}.

\bibitem[Tangkaratt et~al.(2017)Tangkaratt, van Hoof, Parisi, Neumann, Peters,
  and Sugiyama]{Tangkaratt2017}
Voot Tangkaratt, Herke van Hoof, Simone Parisi, Gerhard Neumann, Jan Peters,
  and Masashi Sugiyama.
\newblock Policy search with high-dimensional context variables.
\newblock In \emph{Proceedings of the AAAI Conference on Artificial
  Intelligence}, volume~31, 2017.

\bibitem[Wierstra et~al.(2014)Wierstra, Schaul, Glasmachers, Sun, Peters, and
  Schmidhuber]{Wierstra2014}
Daan Wierstra, Tom Schaul, Tobias Glasmachers, Yi~Sun, Jan Peters, and
  J{\"u}rgen Schmidhuber.
\newblock Natural evolution strategies.
\newblock \emph{The Journal of Machine Learning Research}, 15\penalty0
  (1):\penalty0 949--980, 2014.

\bibitem[Williams(1992)]{Williams1992}
Ronald~J Williams.
\newblock Simple statistical gradient-following algorithms for connectionist
  reinforcement learning.
\newblock \emph{Machine learning}, 8\penalty0 (3):\penalty0 229--256, 1992.

\bibitem[Yu et~al.(2019)Yu, Quillen, He, Julian, Hausman, Finn, and
  Levine]{Yu2019}
Tianhe Yu, Deirdre Quillen, Zhanpeng He, Ryan Julian, Karol Hausman, Chelsea
  Finn, and Sergey Levine.
\newblock Meta-world: A benchmark and evaluation for multi-task and meta
  reinforcement learning.
\newblock In \emph{Conference on Robot Learning (CoRL)}, 2019.
\newblock URL \url{https://arxiv.org/abs/1910.10897}.

\bibitem[Zenkri et~al.(2022)Zenkri, Vien, and Neumann]{zenkri2022hierarchical}
Oussama Zenkri, Ngo~Anh Vien, and Gerhard Neumann.
\newblock Hierarchical policy learning for mechanical search.
\newblock In \emph{2022 International Conference on Robotics and Automation
  (ICRA)}, pages 1954--1960. IEEE, 2022.

\end{thebibliography}


\appendix
\newpage
\section{Derivations of \acrlong{pdmp}.}
\label{app:pdmp}
In this section, we will briefly present the main derivations of \acrshortpl{pdmp}. 
We start with the fundamental aspects of \acrshortpl{dmp} and then derive \acrshortpl{pdmp} from the analytical solution of the \acrshortpl{dmp}' \acrshort{ode}. 
Finally, we present the solution to a initial value problem of the \acrshort{ode}, which allows us to perform smooth replanning during trajectory execution in a computationally efficient manner. 
For the sake of simplicity, we introduce the approach by means of a 1-\acrshort{dof} dynamical system. For higher \acrshort{dof} systems, we refer to the original paper by \cite{li2022prodmps}.

\paragraph{\acrlong{dmp}} 
\citet{Schaal2006, Ijspeert2013} model a single movement execution as a trajectory $\bm{\lambda} = [y_t]_{t=0:T}$ using a second-order linear dynamical system with a non-linear forcing function $f$,
\begin{equation}
    \tau^2\ddot{y} = \alpha(\beta(g-y)-\tau\dot{y})+ f(x), \quad f(x) = x\frac{\sum\varphi_i(x)w_i}{\sum\varphi_i(x)} = x\bm{\varphi}_x^\intercal\bm{w},
    \label{eq:dmp_original}
\end{equation}
where $y = y(t),~\dot{y}=\mathrm{d}y/\mathrm{d}t,~\ddot{y} =\mathrm{d}^2y/\mathrm{d}t^2$ represent the position, velocity, and acceleration of the system at time step $t$, respectively.
$\alpha$ and $\beta$ are spring-damper constants, 
$g$ is a goal attractor, and $\tau$ is a time constant that can be used to adjust the execution speed of the resulting trajectory. To achieve goal convergence, \acrshortpl{dmp} define the forcing function based on an exponentially decaying phase variable $x(t)=\exp(-\alpha_x/\tau \; t)$, where $\varphi_i(x)$ represents the (unnormalized) basis functions. The shape of the trajectory as it converges to the goal is controlled by the weights $w_i \in \bm{w}$, $i=1...N$.
The trajectory of the motion $\bm{\lambda}$ is obtained by integrating the system numerically from the starting time to the target time point. However, this process is often computationally expensive.

\paragraph{Solving the \acrlongpl{dmp}' underlying \acrshort{ode}.}
\citeauthor{li2022prodmps} recognize that the governing equation of \acrshortpl{dmp}, given in Eq.(\ref{eq:dmp_original}), has an analytical solution, as it is a second-order linear non-homogeneous \acrshort{ode} with constant coefficients. To better convey this method, the \acrshort{ode} and its homogeneous counterpart can be rewritten in a standard form as:
{
\begin{align}
    \textbf{Non-homo. ODE:} ~~~\ddot{y} + \frac{\alpha}{\tau}\dot{y} + \frac{\alpha\beta}{\tau^2} y &=\frac{f(x)}{\tau^2}+\frac{\alpha \beta}{\tau^2} g \equiv F(x, g), \label{eq:dmp_non_homo}\\
    \textbf{Homo. ODE:}~~~\ddot{y} + \frac{\alpha}{\tau}\dot{y} + \frac{\alpha\beta}{\tau^2} y &= 0.
    \label{eq:dmp_homo}
\end{align}
}%
With appropriate configuration of the spring-damper coefficients, \ie, $\beta = \alpha/4$ (\cite{Schaal2006, Ijspeert2013}), the system is critically damped and the motion generated by the \acrshortpl{dmp} will settle to the target position smoothly and efficiently. The analytical solution of Eq. (\ref{eq:dmp_non_homo}) in this case takes the form
\begin{align}
    y = c_1y_1 + c_2y_2 -y_1 \int\frac{y_2F}{Y}\mathrm{d}t + y_2\int\frac{y_1F}{Y}\mathrm{d}t, \quad Y = y_1 \dot{y}_2 - \dot{y}_1y_2,
    \label{eq:dmp_closed_form_position}\\
    y_1 = y_1(t) = \exp\left(-\frac{\alpha}{2\tau}t\right), \quad\quad
    y_2 = y_2(t) = t\exp\left(-\frac{\alpha}{2\tau}t\right),
    \label{eq:dmp_y1_y2}
\end{align}
where $y_1$ and $y_2$ are the complementary functions of the homogeneous function in Eq. (\ref{eq:dmp_homo}) and $\dot{y}_1$, $\dot{y}_2$ their corresponding derivatives w.r.t. time.
By utilizing the fundamental of calculus, which states that $\int h(t)\mathrm{d}t = \int_0^th(t')\mathrm{d}t' + c$, where $c \in \mathbb R$ is a constant, the two indefinite integrals in Eq. (\ref{eq:dmp_closed_form_position}) can be transformed into two definite integrals.
During this transformation, the learnable parameters $\bm{w}$ and $g$ which control the shape of the trajectory, can be extracted from the resulting definite integrals.
Finally, the trajectory position and velocity can be expressed in a compact matrix form as
\begin{align}
    y &= c_1y_1 + c_2y_2 + \begin{bmatrix}
                            y_2\bm{p_2}-y_1\bm{p_1} & y_2q_2 - y_1q_1
                          \end{bmatrix}\begin{bmatrix}
                            \bm{w}\\g
                          \end{bmatrix} 
         \label{eq:prodmp_pos}\\
    \dot{y} &= c_1\dot{y}_1 + c_2\dot{y}_2 + \begin{bmatrix}
                            \dot{y}_2\bm{p_2}-\dot{y}_1\bm{p_1} & \dot{y}_2q_2 - \dot{y}_1q_1
                            \end{bmatrix}\begin{bmatrix}\vspace{-0.0cm}\bm{w}\\g
                            \end{bmatrix}
        \label{eq:prodmp_vel},
\end{align}
where $\bm{p}_1$, $\bm{p}_1$, $q_1$, $q_2$ represent the elements used to formulate the definite integrals in the matrix form, as
\begin{align}
    \bm{p}_1(t) = \frac{1}{\tau^2}\int_0^t t'\exp\Big(\frac{\alpha}{2\tau}t'\Big)x(t')\bm{\varphi}_x^\intercal \mathrm{d}t', \quad &
    \bm{p}_2(t) = \frac{1}{\tau^2}\int_0^t \exp\Big(\frac{\alpha}{2\tau}t'\Big)x(t')\bm{\varphi}_x^\intercal \mathrm{d}t',\label{eq:dmp_p1_p2}\\
    q_1(t) = \Big(\frac{\alpha}{2\tau}t  - 1\Big)\exp\Big(\frac{\alpha}{2\tau}t\Big) +1,
    \quad & q_2(t) = \frac{\alpha}{2\tau} \bigg[\exp\Big(\frac{\alpha}{2\tau}t\Big)-1\bigg]. \label{eq:dmp_q1_q2}
\end{align}
It is worth noting that, despite the closed form solution for $q_1$ and $q_2$, $\bm{p}_1$ and $\bm{p}_2$ cannot be obtained analytically because of the complex nature of the $\bm{\varphi}_x$. As a result, they must be computed numerically. 
However, the extraction of the learnable parameters $\bm{w}$ and $g$ from the integrals in Eq. (\ref{eq:prodmp_pos}) and (\ref{eq:prodmp_vel}) enables the sharing of the remaining integrals among all trajectories to be generated. 
In other words, these integrals can be pre-computed offline and used as constant functions during online trajectory computation, which significantly simplifies the trajectory generation procedure and speeds it up. 
These remaining integrals are referred to as the position basis $\ipb(t)$ and velocity basis $\ivb(t)$, and the \acrshortpl{pdmp} represent the position and velocity in a similar manner of \acrshortpl{promp} as:
\begin{equation}
    y(t) = c_1y_1(t) + c_2y_2(t) + \ipb(t)^\intercal \bm{w}_g, \quad \dot{y}(t)=c_1\dot{y}_1(t) + c_2\dot{y}_2(t) + \ivb(t)^\intercal\bm{w}_g, \tag{\ref{eq:prodmp}}
\end{equation}
where $\bm{w}_g$ is a concatenation vector containing $\bm{w}$ and $g$.
\paragraph{Solve the initial value problem.}
To compute the coefficients $c_1$ and $c_2$, a solution to the initial value problem represented by the Eq.(\ref{eq:prodmp}) must be found. 
\citeauthor{li2022prodmps} suggest using the current robot state, which consists of the robot's position and velocity ($y_b, \dot{y}_b$) at the replanning time step $t_b$, as the natural condition for ensuring a smooth transition between the previous and newly generated trajectory.   
We denote the values of the complementary functions and their derivatives at time $t_b$ as $y_{1_b}, y_{2_b}, \dot{y}_{1_b} \dot{y}_{2_b}$, and the values of the position and velocity basis functions as $\ipbb, \ivbb$.
By substituting these values into Eq.(\ref{eq:prodmp}), $c_1$ and $c_2$ can be calculated as:
\begin{equation}
    \begin{bmatrix}c_1\\c_2\end{bmatrix}
    = \begin{bmatrix}\frac{\dot{y}_{2_b}y_b-y_{2_b}\dot{y}_b}{y_{1_b}\dot{y}_{2_b}-y_{2_b}\dot{y}_{1_b}} +\frac{y_{2_b}\ivbbt-\dot{y}_{2_b}\ipbbt}{y_{1_b}\dot{y}_{2_b}-y_{2_b}\dot{y}_{1_b}}\bm{w}_g\\[0.7em]
    \frac{y_{1_b}\dot{y}_b-\dot{y}_{1_b}y_b}{y_{1_b}\dot{y}_{2_b}-y_{2_b}\dot{y}_{1_b}}+\frac{\dot{y}_{1_b}\ipbbt- y_{1_b}\ivbbt}{y_{1_b}\dot{y}_{2_b}-y_{2_b}\dot{y}_{1_b}}\bm{w}_g\end{bmatrix}.
    \label{eq:dmp_c}
\end{equation}

\newpage

\section{Trust Region Projection Layers with KL-Divergence}
\label{app:kl_trpl}
As already mentioned in the main text, TRPLs \cite{Otto2021} present a scalable and mathematically sound approach for enforcing trust regions in step-based deep RL.
The layer takes the output of a standard Gaussian policy as input in terms of mean $\bm{\mu}$ and variance $\bm{\Sigma}$ and projects it into the trust region if the given mean and variance violate their respective bounds. 
This projection is done for each input state individually. 
Subsequently, the projected Gaussian policy distribution with parameters $\tilde{\bm{\mu}}$, $\tilde{\bm{\Sigma}}$ is used for any further steps, e.\,g. for sampling and/or loss computation.
Formally, the layer solves the following two optimization problems for each state $\bm{s}$
    \begin{align}
        \argmin_{\til{\bm{\mu}}_s} d_\textrm{mean} \left(\til{\bm{\mu}}_s, \bm{\mu}(s) \right),  \quad &\st \quad d_\textrm{mean} \left(\til{\bm{\mu}}_s,  \old{\bm{\mu}}(s) \right) \leq \epsilon_{\bm{\mu}}, \quad \textrm{and} 
        \label{eq:apx:generic_mean}
        \\ 
        \argmin_{\til{\bm{\Sigma}}_s} d_\textrm{cov} \left(\til{\bm{\Sigma}}_s, \bm{\Sigma}(\bm{s}) \right), \quad &\st \quad d_\textrm{cov} \left(\til{\bm{\Sigma}}_s, \old{\bm{\Sigma}}(\bm{s}) \right) \leq \epsilon_\Sigma,`
        \label{eq:apx:generic_cov}
    \end{align}
where $\tilde{\bm{\mu}}_s$ and $\tilde{ \bm{\Sigma}}_s$ are the optimization variables for input state $\bm{s}$ and $\epsilon_\mu$ and $\epsilon_\Sigma$ are the trust region bounds for mean and covariance, respectively.
Finally, $\old{\mu}$ and $\old{\Sigma}$ are the reference mean and covariance for the trust region and $d_\textrm{mean}$ as well as $d_\textrm{cov}$ are the similarity metrics for the mean and covariance of a decomposable distance or divergence measure. 
As we only leverage the KL-divergence projection, we will provide only details for this particular projection below. 
For the other two projections we refer the reader to \citet{Otto2021}.

Inserting the mean part of the Gaussian KL divergence into \Eqref{eq:apx:generic_mean} yields
\begin{equation*}
    	\argmin_{\til{\mu}} \vecT{\left(\mu - \til{\mu}\right)} \oldInv{\Sigma} \left(\mu - \til{\mu}\right) 
    	\quad \st \quad \vecT{\left(\old{\mu} - \til{\mu}\right)} \oldInv{\Sigma} \left(\old{\mu} - \til{\mu}\right) \leq \epsilon_\mu.
\end{equation*}

After differentiating the dual w.r.t. $\til{\mu}$, we can solve for the projected mean 
\begin{align*}
\til{\mu} &= \frac{\mu + \omega  \old{\mu}}{1 + \omega} \quad \text{with} \quad \omega = \sqrt{\frac{\vecT{\left(\old{\mu} - \mu\right)} \oldInv{\Sigma} \left(\old{\mu} - \mu\right)}{\epsilon_\mu}} - 1,  
\end{align*}
leveraging the optimal Lagrange multiplier $\omega$.
Similarly, we can insert the covariance part of the Gaussian KL divergence into \Eqref{eq:apx:generic_cov}, which results in
\begin{align*}
	\argmin_{\til{\Sigma}}  \textrm{tr}\left(\inv{\Sigma}\til{\Sigma}\right) + \log \frac{|\Sigma|}{|\til{\Sigma}|}, \quad
	\st \quad \textrm{tr}\left(\oldInv{\Sigma} \til{\Sigma} \right) - d + \log\frac{|\old{\Sigma}|}{|\til{\Sigma}|} \leq \epsilon_\Sigma,
\end{align*}
where $d$ is the number of degrees of freedom (DoF).
Once again, differentiating and solving the dual $g(\eta)$ for the projected covariance yields
\begin{align*}
    \tilde{\Sigma} = \left( \dfrac{\eta^* \oldInv{\Sigma} + \Sigma^{-1}}{\eta^* + 1 } \right)^{-1} \quad \text{with} \quad \eta^* = \argmin_{\eta} g(\eta), \; \st \; \eta \ge 0.
    \label{eq:kl_proj_cov}
\end{align*}
Here, the the optimal Lagrange multiplier $\eta^*$ cannot be computed in closed form, however, a standard numerical optimizer, such as BFGS, is able to efficiently find it. 
This can be made differentiable by taking the differentials of the KKT conditions of the dual. 
For more details, we refer to the original work \citep{Otto2021}.

\section{Environment Details}
\label{app:details}
\begin{figure}
    \centering
    \includegraphics[width=0.25\textwidth]{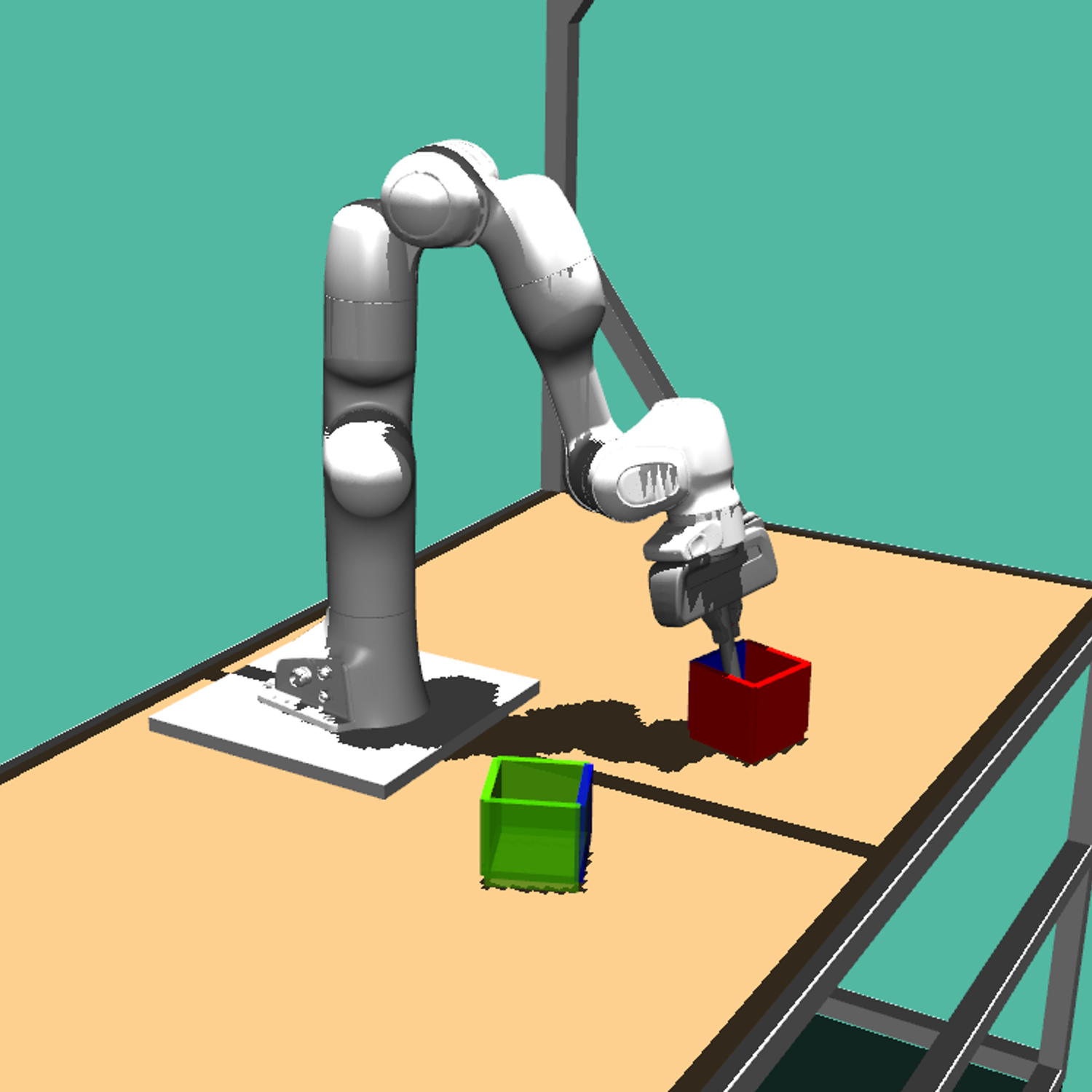}%
    \includegraphics[width=0.25\textwidth]{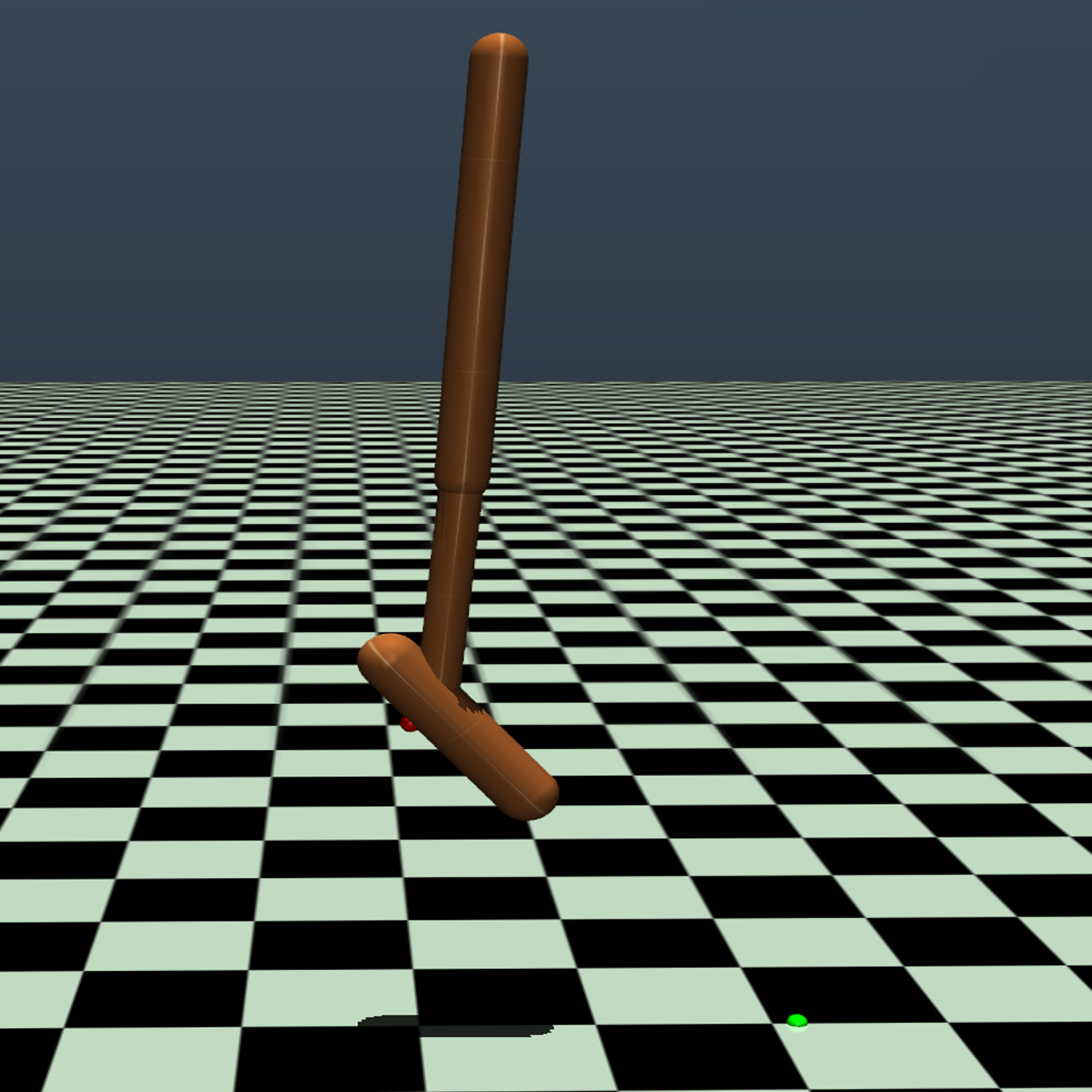}%
    \includegraphics[width=0.25\textwidth]{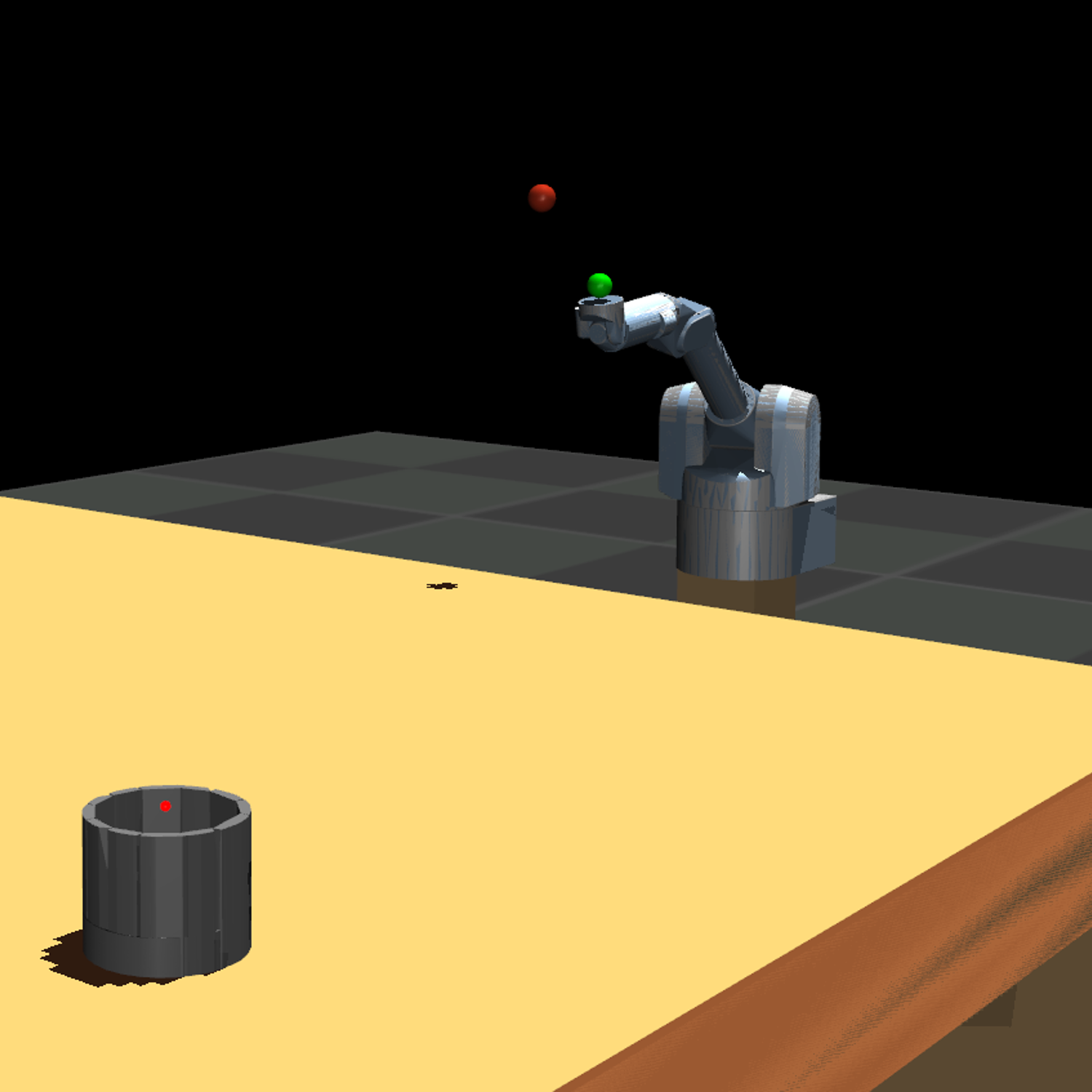}%
    \includegraphics[width=0.25\textwidth]{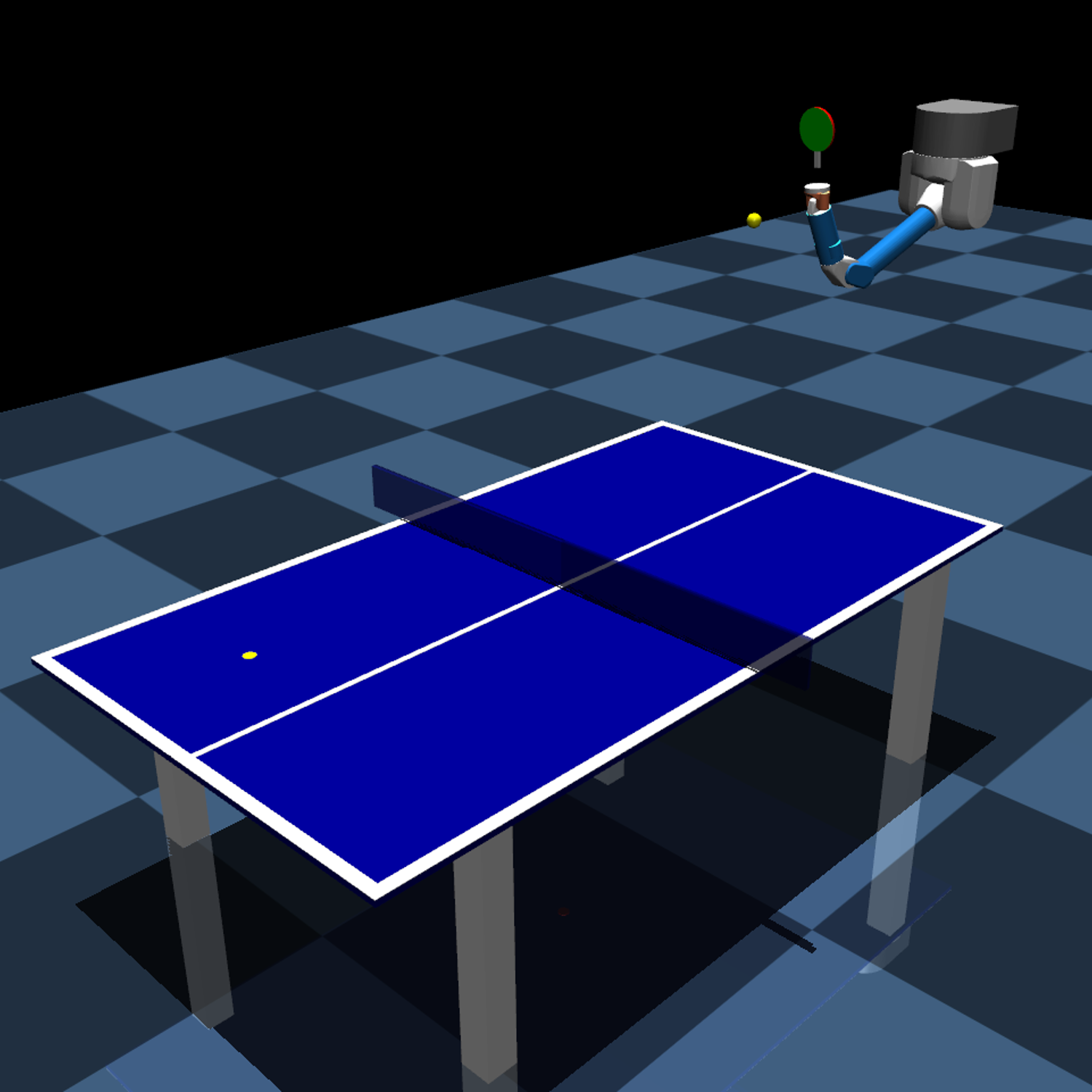}%
    \caption{Visualization of the four control tasks box pushing, hopper jumping, beer pong, and table tennis.}
    \label{fig:environments}
\end{figure}

{\color{black}
\subsection{Reacher5d}
\label{app:reacher}
For the Reacher task we modify the original OpenAI gym Reacher-v2 by adding three additional joints, resulting in a total of five joints. 
The task goal is still to minimize the distance between the goal point $\mathbf{p}_{goal}$ and the end-effector $\mathbf{p}$.
We, however, only sample the goal point for $y\geq0$, i.\,e. in the first two quadrants, to slightly reduce task complexity while maintaining the increased control complexity.
The observation space remains unchanged, unless for the sparse reward where we additionally add the current step value to make learning possible for step-based methods.
The context space only contains the coordinates of the goal position. 
The action space is the 5d equivalent to the original version.

For the reward the original setting leverages the goal distance
\[
 R_\text{goal} = \lVert \mathbf{p}-\mathbf{p}_{goal}\rVert_2
\]
and the action cost
\begin{align*}
    \tau_t = \sum_i^K (a^i_t)^2,
\end{align*}

\textbf{Dense Reward.} The dense reward in the 5d setting, hence, stays the same and the agent receives in each time step $t$
\begin{equation*}
    R_\text{tot} = - \tau_t - R_{\text{goal}}
\end{equation*}
\textbf{Sparse Reward.} The sparse reward only returns the task reward in the last time step $T$ and additionally adds a velocity penalty $R_\text{vel} = \sum_i^K (\dot{q}_T^i)^2$, where $\mathbf{\dot{q}}$ are the joint velocities, to avoid overshooting
\begin{equation*}
    R_\text{tot} = \begin{cases}
        - \tau_t 
        &t<T, 
        \\
        - \tau_t 
        -200 R_\text{goal}
        -10 R_\text{vel}
        &t=T.
        \end{cases}
\end{equation*}

}

\subsection{Box Pushing}
\label{app:box_pushing}
The goal of the box-pushing task is to move a box to a specified goal location and orientation using the seven DoF Franka Emika Panda.
Hence, the context space for this task is the goal position $x \in [0.3, 0.6]$, $y \in [-0.45, 0.45]$ and the goal orientation $\theta \in [0, 2\pi]$. 
{\color{black}In addition to the contexts, the observation space for the step-based algorithms contains the positions and velocities of the joint angles, as well as position and orientation quaternions for the actual box and the target. 
For the action space we use the torques per joint and additionally add gravity compensation in each time step, that does not have to be learnt.}
The task is considered successfully solved if the position distance $\leq 0.05$m and  the orientation error $\leq 0.5$rad.
For the total reward we consider different sub-rewards. 
First, the distance to the goal
\[
 R_\text{goal} = \lVert\mathbf{p}-\mathbf{p}_{goal}\rVert,
\]
where $\mathbf{p}$ is the box position and $\mathbf{p}_{goal}$ the goal position itself.
Second, the rotation distance
\[
R_\text{rotation} =\frac{1}{\pi} \arccos{|\mathbf{r}\cdot\mathbf{r}_{goal}|},
\]
where $\mathbf{r}$ and $\mathbf{r}_{goal}$ are the box orientation and goal orientation in quaternion, respectively.
Third, an incentive to keep the rod within the box
\[
R_\text{rod} = \text{clip}(||\mathbf{p}-\mathbf{h}_{pos}||, 0.05, 10)
\]
where $\mathbf{h}_{pos}$ is the position of the rod tip.
Fourth, a similar incentive that encourages to maintain the rod in a desired rotation
\[
R_\text{rod\_rotation} = \text{clip} (\frac{2}{\pi} \arccos{|\mathbf{h}_{rot}\cdot\mathbf{h}_0|} , 0.25, 2),
\]
where $\mathbf{h}_{rot}$ and $\mathbf{h}_{0}=(0.0, 1.0, 0.0, 0.0)$ are the current and desired rod orientation in quaternion, respectively.
And lastly, we utilize the following error
\[\text{err}(\mathbf{q}, \mathbf{\dot{q}}) = \sum_{i \in \{i | |q_i| > |q_i^{b}|\}}{(|q_i|-|q^{b}_i|)}
+ \sum_{j \in \{j | |\dot{q}_j| >|\dot{q}_j^{b}|\}}{(|\dot{q}_j|-|\dot{q}^{b}_j|)}.
\]
Here, $\mathbf{q}$, $\mathbf{\dot{q}}$, $\mathbf{q}^{b}$, and $\mathbf{\dot{q}}^{b}$ are the robot joint's position and velocity as well as their respective bounds.
Additionally, we consider an action cost in each time step $t$
\begin{align*}
    \tau_t = \sum_i^K (a^i_t)^2,
\end{align*}
where $K=7$ is the number of DoF.
Similar to the aforementioned reacher task, we consider both dense and sparse reward setups. 

\textbf{Dense Reward.} The dense reward provides information about the goal and rotation distance in each time step $t$ on top of the utility rewards
\begin{equation*}
    R_\text{tot} = 
    - R_{\text{rod}}
    - R_{\text{rod\_rotation}}
    - 5 e^{-4}\tau_t 
    - \text{err}(\bm{q}, \bm{\dot{q}})
    -3.5 R_{\text{goal}}
    - 2 R_{\text{rotation}}.
\end{equation*}
\textbf{Temporal Sparse Reward.} The time-dependent sparse reward is similar to the dense reward, but only returns the goal and rotation distance in the last time step $T$
\begin{equation*}
    R_\text{tot} = \begin{cases}
        - R_\text{rod}
        - R_\text{rod\_rotation}
        - 0.02 \tau_t 
        - \text{err}(\mathbf{q}, \mathbf{\dot{q}}),
        &t<T, 
        \\
        - R_\text{rod}
        - R_\text{rod\_rotation}
        - 0.02 \tau_t 
        - \text{err}(\mathbf{q}, \mathbf{\dot{q}})
        -350 R_\text{goal}
        -200R_\text{rotation},
        &t=T.
        \end{cases}
\end{equation*}
\textbf{Goal Switching.} To demonstrate the ability of our algorithm to handle the changing goal. We randomly switch to a new target at $20\%$ of the max episode length. To ensure that the new target is solvable within the given episode length, we sample its position near to the old target position. The new target position, denoted as $[x_{new}, y_{new}]$, is computed as follows:
\begin{equation*}
    [x_{new}, y_{new}] = [x_{old}, y_{old}] + [\Delta x, \Delta y],
\end{equation*}
where $\Delta x, \Delta y$ are randomly sampled within the range $[-0.25, 0.2]$. Additionally, the new target orientation is determined by uniformly sampling a value from the range of $[0, 2\pi]$.
\subsection{Hopper Jump}
\label{app:hopper_jump}
In the Hopper jump task the agent has to learn to jump as high as possible and land on a certain goal position at the same time. We consider five basis functions per joint resulting in an 15 dimensional weight space. The context is four-dimensional consisting of the initial joint angles $\theta \in [-0.5, 0]$, $\gamma \in [-0.2, 0]$, $\phi \in [0, 0.785]$ and the goal landing position $x \in [0.3, 1.35]$. 
{\color{black}The full observation space extends the original observation space from the OpenAI gym Hopper by adding the x-value of the goal position and the x-y-z difference between the goal point and the reference point at the Hopper's foot.
The action space is the same as for the original Hopper task.}
We consider a non-Markovian reward function for the episode-based algorithms and a step-based reward for PPO, which we have extensively designed to obtain the highest possible jump.

\textbf{Non-Markovian Reward.} In each time-step $t$ we provide an action cost 
\begin{align*}
    \tau_t = 10^{-3}\sum_i^K (a^i_t)^2,
\end{align*}
where $K=3$ is the number of DoF. In the last time-step $T$ of the episode we provide a reward which contains information about the whole episode as
\begin{align*}
    R_{height} &= 10h_{max}, \\
    R_{gdist} &= ||p_{foot, T} - p_{goal}||_2,\\
    R_{cdist} &= ||p_{foot,contact} - p_{goal}||_2,\\
    R_{healthy} &= \left\{\begin{array}{ll} 2 & \textrm{if } z_T \in [0.5, \infty] \textrm{and } \theta, \gamma, \phi \in [-\infty, \infty]\\
                                           0 & \textrm{else} \end{array}\right.,
\end{align*}
where $h_{max}$ is the maximum jump height in z-direction of the center of mass reached during the whole episode, $p_{foot, t}$ is the x-y-z position of the foot's heel at time step $t$, $p_{foot,contact}$ is the foot's heel position when having a contact with the ground after the first jump, $p_{goal}$ is the goal landing position of the heel. $R_{healthy}$ is a slightly modified reward of the healthy reward defined in the original hopper task. The hopper is considered as 'healthy' if the z position of the center of mass is within the range $[0.5m, \infty]$. This encourages the hopper to stand at the end of the episode. Note that all states need to be within the range $[-100, 100]$ for $R_{healthy}$. Since this is defined in the hopper task from OpenAI already, we haven't mentioned it here. The total reward at the end of an episode is given as
\begin{align*}
    R_{tot} = -\sum_{t=0}^T\tau_t + R_{height} + R_{gdist} + R_{cdist} + R_{healthy}.
\end{align*}

\textbf{Step-Based Reward.} We consider a step-based alternative reward such that PPO is also able to learn a meaningful behavior on this task. We have tuned the reward such that we can obtain the best performance. The observation space is the same as in the original hopper task from OpenAI extended with the goal landing position and the current distance of the foot's heel and the goal landing postion.  
We again consider the action cost in each time-step $t$
\begin{align*}
    \tau_t = 10^{-3}\sum_i^K (a^i_t)^2,
\end{align*}
and additionally consider the rewards
\begin{align*}
    R_{height, t} &= 3h_t\\
    R_{gdist, t} &= 3 ||p_{foot, t} - p_{goal}||_2 \\
    R_{healthy, t} &= \left\{\begin{array}{ll} 1 & \textrm{if } z_t \in [0.5, \infty] \textrm{and } \theta, \gamma, \phi \in [-\infty, \infty]\\
                                           0 & \textrm{else} \end{array}\right.,
\end{align*}
where these rewards are now returned to the agent in each time-step $t$, resulting in the reward per time-step 
\begin{align*}
    r_t(s_t, a_t) = -\tau_t + R_{height, t} + R_{gdist, t} + R_{healthy, t}.
\end{align*}

\subsection{Beer Pong}
\label{app:beer_pong}
In the Beer Pong task the $K=7$ Degrees of Freedom (DoF) robot has to throw a ball into a cup on a big table. The context is defined by the cup's two dimensional position on the table which lies in the range $x\in [-1.42, 1.42]$, $y\in [-4.05, -1.25]$. 
{\color{black}
For the step-based algorithms we consider cosine and sine of the robot's joint angles, the angle velocities, the ball's distance to the bottom of the cup, the ball's distance to the top of the cup, the cup position and the current time step.}
The action space for the step-based algorithms is defined as the torques for each joint, the parameter space for the episode-based methods is 15 dimensional which consists of the two weights for the basis functions per joint and the duration of the throwing trajectory, i.e. the ball release time.

We generally consider action penalties
\begin{align*}
    \tau_t = \frac{1}{K}\sum_i^K (a^i_t)^2,
\end{align*}
consisting of the sum of squared torques per joint.
For $t<T$ we consider the reward
\begin{align*}
    r_t(s_t,a_t) = -\alpha_t \tau_t,
\end{align*}
with $\alpha_t = 10^{-2}$.
For $t=T$ we consider the non-Markovian reward

\begin{equation*}
    R_{task} = \begin{cases}
    -4 
    - min(||p_{c,top}-p_{b,1:T}||_2^2) 
    - 0.5||p_{c,bottom}-p_{b,T}||_2^2 \cdots \\
    \cdots - 2||p_{c,bottom}-p_{b,k}||_2^2
    -\alpha_T\tau, 
    & \textrm{if cond. 1} \\
    -4 
    - min(||p_{c,top}-p_{b,1:T}||_2^2) 
    - 0.5||p_{c,bottom}-p_{b,T}||_2^2 
    -\alpha_T\tau, 
    & \textrm{if cond. 2} \\
    -2 
    - min(||p_{c,top}-p_{b,1:T}||_2^2) 
    - 0.5||p_{c,bottom}-p_{b,T}||_2^2 
    -\alpha_T\tau, 
    & \textrm{if cond. 3}\\
    -||p_{c,bottom}-p_{b,T}||_2^2 
    - \alpha_T \tau, 
    & \textrm{if cond. 4}
    \end{cases}
\end{equation*}

\begin{align*}
    R_{task} = \left\{
    \begin{array}{llll} -4 - min(||p_{c,top}-p_{b,1:T}||_2^2) - 0.5||p_{c,bottom}-p_{b,T}||_2^2 \cdots \\
                                    \cdots - 2||p_{c,bottom}-p_{b,k}||_2^2-\alpha_T\tau, & \textrm{if cond. 1} \\
                                    -4 - min(||p_{c,top}-p_{b,1:T}||_2^2) - 0.5||p_{c,bottom}-p_{b,T}||_2^2 -\alpha_T\tau, & \textrm{if cond. 2} \\
                                    -2 - min(||p_{c,top}-p_{b,1:T}||_2^2) - 0.5||p_{c,bottom}-p_{b,T}||_2^2 -\alpha_T\tau, & \textrm{if cond. 3} \\
                                                         -||p_{c,bottom}-p_{b,T}||_2^2 - \alpha_T \tau, & \textrm{if cond. 4}
                                                         \end{array}\right.,
\end{align*}
where $p_{c, top}$ is the position of the top edge of the cup, $p_{c, bottom}$ is the ground position of the cup, $p_{b,t}$ is the position of the ball at time point $t$, and $\tau$ is the squared mean torque over all joints during one rollout and $\alpha_T=10^{-4}$.  
The different conditions are:
\begin{itemize}
    \item cond. 1: The ball had a contact with the ground before having a contact with the table.
    \item cond. 2: The ball is not in the cup and had no table contact
    \item cond. 3: The ball is not in the cup and had table contact
    \item cond. 4: The ball is in the cup.
\end{itemize}
Note that $p_{b,k}$ is the ball's and the ground's contact position and is only given, if the ball had a contact with the ground first.

At time step $t=T$ we also give information whether the agent's chosen ball release time $B$ was reasonable
\begin{align*}
    R_{release} = \left\{\begin{array}{l}-30-10(B-B_{min})^2, ~~ \textrm{if } B < B_{min}\\
                                           -30-10(B-B_{max})^2, ~~ \textrm{if } B < B_{max} \end{array}\right.,
\end{align*}
where we define $B_{min} = 0.1s$ and $B_{max} = 1s$, such that the agent is encouraged to throw the ball within the time range $[B_{min}, B_{max}]$.

The total return over the whole episode is therefore given as 
\begin{align*}
    R_{tot} = \sum_{t=1}^{T-1} r_t(s_t,a_t) + R_{task} + R_{release}
\end{align*}

A throw is considered as successful if the ball is in the cup at the end of an episode.

\subsection{Table Tennis}
\label{app:table_tennis}
We consider table tennis for the entire table, i.\,e. incoming balls are anywhere on the side of the robot and goal locations anywhere on the opponents side. 
The goal is to use the 7 \glspl{dof} robotic arm to hit the incoming ball based on its landing position and return it as close as possible to the specified goal location.
As context space we consider the initial ball position $x \in [-1, -0.2]$, $y \in [-0.65, 0.65]$ and the goal position $x \in [-1.2, -0.2]$, $y \in [-0.6, 0.6]$. 
{\color{black}The full observation space again contains the positions and velocities of the joints on top of the above context information.
The torques of the joints make up the action space.}
For this experiment, we do not use any gravity compensation and allow in the episode-based setting to learn the start time $t_0$ and the trajectory duration $T$.
The task is considered successful if the returned ball lands on the opponent's side of the table and within $\leq 0.2$m to the goal location. 
The max episode length of the table tennis environment is $350$ steps. However, to accelerate the simulation, the episode will end immediately if any of the following terminated conditions are met: 
\begin{itemize}
    \item terminated cond. 1: A contact between the ball and the floor is detected,
    \item terminated cond. 2: The agent has hit the ball and then a contact between the ball and the table is detected. 
\end{itemize}

The reward signal in the table tennis environment is defined as 
\begin{equation*}
    r_{task} = \begin{cases}
        0, & \text{if cond. 1} \\
        0.2-0.2\tanh{(\min{||\mathbf{p}_{r}-\mathbf{p}_{b}||^2})}, & \text{if cond. 2} \\
        3-2\tanh{(\min{||\mathbf{p}_{r}-\mathbf{p}_{b}||^2})}-\tanh{(||\mathbf{p}_{l}-\mathbf{p}_{goal}||^2)}, &\text{if cond. 3}\\
        6-2\tanh{(\min{||\mathbf{p}_{r}-\mathbf{p}_{b}||^2})}-4\tanh{(||\mathbf{p}_{l}-\mathbf{p}_{goal}||}^2), &\text{if cond. 4}\\
        7-2\tanh{(\min{||\mathbf{p}_{r}-\mathbf{p}_{b}||^2})}-4\tanh{(||\mathbf{p}_{l}-\mathbf{p}_{goal}||}^2), &\text{if cond. 5}
        \end{cases}
\end{equation*}
where $\mathbf{p}_r$ is the position of racket center, $\mathbf{p}_b$ is the position of the ball, $\mathbf{p}_{l}$ is the ball landing position, $\mathbf{p}_{goal}$ is the target position. The different conditions are
\begin{itemize}
    \item cond. 1: the end of episode is not reached,
    \item cond. 2: the end of episode is reached,
    \item cond. 3: cond.2 is satisfied and robot did hit the ball,
    \item cond. 4: cond.3 is satisfied and the returned ball landed on the table,
    \item cond. 5: cond.4 is satisfied and the landing position is at the opponent's side.
\end{itemize}
The episode ends when any of the following conditions are met
\begin{itemize}
    \item the maximum horizon length is reached
    \item ball did land on the floor without hitting
    \item ball did land on the floor or table after hitting
\end{itemize}

For BBRL-PPO and BBRL-TRPL, the whole desired trajectory is obtained ahead of environment interaction, making use of this property we can collect some samples without physical simulation. The reward function based on this desired trajectory is defined as
\begin{equation*}
    r_{traj} = -\sum_{(i,j)}{|\tau_{ij}^d| - |q^b_j|} , \quad 
    (i,j) \in \{(i,j)\mid |\tau_{ij}^d| > |q^b_j|\}
\end{equation*}
where $\tau^d$ is the desired trajectory, $i$ is the time index, $j$ is the joint index, $q^b$ is the joint position upperbound. The desired trajectory is considered as invalid if $r_{traj} < 0$, an invalid trajectory will not be executed by robot. The overall reward for BBRL is defined as:
\begin{equation*}
    r = \begin{cases}
    r_{traj}, & r_{traj} < 0 \\
    r_{task}, & \text{otherwise}
    \end{cases}
\end{equation*}
\textbf{Goal Switching.} 
 To evaluate the capability of our approach in handling goal changes in the presence of non-Markovian reward, we designed a goal-switching task based on the table tennis environment. Given that the episode lengths are not fixed in this environment, we fixed the target changing time at the $99$-th step after the episode begins. To simplify the task and make it easier to visualize, we restricted the range of the randomly sampled initial target to the left half of the table, specifically $y \in [-0.65, 0], x \in [-1.2, 0.2]$. At the $99$-th step, there is $50\%$ of chance that the goal is switched to another random position from the right side of the table, namely $y \in [0, 0.65], x \in [-1.2, 0.2]$.  

\textbf{Wind as External Perturbation.}
To further investigate the performance of our approach in handling environmental perturbations, we introduced artificial wind to the environment. At the beginning of each episode, we randomly sample a value $f \in [-0.1, 0.1]$ to represent the constant wind force. This force was then applied as an external force to the ball at each simulation step. It's important to note that, in this specific task, we also augmented the observation space of the agent to include the velocity of the ball. By incorporating this information, the agent was able to infer the underlying "wind speed" and adjust its behavior accordingly. Since this information is not directly observable at the beginning of the episode, episode-based policies, struggled to solve the task. 

\section{Additional Evaluations}
\label{app:evaluations}
\begin{figure}[t!]
    \centering
    \begin{subfigure}{0.4\textwidth}
        \caption{Hopper Jump - Height Trajectory}
        \label{fig:hopper_jump_traj:traj}
        \resizebox{\textwidth}{!}{\input{tikz_plots/hopper_jump/hj_iqm_height_traj}}%
    \end{subfigure}%
    \begin{subfigure}{0.4\textwidth}
        \caption{5D Reacher - Sparse}
        \label{fig:hopper_jump_traj:reacher}
        \resizebox{\textwidth}{!}{\begin{tikzpicture}
\begin{axis}[
legend cell align={left},
legend style={
  fill opacity=0.8,
  draw opacity=1,
  text opacity=1,
  at={(0.97,0.03)},
  anchor=south east,
  draw=lightgray204
},
title={5D Reacher - Sparse},
tick align=outside,
tick pos=left,
x grid style={darkgray176},
xlabel={Number Environment Interactions},
xmajorgrids,
xmin=-1472000, xmax=31912000,
xtick style={color=black},
y grid style={darkgray176},
ylabel={Reward},
ymajorgrids,
ymin=-12500, ymax=5,
ytick style={color=black}
]
\path [draw=C8, fill=C8, opacity=0.2]
(axis cs:0,-804930.130645931)
--(axis cs:0,-1278894.09449712)
--(axis cs:400000,-16572.5208399925)
--(axis cs:800000,-27382.2598870615)
--(axis cs:1200000,-20781.0124473907)
--(axis cs:1600000,-45148.294311903)
--(axis cs:2000000,-20082.5248534612)
--(axis cs:2400000,-17448.479055946)
--(axis cs:2800000,-103206.15146879)
--(axis cs:3200000,-23192.3845824757)
--(axis cs:3600000,-58283.9074550735)
--(axis cs:4000000,-28636.3524705853)
--(axis cs:4400000,-32545.1620167667)
--(axis cs:4800000,-14407.48670961)
--(axis cs:5200000,-33564.2880822662)
--(axis cs:5600000,-22920.4363017282)
--(axis cs:6000000,-19840.4275005392)
--(axis cs:6400000,-26495.5362260327)
--(axis cs:6800000,-45718.811172274)
--(axis cs:7200000,-19548.5917888007)
--(axis cs:7600000,-36637.7472059447)
--(axis cs:8000000,-34488.2164246332)
--(axis cs:8400000,-17110.9504357945)
--(axis cs:8800000,-17693.3063968742)
--(axis cs:9200000,-14500.766995704)
--(axis cs:9600000,-17235.33149568)
--(axis cs:10000000,-19298.5469884752)
--(axis cs:10400000,-19243.3937680765)
--(axis cs:10800000,-16074.3957659797)
--(axis cs:11200000,-12364.5777672952)
--(axis cs:11600000,-20262.4735354012)
--(axis cs:12000000,-14126.63987363)
--(axis cs:12400000,-10643.4230272382)
--(axis cs:12800000,-16298.3035389125)
--(axis cs:13200000,-21273.3543641582)
--(axis cs:13600000,-14533.929151352)
--(axis cs:14000000,-14561.7585231272)
--(axis cs:14400000,-15066.2369797982)
--(axis cs:14800000,-11133.473057644)
--(axis cs:15200000,-14078.298017307)
--(axis cs:15600000,-15110.2952657327)
--(axis cs:16000000,-14872.3282812033)
--(axis cs:16400000,-16716.7497312435)
--(axis cs:16800000,-16139.973035384)
--(axis cs:17200000,-11400.5383604987)
--(axis cs:17600000,-10265.1840167385)
--(axis cs:18000000,-9981.72245764575)
--(axis cs:18400000,-9668.51891450025)
--(axis cs:18800000,-10541.4607907837)
--(axis cs:19200000,-16455.3922856197)
--(axis cs:19600000,-12449.0732760778)
--(axis cs:20000000,-16840.8545616757)
--(axis cs:20400000,-20849.0079780733)
--(axis cs:20800000,-13149.3912218525)
--(axis cs:21200000,-12774.3004152557)
--(axis cs:21600000,-16794.0186884842)
--(axis cs:22000000,-20000.8526771405)
--(axis cs:22400000,-20411.4860795577)
--(axis cs:22800000,-20245.3742300272)
--(axis cs:23200000,-20906.334386445)
--(axis cs:23600000,-22086.8820392477)
--(axis cs:24000000,-17500.7618157068)
--(axis cs:24400000,-8922.24153647975)
--(axis cs:24800000,-17873.7208003277)
--(axis cs:25200000,-12244.3871920452)
--(axis cs:25600000,-14638.9756391048)
--(axis cs:26000000,-11518.5623958585)
--(axis cs:26400000,-10327.412604866)
--(axis cs:26800000,-12112.9449201955)
--(axis cs:27200000,-27753.0977247535)
--(axis cs:27600000,-9231.7304188495)
--(axis cs:28000000,-12700.4111832012)
--(axis cs:28400000,-9233.65771457675)
--(axis cs:28800000,-12827.1825733297)
--(axis cs:29200000,-11917.257162812)
--(axis cs:29600000,-11621.0647866475)
--(axis cs:29600000,-4200.86796382875)
--(axis cs:29600000,-4200.86796382875)
--(axis cs:29200000,-4236.228010763)
--(axis cs:28800000,-3026.36408900125)
--(axis cs:28400000,-3809.1742956085)
--(axis cs:28000000,-4616.44681985225)
--(axis cs:27600000,-4063.27269159875)
--(axis cs:27200000,-7258.8986200075)
--(axis cs:26800000,-4051.84929265751)
--(axis cs:26400000,-2690.218663658)
--(axis cs:26000000,-4407.34407152625)
--(axis cs:25600000,-3517.62984714)
--(axis cs:25200000,-5596.52211195225)
--(axis cs:24800000,-3570.20632473925)
--(axis cs:24400000,-4177.52918310525)
--(axis cs:24000000,-4162.5543497175)
--(axis cs:23600000,-5147.1715048255)
--(axis cs:23200000,-4299.765578122)
--(axis cs:22800000,-3818.4895706375)
--(axis cs:22400000,-6412.125237483)
--(axis cs:22000000,-6994.67180463925)
--(axis cs:21600000,-5377.79959075225)
--(axis cs:21200000,-4000.334978613)
--(axis cs:20800000,-5432.06228207025)
--(axis cs:20400000,-5747.3403836475)
--(axis cs:20000000,-7664.5289134105)
--(axis cs:19600000,-6105.94830406025)
--(axis cs:19200000,-6310.594683876)
--(axis cs:18800000,-5017.9308956285)
--(axis cs:18400000,-3684.50408930725)
--(axis cs:18000000,-3482.55558418375)
--(axis cs:17600000,-4781.000901916)
--(axis cs:17200000,-5191.92466328226)
--(axis cs:16800000,-8259.04373392325)
--(axis cs:16400000,-4999.81107279025)
--(axis cs:16000000,-5940.2032437075)
--(axis cs:15600000,-5369.2695458555)
--(axis cs:15200000,-6451.91963178251)
--(axis cs:14800000,-5693.99233834325)
--(axis cs:14400000,-5764.1338813055)
--(axis cs:14000000,-4572.95982016425)
--(axis cs:13600000,-6966.8997944495)
--(axis cs:13200000,-8992.787871566)
--(axis cs:12800000,-5078.26902438575)
--(axis cs:12400000,-5591.434639889)
--(axis cs:12000000,-4982.9118448405)
--(axis cs:11600000,-5331.08390882225)
--(axis cs:11200000,-5700.426595168)
--(axis cs:10800000,-5618.57333190925)
--(axis cs:10400000,-3770.04573435775)
--(axis cs:10000000,-5696.608614398)
--(axis cs:9600000,-5722.98982227325)
--(axis cs:9200000,-7165.0507746175)
--(axis cs:8800000,-4952.021905382)
--(axis cs:8400000,-5077.9693937625)
--(axis cs:8000000,-4514.38141022476)
--(axis cs:7600000,-6174.05420301075)
--(axis cs:7200000,-6431.02442594125)
--(axis cs:6800000,-12637.3795918443)
--(axis cs:6400000,-9344.06129204726)
--(axis cs:6000000,-7043.449038697)
--(axis cs:5600000,-6866.70914288075)
--(axis cs:5200000,-10298.964978899)
--(axis cs:4800000,-5325.96123530275)
--(axis cs:4400000,-5626.63362145)
--(axis cs:4000000,-7713.0795632195)
--(axis cs:3600000,-8772.63975426075)
--(axis cs:3200000,-8827.36879114651)
--(axis cs:2800000,-10754.1427100465)
--(axis cs:2400000,-5348.1663490705)
--(axis cs:2000000,-6949.42007200225)
--(axis cs:1600000,-7823.32898501725)
--(axis cs:1200000,-6636.08707317825)
--(axis cs:800000,-5108.77938039)
--(axis cs:400000,-7201.46020312325)
--(axis cs:0,-804930.130645931)
--cycle;

\addplot [thick, C8, mark=*, mark size=0, mark options={solid}]
table {%
0 -1064468.61451158
400000 -10978.49556842
800000 -10397.16034207
1200000 -10310.74732736
1600000 -14283.00320912
2000000 -11712.0594854
2400000 -10599.77788235
2800000 -22525.18933141
3200000 -14629.81311805
3600000 -17725.4117261
4000000 -16969.98937502
4400000 -10515.49615643
4800000 -8531.91742166
5200000 -16970.42190651
5600000 -13532.43151365
6000000 -12149.82933308
6400000 -15352.28316259
6800000 -23782.43799484
7200000 -10980.40945137
7600000 -15620.00137169
8000000 -13242.45653198
8400000 -8729.51952771
8800000 -7655.02173237
9200000 -10152.67378537
9600000 -9581.22727557
10000000 -11430.87293246
10400000 -7051.42121806
10800000 -9584.09670305
11200000 -8154.64006133
11600000 -9832.93436525
12000000 -7329.39464119
12400000 -7865.76107402
12800000 -8402.81601146
13200000 -14842.8480109
13600000 -10036.91055484
14000000 -7089.97142322
14400000 -8963.71806051
14800000 -8278.69872128
15200000 -9232.47638948
15600000 -8921.87787325
16000000 -8311.4958769
16400000 -8487.37746984
16800000 -11838.81749544
17200000 -8488.89542955
17600000 -7164.89055584
18000000 -5989.47279515
18400000 -6113.92729756
18800000 -7004.23499418
19200000 -9249.12027928
19600000 -8358.71472537
20000000 -11789.78151165
20400000 -10369.59041593
20800000 -8889.2402405
21200000 -6528.47439316
21600000 -7354.27993123
22000000 -10326.02595758
22400000 -11319.33403145
22800000 -8731.52957495
23200000 -7952.08222538
23600000 -10282.05082861
24000000 -8043.50565888
24400000 -6085.83881868
24800000 -8203.46478492
25200000 -7881.35099742
25600000 -8168.60751249
26000000 -6707.59894717
26400000 -4952.25825964
26800000 -6989.63463294
27200000 -14470.64041691
27600000 -5626.3505008
28000000 -7238.83479514
28400000 -6050.25242169
28800000 -6453.87455266
29200000 -7532.2779253
29600000 -7916.66048917
};
\addlegendentry{SAC}
\end{axis}
\end{tikzpicture}}%
    \end{subfigure}%
    \caption{\subref{fig:hopper_jump_traj:traj} The improved performance on the Hopper Jump task is also demonstrated on the jumping profile for a fixed context. While \gls{method-bb} jumps once as high as possible, \gls{ppo} constantly tries to maximize the height at each time step which leads to several jumps throughout the episode and consequently to a lower maximum height.
    \subref{fig:hopper_jump_traj:reacher} Learning curve of \gls{sac} for the sparse reward of the 5D Reacher task.}
    \label{fig:hopper_jump_traj}
\end{figure}
\clearpage

\section{Hyperparameters}
\label{app:parameters}
For all methods, where applicable, we optimized the learning rate, sample size, batch size, number of layers, and the number of epochs. 
For all \gls{mp} based methods, we additionally optimized the number of basis functions. 
Moreover, we found that \gls{ndp} requires tuning of the scale of the predicted \gls{dmp} weights, which was hard-coded to $100$ in the original code base.
However, this value only worked for the Meta-World tasks, but not for the other tasks, hence we adjusted it to allow for a fair comparison.
The population size of ES is always half the number of samples because two function evaluations are used per parameter vector.

\begin{table}[ht]
\centering
\caption{Hyperparameters for the 5D Reacher experiments.}
\label{tab:reacher-HP}
\begin{adjustbox}{max width=\textwidth}
\begin{tabular}{lcccccccc}
                                 & \acrshort{ppo}        & \acrshort{ndp}        & \acrshort{trpl}       & \acrshort{sac}         & \acrshort{cmore} & \acrshort{es}         & \acrshort{method-bb-ppo}   & \acrshort{method-bb}   \\ 
\hline
\multicolumn{9}{l}{}                                                                                                                  \\
number samples                   & 16000      & 16000      & 16000      & 1000        & 120   & 200        & 64         & 64          \\
GAE $\lambda$                    & 0.95       & 0.95       & 0.95       & 0.95        & n.a.  & n.a.       & n.a.       & n.a.        \\
discount factor                  & 0.99       & 0.99       & 0.99       & 0.99        & n.a.  & n.a.       & n.a.       & n.a.        \\
\multicolumn{9}{l}{}                                                                                                                  \\ 
\hline
\multicolumn{9}{l}{}                                                                                                                  \\
$\epsilon_\mu$/$\epsilon$        & n.a.       & n.a.       & 0.005      & n.a.        & 0.1   & n.a.       & n.a.       & 0.05        \\
$\epsilon_\Sigma$                & n.a.       & n.a.       & 0.0005     & n.a.        & n.a.  & n.a.       & n.a.       & 0.0005      \\
\multicolumn{9}{l}{}                                                                                                                  \\ 
\hline
\multicolumn{9}{l}{}                                                                                                                  \\
optimizer                        & adam       & adam       & adam       & adam        & n.a.  & adam       & adam       & adam        \\
epochs                           & 10         & 10         & 20         & 1000        & n.a.  & n.a.       & 100        & 100         \\
learning rate                    & 3e-4       & 3e-4       & 5e-5       & 3e-4        & n.a.  & 1e-2       & 3e-4       & 3e-4        \\
use critic                       & True       & True       & True       & True        & False & False      & False      & False       \\
epochs critic                    & 10         & 10         & 10         & 1000        & n.a.  & n.a.       & n.a.       & n.a.        \\
learning rate critic (and alpha) & 3e-4       & 3e-4       & 3e-4       & 3e-4        & n.a.  & n.a.       & n.a.       & n.a.        \\
number minibatches               & 32         & 32         & 64         & n.a.        & n.a.  & n.a.       & 1          & 1           \\
batch size                       & n.a.       & n.a.       & n.a.       & 256         & n.a.  & n.a.       & n.a.       & n.a.        \\
buffer size                      & n.a.       & n.a.       & n.a.       & 1e6         & n.a.  & n.a.       & n.a.       & n.a.        \\
learning starts                  & 0          & 0          & 0          & 10000       & 0     & 0          & 0          & 0           \\
polyak\_weight                   & n.a.       & n.a.       & n.a.       & 5e-3        & n.a.  & n.a.       & n.a.       & n.a.        \\
trust region loss weight         & n.a.       & n.a.       & 10         & n.a.        & n.a.  & n.a.       & n.a.       & 10          \\
\multicolumn{9}{l}{}                                                                                                                  \\ 
\hline
\multicolumn{9}{l}{}                                                                                                                  \\
normalized observations          & True       & True       & True       & False       & False & False      & False      & False       \\
normalized rewards               & True       & True       & False      & False       & False & False      & False      & False       \\
observation clip                 & 10.0       & 10.0       & n.a.       & n.a.        & n.a.  & n.a.       & n.a.       & n.a.        \\
reward clip                      & 10.0       & 10.0       & n.a.       & n.a.        & n.a.  & n.a.       & n.a.       & n.a.        \\
critic clip                      & 0.2        & 0.2        & n.a.       & n.a.        & n.a.  & n.a.       & 0.2        & n.a.        \\
importance ratio clip            & 0.2        & 0.2        & n.a.       & n.a.        & n.a.  & n.a.       & 0.2        & n.a.        \\
\multicolumn{9}{l}{}                                                                                                                  \\ 
\hline
\multicolumn{9}{l}{}                                                                                                                  \\
hidden layers                    & {[}32, 32] & {[}32, 32] & {[}32, 32] & {[}128,128] & n.a.  & {[}32, 32] & {[}32, 32] & {[}32, 32]  \\
hidden layers critic             & {[}32, 32] & {[}32, 32] & {[}32, 32] & {[}128,128] & n.a.  & n.a.       & n.a.       & n.a.        \\
hidden activation                & tanh       & tanh       & tanh       & relu        & n.a.  & tanh       & tanh       & tanh        \\
initial std                      & 1.0        & 1.0        & 1.0        & 1.0         & 1.0   & 1.0        & 1.0        & 1.0         \\
\multicolumn{9}{l}{}                                                                                                                  \\ 
\hline
\multicolumn{9}{l}{}                                                                                                                  \\
number basis functions           & n.a.       & 5          & n.a.       & n.a.        & 5     & n.a.       & 5          & 5           \\
number zero basis                & n.a.       & n.a.       & n.a.       & n.a.        & 1     & n.a.       & 1          & 1           \\
weight scale                     & n.a.       & 20         & n.a.       & n.a.        & n.a.  & n.a.       & n.a.       & n.a.       
\end{tabular}
\end{adjustbox}
\end{table}

\begin{table}[ht]
\centering
\caption{Hyperparameters for the box pushing experiments.}
\label{tab:boxpushing-HP}
\begin{adjustbox}{max width=\textwidth}
\begin{tabular}{lcccccccc}
                                 & \acrshort{ppo}        & \acrshort{ndp}        & \acrshort{trpl}       & \acrshort{sac}         & \acrshort{es}   & \acrshort{method}   & \acrshort{method-bb-ppo}   & \acrshort{method-bb}   \\ 
\hline
\multicolumn{9}{l}{}                                                                                                  \\
number samples                   & 16000        & 16000        & 16000        & 1000         & 250      & 160       & 160          & 40          \\
GAE $\lambda$                    & 0.95         & 0.95         & 0.95         & 0.95         & n.a.     & n.a.      & n.a.         & n.a.         \\
discount factor                  & 1.0         & 0.99         & 1.0
& 0.99         & n.a.     & 1.0      & n.a.         & n.a.         \\
\multicolumn{9}{l}{}                                                                                                                      \\ 
\hline
\multicolumn{9}{l}{}                                                                                                                      \\
$\epsilon_\mu$                   & n.a.         & n.a.         & 0.005        & n.a.         & n.a.     &0.05         & n.a.         & 0.05        \\
$\epsilon_\Sigma$                & n.a.         & n.a.         & 0.00005      & n.a.         & n.a.     &0.0005         & n.a.         & 0.0005       \\
\multicolumn{9}{l}{}                                                                                                                      \\ 
\hline
\multicolumn{9}{l}{}                                                                                                                      \\
optimizer                        & adam         & adam         & adam         & adam         & adam     & adam         & adam         & adam         \\
epochs                           & 10           & 10           & 20           & 1000         & n.a.     & 20           & 100          & 20          \\
learning rate                    & 1e-4         & 1e-4         & 5e-5         & 1e-4         & 1e-2     & 3e-4         & 1e-4         & 3e-4         \\
use critic                       & True         & True         & True         & True         & False    & True         & True         & True         \\
epochs critic                    & 10           & 10           & 10           & 1000         & n.a.     & 10    & 100          & 10          \\
learning rate critic (and alpha) & 1e-4         & 1e-4         & 2e-4         & 1e-4         & n.a.     & 3e-4    & 1e-4         & 3e-4         \\
number minibatches               & 40           & 32           & 40           & n.a.         & n.a.     & 1    & 1            & 1            \\
batch size                       & n.a.         & n.a.         & n.a.         & 256          & n.a.     & n.a.    & n.a.         & n.a.         \\
buffer size                      & n.a.         & n.a.         & n.a.         & 1e6          & n.a.     & n.a.    & n.a.         & n.a.         \\
learning starts                  & 0            & 0            & 0            & 10000        & 0        & 0    & 0            & 0            \\
polyak\_weight                   & n.a.         & n.a.         & n.a.         & 5e-3         & n.a.     & n.a.    & n.a.         & n.a.         \\
trust region loss weight         & n.a.         & n.a.         & 10           & n.a.         & n.a.     & 10    & n.a.         & 10           \\
\multicolumn{9}{l}{}                                                                                                                      \\ 
\hline
\multicolumn{9}{l}{}                                                                                                                      \\
normalized observations          & True         & True         & True         & False        & False    & False    & False        & False        \\
normalized rewards               & True         & True         & False        & False        & False    & False    & False        & False        \\
observation clip                 & 10.0         & 10.0         & n.a.         & n.a.         & n.a.     & n.a.    & n.a.         & n.a.         \\
reward clip                      & 10.0         & 10.0         & n.a.         & n.a.         & n.a.     & n.a.    & n.a.         & n.a.         \\
critic clip                      & 0.2          & 0.2          & n.a.         & n.a.         & n.a.     & n.a.    & 0.2          & n.a.         \\
importance ratio clip            & 0.2          & 0.2          & n.a.         & n.a.         & n.a.     & n.a.    & 0.2          & n.a.         \\
\multicolumn{9}{l}{}                                                                                                                      \\ 
\hline
\multicolumn{9}{l}{}                                                                                                                      \\
hidden layers                    & {[}256, 256] & {[}256, 256] & {[}256, 256] & {[}256, 256] & {[}256, 256] &{[}128, 128] & {[}128, 128] & {[}128, 128] \\
hidden layers critic             & {[}256, 256] & {[}256, 256] & {[}256, 256] & {[}256, 256] & n.a.         &{[}32, 32]   & {[}32, 32]   & {[}32, 32]   \\
hidden activation                & tanh         & tanh         & tanh         & relu         & tanh         & relu        & tanh         & relu         \\
initial std                      & 1.0          & 1.0          & 1.0          & 1.0          & 1.0          & 1.0         & 1.0          & 1.0          \\
\multicolumn{9}{l}{}                                                                                                                      \\ 
\hline
\multicolumn{9}{l}{}                                                                                                                                        \\
\gls{mp} type                    & n.a.         & \gls{dmp}    & n.a.         & n.a.         & n.a.        & \gls{pdmp}   & \gls{promp}    & \gls{promp} \\
number basis functions           & n.a.         & 5            & n.a.         & n.a.         & n.a.         & 4           & 5            & 5            \\
number zero basis                & n.a.         & n.a.         & n.a.         & n.a.         & n.a.         & 0           & 1            & 1            \\
$k$                              & n.a.         & 5            & n.a.         & n.a.         & n.a.         & 25          & 100          & 100          \\
weight scale                     & n.a.         & 10           & n.a.         & n.a.         & n.a.         & n.a.        & n.a.         & n.a.       
\end{tabular}
\end{adjustbox}
\end{table}

\begin{table}[ht]
\centering
\centering
\caption{Hyperparameters for the Meta-World experiments.}
\label{tab:metaworld-HP}
\begin{adjustbox}{max width=\textwidth}
\begin{tabular}{lcccccccc}
                                 & \acrshort{ppo}        & \acrshort{ndp}        & \acrshort{trpl}       & \acrshort{sac}         & \acrshort{es}         & \acrshort{method}   & \acrshort{method-bb-ppo}   & \acrshort{method-bb}   \\ 
\hline
\multicolumn{9}{l}{}                                                                                                                  \\
number samples                   & 16000        & 16000        & 16000        & 1000         & 200          & 64           & 16         & 16         \\
GAE $\lambda$                    & 0.95         & 0.95         & 0.95         & 0.95         & n.a.         & 1            & n.a.       & n.a.       \\
discount factor                  & 0.99         & 0.99         & 0.99         & 0.99         & n.a.         & 1            & n.a.       & n.a.       \\
\multicolumn{9}{l}{}                                                                                                                                 \\
\hline
\multicolumn{9}{l}{}                                                                                                                                 \\
$\epsilon_\mu$                   & n.a.         & n.a.         & 0.005        & n.a.         & n.a.         & 0.075        & n.a.       & 0.005      \\
$\epsilon_\Sigma$                & n.a.         & n.a.         & 0.0005       & n.a.         & n.a.         & 0.0005       & n.a.       & 0.0005     \\
\multicolumn{9}{l}{}                                                                                                                                 \\
\hline
\multicolumn{9}{l}{}                                                                                                                                 \\
optimizer                        & adam         & adam         & adam         & adam         & adam         & adam         & adam       & adam       \\
epochs                           & 10           & 10           & 20           & 1000         & n.a.         & 10           & 100        & 100        \\
learning rate                    & 3e-4         & 3e-4         & 5e-5         & 3e-4         & 1e-2         & 5e-5         & 3e-4       & 3e-4       \\
use critic                       & True         & True         & True         & True         & False        & True         & False      & False      \\
epochs critic                    & 10           & 10           & 10           & 1000         & n.a.         & 10           & n.a.       & n.a.       \\
learning rate critic (and alpha) & 3e-4         & 3e-4         & 3e-4         & 3e-4         & n.a.         & 3e-4         & n.a.       & n.a.       \\
number minibatches               & 32           & 32           & 64           & n.a.         & n.a.         & 32           & 1          & 1          \\
batch size                       & n.a.         & n.a.         & n.a.         & 256          & n.a.         & n.a.         & n.a.       & n.a.        \\
buffer size                      & n.a.         & n.a.         & n.a.         & 1e6          & n.a.         & n.a.         & n.a.       & n.a.        \\
learning starts                  & 0            & 0            & 0            & 10000        & 0            & 0            & 0          & 0           \\
polyak\_weight                   & n.a.         & n.a.         & n.a.         & 5e-3         & n.a.         & n.a.         & n.a.       & n.a.        \\
trust region loss weight         & n.a.         & n.a.         & 10           & n.a.         & n.a.         & 10           & n.a.       & 10         \\
\multicolumn{9}{l}{}                                                                                                                                 \\
\hline
\multicolumn{9}{l}{}                                                                                                                                 \\
normalized observations          & True         & True         & True         & False        & False        & True         & False      & False      \\
normalized rewards               & True         & True         & False        & False        & False        & False        & False      & False      \\
observation clip                 & 10.0         & 10.0         & n.a.         & n.a.         & n.a.         & n.a.         & n.a.       & n.a.       \\
reward clip                      & 10.0         & 10.0         & n.a.         & n.a.         & n.a.         & n.a.         & n.a.       & n.a.       \\
critic clip                      & 0.2          & 0.2          & n.a.         & n.a.         & n.a.         & n.a.         & 0.2        & n.a.       \\
importance ratio clip            & 0.2          & 0.2          & n.a.         & n.a.         & n.a.         & n.a.         & 0.2        & n.a.       \\
\multicolumn{9}{l}{}                                                                                                                                 \\
\hline
\multicolumn{9}{l}{}                                                                                                                                 \\
hidden layers                    & {[}128, 128] & {[}128, 128] & {[}128, 128] & {[}256, 256] & {[}128, 128] & {[}256, 256] & {[}32, 32] & {[}32, 32] \\
hidden layers critic             & {[}128, 128] & {[}128, 128] & {[}128, 128] & {[}256, 256] & n.a.         & {[}256, 256] & n.a.       & n.a.       \\
hidden activation                & tanh         & tanh         & tanh         & relu         & tanh         & tanh         & tanh       & relu       \\
initial std                      & 1.0          & 1.0          & 1.0          & 1.0          & 1.0          & 1.0          & 1.0        & 1.0        \\
\multicolumn{9}{l}{}                                                                                                                                 \\
\hline
\multicolumn{9}{l}{}                                                                                                                                 \\
\gls{mp} type                    & n.a.         & \gls{dmp}    & n.a.         & n.a.         & n.a.         & \gls{pdmp}   & \gls{promp}& \gls{promp}\\
number basis functions           & n.a.         & 5            & n.a.         & n.a.         & n.a.         & 3            & 5          & 5          \\
number zero basis                & n.a.         & n.a.         & n.a.         & n.a.         & n.a.         & n.a.         & 1          & 1          \\
$k$                              & n.a.         & n.a.         & n.a.         & n.a.         & n.a.         & 100          & n.a.       & n.a.       \\
weight scale                     & n.a.         & 100          & n.a.         & n.a.         & n.a.         & 10           & 10         & 10        
\end{tabular}
\end{adjustbox}
\end{table}

\begin{table}[h]
\centering
\centering
\caption{Hyperparameters for the Hopper Jump experiments.}
\label{tab:hopper-HP}
\begin{adjustbox}{max width=\textwidth}
\begin{tabular}{lccccccc}
                                 & \acrshort{ppo}        & \acrshort{trpl}       & \acrshort{sac}         & \acrshort{cmore}    & \acrshort{es}     & \acrshort{method-bb-ppo}   & \acrshort{method-bb}   \\ 
\hline
\multicolumn{8}{l}{}                                                                                              \\
number samples                   & 16384        & 16384        & 1000         & 60    & 200          & 64         & 64          \\
GAE $\lambda$                    & 0.95         & 0.95         & 0.95         & n.a.  & n.a.         & n.a.       & n.a.        \\
discount factor                  & 0.99         & 0.99         & 0.99         & n.a.  & n.a.         & n.a.       & n.a.        \\
\multicolumn{8}{l}{}                                                                                              \\ 
\hline
\multicolumn{8}{l}{}                                                                                              \\
$\epsilon_\mu$/$\epsilon$        & n.a.         & 0.005        & n.a.         & 0.1   & n.a.         & n.a.       & 0.005       \\
$\epsilon_\Sigma$                & n.a.         & 0.00005      & n.a.         & n.a.  & n.a.         & n.a.       & 0.0005      \\
\multicolumn{8}{l}{}                                                                                              \\ 
\hline
\multicolumn{8}{l}{}                                                                                              \\
optimizer                        & adam         & adam         & adam         & n.a.  & adam         & adam       & adam        \\
epochs                           & 10           & 20           & 1000         & n.a.  & n.a.         & 100        & 100         \\
learning rate                    & 3e-4         & 5e-5         & 1e-4         & n.a.  & 0.01         & 1e-4       & 5e-5        \\
use critic                       & True         & True         & True         & False & False        & False      & False       \\
epochs critic                    & 10           & 10           & 1000         & n.a.  & n.a.         & n.a.       & n.a.        \\
learning rate critic (and alpha) & 3e-4         & 3e-4         & 1e-4         & n.a.  & n.a.         & n.a.       & n.a.        \\
number minibatches               & 32           & 64           & n.a.         & n.a.  & n.a.         & 1          & 1           \\
batch size                       & n.a.         & n.a.         & 256          & n.a.  & n.a.         & n.a.       & n.a.        \\
buffer size                      & n.a.         & n.a.         & 1e6          & n.a.  & n.a.         & n.a.       & n.a.        \\
learning starts                  & 0            & 0            & 10000        & 0     & 0            & 0          & 0           \\
polyak\_weight                   & n.a.         & n.a.         & 5e-3         & n.a.  & n.a.         & n.a.       & n.a.        \\
trust region loss weight         & n.a.         & 10           & n.a.         & n.a.  & n.a.         & n.a.       & 25          \\
\multicolumn{8}{l}{}                                                                                              \\ 
\hline
\multicolumn{8}{l}{}                                                                                              \\
normalized observations          & True         & True         & False        & False & False        & False      & False       \\
normalized rewards               & True         & False        & False        & False & False        & False      & False       \\
observation clip                 & 10.0         & n.a.         & n.a.         & n.a.  & n.a.         & n.a.       & n.a.        \\
reward clip                      & 10.0         & n.a.         & n.a.         & n.a.  & n.a.         & n.a.       & n.a.        \\
critic clip                      & 0.2          & n.a.         & n.a.         & n.a.  & n.a.         & 0.2        & n.a.        \\
importance ratio clip            & 0.2          & n.a.         & n.a.         & n.a.  & n.a.         & 0.2        & n.a.        \\
\multicolumn{8}{l}{}                                                                                              \\ 
\hline
\multicolumn{8}{l}{}                                                                                              \\
hidden layers                    & {[}128, 128] & {[}128, 128] & {[}128, 128] & n.a   & {[}128, 128] & {[}32, 32] & {[}32, 32]  \\
hidden layers critic             & {[}128, 128] & {[}128, 128] & {[}128, 128] & n.a   & n.a          & n.a        & n.a         \\
hidden activation                & tanh         & tanh         & relu         & n.a.  & tanh         & tanh       & tanh        \\
initial std                      & 1.0          & 1.0          & 1.0          & 1.0   & 1.0          & 1.0        & 1.0         \\
\multicolumn{8}{l}{}                                                                                              \\ 
\hline
\multicolumn{8}{l}{}                                                                                              \\
number basis functions           & n.a.         & n.a.         & n.a.         & 5    & n.a.         & 5          & 5           \\
number zero basis                & n.a.         & n.a.         & n.a.         & 1    & n.a.         & 1          & 1           \\
\end{tabular}
\end{adjustbox}
\end{table}

\begin{table}[ht]
\centering
\centering
\caption{Hyperparameters for the Beer Pong experiments.}
\label{tab:beerpong}
\begin{adjustbox}{max width=\textwidth}
\begin{tabular}{lcccc}
                                 & PPO          & CMORE & BBRL-PPO   & BBRL-TRPL   \\ 
\hline
\multicolumn{5}{l}{}                                                               \\
number samples                   & 16384        & 60    & 160        & 160         \\
GAE $\lambda$                    & 0.95         & n.a.  & n.a.       & n.a.        \\
discount factor                  & 0.99         & n.a.  & n.a.       & n.a.        \\
\multicolumn{5}{l}{}                                                               \\ 
\hline
\multicolumn{5}{l}{}                                                               \\
$\epsilon_\mu$/$\epsilon$        & n.a.         & 0.1   & n.a.       & 0.005       \\
$\epsilon_\Sigma$                & n.a.         & n.a.  & n.a.       & 0.0005      \\
\multicolumn{5}{l}{}                                                               \\ 
\hline
\multicolumn{5}{l}{}                                                               \\
optimizer                        & adam         & n.a.  & adam       & adam        \\
epochs                           & 10           & n.a.  & 100        & 100         \\
learning rate                    & 3e-4         & n.a.  & 1e-4       & 5e-5        \\
use critic                       & True         & False & False      & False       \\
epochs critic                    & 10           & n.a.  & n.a.       & n.a.        \\
learning rate critic (and alpha) & 3e-4         & n.a.  & n.a.       & n.a.        \\
number minibatches               & 32           & n.a.  & 1          & 1           \\
trust region loss weight         & n.a.         & n.a.  & n.a.       & 25          \\
\multicolumn{5}{l}{}                                                               \\ 
\hline
\multicolumn{5}{l}{}                                                               \\
normalized observations          & True         & False & False      & False       \\
normalized rewards               & True         & False & False      & False       \\
observation clip                 & 10.0         & n.a.  & n.a.       & n.a.        \\
reward clip                      & 10.0         & n.a.  & n.a.       & n.a.        \\
critic clip                      & 0.2          & n.a.  & 0.2        & n.a.        \\
importance ratio clip            & 0.2          & n.a.  & 0.2        & n.a.        \\
\multicolumn{5}{l}{}                                                               \\ 
\hline
\multicolumn{5}{l}{}                                                               \\
hidden layers                    & {[}128, 128] & n.a.  & {[}32, 32] & {[}32, 32]  \\
hidden layers critic             & {[}128, 128] & n.a.  & n.a.       & n.a.        \\
hidden activation                & tanh         & n.a.  & tanh       & tanh        \\
initial std                      & 1.0          & 1.0   & 1.0        & 1.0         \\
\multicolumn{5}{l}{}                                                               \\ 
\hline
\multicolumn{5}{l}{}                                                               \\
number basis functions           & n.a.         & 2     & 2          & 2           \\
number zero basis                & n.a.         & 2     & 2          & 2           \\
\end{tabular}
\end{adjustbox}
\end{table}

\begin{table}[h]
\centering
\centering
\caption{Hyperparameters for the Table Tennis experiments.}
\label{tab:table tennis}
\begin{adjustbox}{max width=\textwidth}
\begin{tabular}{lccccc}
                                 & PPO          & TRPL      & \acrshort{method}    & BBRL-PPO & BBRL-TRPL  \\ 
\hline
\multicolumn{6}{l}{}                                                                   \\
number samples                   & 16000        & 16000    & 360           & 200      & 200        \\
GAE $\lambda$                    & 0.95         & 0.95     & n.a.    & n.a.     & n.a.       \\
discount factor                  & 0.99         & 0.99     & 1.0    & n.a.     & n.a.       \\
\multicolumn{6}{l}{}                                                                   \\ 
\hline
\multicolumn{6}{l}{}                                                                   \\
$\epsilon_\mu$                   & n.a.         & 0.0005   & 0.005    & n.a.     & 0.0005     \\
$\epsilon_\Sigma$                & n.a.         & 0.00005  & 0.0005    & n.a.     & 0.00005    \\
\multicolumn{6}{l}{}                                                                   \\ 
\hline
\multicolumn{6}{l}{}                                                                   \\
optimizer                        & adam         & adam     & adam     & adam     & adam       \\
epochs                           & 10           & 20       & 20    & 100      & 100        \\
learning rate                    & 1e-4         & 5e-5     & 2e-4    & 1e-4     & 3e-4       \\
use critic                       & True         & True     & True    & True     & True       \\
epochs critic                    & 10           & 10       & 10    & 100      & 100        \\
learning rate critic (and alpha) & 1e-4         & 1e-4     & 2e-4    & 1e-4     & 3e-4       \\
number minibatches               & 40           & 40       & 1    & 1        & 1          \\
trust region loss weight         & n.a.         & 10.0     & 10    & n.a.     & 25         \\
\multicolumn{6}{l}{}                                                                   \\ 
\hline
\multicolumn{6}{l}{}                                                                   \\
normalized observations          & True         & True     & False    & False    & False      \\
normalized rewards               & True         & False    & False    & False    & False      \\
observation clip                 & 10.0         & n.a.     & n.a.    & n.a.     & n.a.       \\
reward clip                      & 10.0         & n.a.     & n.a.    & n.a.     & n.a.       \\
critic clip                      & 0.2          & n.a.     & n.a.     & 0.2      & n.a.       \\
importance ratio clip            & 0.2          & n.a.     & n.a.    & 0.2      & n.a.       \\
\multicolumn{6}{l}{}                                                                   \\ 
\hline
\multicolumn{6}{l}{}                                                                   \\
hidden layers                    & {[}256, 256] & {[}256, 256] & {[}256] & {[}256]  & {[}256]    \\
hidden layers critic             & {[}256, 256] & {[}256, 256] & {[}256] & {[}256]  & {[}256]    \\
hidden activation                & tanh         & tanh         & relu    & tanh     & relu       \\
initial std                      & 1.0          & 1.0          & 1.0     & 1.0      & 1.0        \\
\multicolumn{6}{l}{}                                                                   \\ 
\hline
\multicolumn{6}{l}{}                                                                   \\
\gls{mp} type                    & n.a.         & n.a.         & \gls{pdmp} & \gls{promp}   & \gls{promp}\\
number basis functions           & n.a.         & n.a.         & 3       & 3        & 3          \\
number zero basis                & n.a.         & n.a.         & 0       & 1        & 1          \\
$k$                              & n.a.         & n.a.         & 50      & n.a.     & n.a.        \\
weight scale                     & n.a.         & n.a.         & n.a.    & n.a.     & n.a.
\end{tabular}
\end{adjustbox}
\end{table}

\clearpage

\end{document}